\documentclass[twoside]{article}

\usepackage[accepted]{aistats2024}
\usepackage{natbib}
\usepackage[utf8]{inputenc} 
\usepackage[T1]{fontenc}    
\usepackage{hyperref}       
\usepackage{url}            
\usepackage{booktabs}       
\usepackage{amsfonts}       
\usepackage{nicefrac}       
\usepackage{microtype}      
\usepackage{xcolor}         
\usepackage{algpseudocode}
\usepackage{algorithm}
\usepackage{lipsum}
\usepackage{wrapfig}
\usepackage{mathtools}
\usepackage{amsthm}
\usepackage{amsmath,amsfonts,amssymb,color}
\usepackage{mathtools}
\usepackage{dsfont}
\usepackage{epsfig}
\usepackage{epstopdf}
\usepackage{pifont}
\usepackage{caption}
\usepackage{subcaption}
\usepackage{wrapfig}

\usepackage[utf8]{inputenc} 
\usepackage[T1]{fontenc}    
\usepackage{hyperref}       
\usepackage{url}            
\usepackage{booktabs}       
\usepackage{amsfonts}       
\usepackage{nicefrac}       
\usepackage{microtype}      
\usepackage{xcolor}         
\usepackage{algpseudocode}
\usepackage{algorithm}
\usepackage{lipsum}
\usepackage{wrapfig}
\usepackage{mathtools}
\usepackage{amsthm}
\usepackage{amsmath,amsfonts,amssymb,color}
\usepackage{mathtools}
\usepackage{dsfont}
\usepackage{epsfig}
\usepackage{epstopdf}
\usepackage{pifont}

\newcommand\tdrt{\tilde{r}_{t}}
\newcommand\tdxt{\tilde{\boldsymbol{\theta}}_t}
\newcommand\tdxtp{\tilde{\boldsymbol{\theta}}_{t+1}}
\newcommand\tdrp{\tilde{r}_{t+1}}
\newcommand\e{{\mathbf{d}}}
\newcommand\thet{{\boldsymbol{\theta}}}
\newcommand\xk{{\boldsymbol{\theta}_t}}
\newcommand\xktau{{\boldsymbol{\theta}_{t-\tau_t}}}
\newcommand\xs{{\boldsymbol{\theta}^*}}
\newcommand\xkp{{\boldsymbol{\theta}_{t+1}}}

\newcommand\ok{o_t}

\newcommand{\E}[1]{\mathbb{E}\left[#1\right]}

\newcommand{\tmix}{\tau_{mix}}
\newcommand{\tmax}{\tau_{max}}
\newcommand{\et}{\mathbf{e}_t}
\newcommand{\bfg}{\mathbf{g}}
\newcommand{\barg}{\bar{\mathbf{g}}}
\newcommand{\bfgofxo}{\mathbf{g}(\xk, o_t)}
\newcommand{\eqal}[2]{\begin{equation}\begin{aligned}\label{#1}
			#2
\end{aligned}\end{equation}}

\newcommand{\oktau}{o_{t-\tau_t}}
\newcommand\xktaua{{{\thet}_{t-\tau}}}

\newtheorem{problem}{Problem}
\newtheorem{definition}{Definition}
\newtheorem{theorem}{Theorem}
\newtheorem{lemma}{Lemma}
\newtheorem{proposition}{Proposition}
\newtheorem{corollary}{Corollary}
\newtheorem{remark}{Remark}
\newtheorem{assumption}{Assumption}
\usepackage{breqn}
\usepackage{array}

\DeclarePairedDelimiterX{\norm}[1]{\lVert}{\rVert}{#1}
\definecolor{mygreen}{rgb}{0.0, 0.5, 0.0}
\definecolor{winered}{rgb}{0.8,0,0}
\definecolor{myblue}{rgb}{0,0,0.8}

\hypersetup{
    colorlinks=true,
    linkcolor={winered},
    citecolor={myblue}
}

\begin{document}

%
\runningtitle{Stochastic Approximation with Delayed Updates: Finite-Time Rates under Markovian Sampling}

%
\runningauthor{Adibi, Dal Fabbro, Schenato, Kulkarni, Poor, Pappas, Hassani, Mitra}

\twocolumn[
\aistatstitle{Stochastic Approximation with Delayed Updates: \\ Finite-Time Rates under Markovian Sampling}
\aistatsauthor{ Arman Adibi$^*$ \And Nicolò Dal Fabbro$^*$ \And Luca Schenato \And Sanjeev Kulkarni}
\aistatsaddress{ Princeton  University \And   University of Pennsylvania \And  University of Padova \And  Princeton  University} 
\aistatsauthor{  H. Vincent Poor\And George J. Pappas\And Hamed Hassani \And Aritra Mitra}
\aistatsaddress{ Princeton  University \And   University of Pennsylvania \And  University of Pennsylvania \And  NC State University} 
]
\def\thefootnote{*}\footnotetext{These authors contributed equally to this work.}\def\thefootnote{\arabic{footnote}}

\begin{abstract} Motivated by applications in large-scale and multi-agent reinforcement learning, we study the non-asymptotic performance of stochastic approximation (SA) schemes with delayed updates under Markovian sampling. While the effect of delays has been extensively studied for optimization, the manner in which they interact with the underlying Markov process to shape the finite-time performance of SA remains poorly understood. In this context, our first main contribution is to show that under time-varying bounded delays, the delayed SA update rule guarantees exponentially fast convergence of the \emph{last iterate} to a ball around the SA operator's fixed point. Notably, our bound is \emph{tight} in its dependence on both the maximum delay $\tau_{max}$, and the mixing time $\tau_{mix}$. To achieve this tight bound, we develop a novel inductive proof technique that, unlike various existing delayed-optimization analyses, relies on establishing uniform boundedness of the iterates. As such, our proof may be of independent interest. Next, to mitigate the impact of the maximum delay on the convergence rate, we provide the first finite-time analysis of a delay-adaptive SA scheme under Markovian sampling. In particular, we show that the exponent of convergence of this scheme gets scaled down by $\tau_{avg}$, as opposed to $\tau_{max}$ for the vanilla delayed SA rule; here, $\tau_{avg}$ denotes the average delay across all iterations. Moreover, the adaptive scheme requires no prior knowledge of the delay sequence for step-size tuning. Our theoretical findings shed light on the finite-time effects of delays for a broad class of algorithms, including TD learning, Q-learning, and stochastic gradient descent under Markovian sampling.
\end{abstract}
\section{INTRODUCTION}
\label{sec:intro}

The goal of Stochastic Approximation (SA) theory, introduced by~\cite{robbins1951stochastic}, is to find the root (or fixed point) of an operator, given access to noisy observations. The general framework of SA finds applications in various fields like control, optimization, and reinforcement learning (RL). Recently, the surge of interest in distributed and asynchronous learning has motivated the study of variants of SA - e.g., stochastic gradient descent (SGD) - in the presence of delays. This leads to the study of iterative optimization algorithms where the gradients used for the iterative updates are computed at potentially stale iterates from the past~\citep{feyzmahdavian2016asynchronous, koloskova2022sharper}. Notably, the literature on understanding the effects of delays in SA has focused almost exclusively on optimization. 

In particular, while delayed versions of more general SA algorithms have been shown to perform well in practice in the context of asynchronous and multi-agent RL~\citep{bouteiller2020reinforcement}, there is little to no theory to substantiate these empirical observations. Bridging the above gap is the main objective of this paper. However, this task is non-trivial since the transition from optimization to RL requires contending with one major challenge: in the latter, the noisy observations are typically generated from a Markov process. As such, unlike the i.i.d. sampling assumption in optimization, the observations in RL are \textit{temporally correlated}. The interplay between such temporal correlations and delayed updates remains poorly understood in RL. Given this motivation, we provide the first comprehensive \textit{finite-time} analysis of SA schemes under Markovian sampling and delayed updates. As we explain later in the paper, this entails the development of novel analysis techniques that may be of independent interest to both the optimization and RL communities. 


\subsection{Related Works}
To position our work and contributions, we now provide a brief summary of relevant literature. A more elaborate survey is deferred to Appendix~\ref{sec:related}. 

$\bullet$ \textit{SA under Markovian Sampling.} The asymptotic convergence of SA under correlated Markov randomness was thoroughly investigated for temporal difference (TD) learning in the pioneering work by~\cite{tsitsiklis1997analysis}. However, 
\emph{finite-time} analyses for TD learning and linear SA were only recently provided  by~\cite{bhandari2018, srikant2019finite}. These contributions were followed by finite-time analyses of more general nonlinear SA algorithms, such as Q-learning~\citep{chen2023concentration, qu2020finite}. More recent work has delved into the non-asymptotic convergence of decentralized SA~\citep{zeng2022finite, doanDistributed}. 

$\bullet$ \textit{Delays and Asynchrony in Optimization.} The study of delays and asynchrony in distributed optimization dates back to the seminal work by~\cite{Tsitsiklis86}. While the focus of this work was on asymptotic convergence, the current focus in ML has shifted towards finite-sample guarantees. As such, motivated by the growing popularity of distributed ML, a significant body of work has explored the \emph{finite-time} effects of delays on the performance of stochastic optimization algorithms such as stochastic gradient descent (SGD), and variants thereof~\citep{feyzmahdavian2016asynchronous, gurbuzbalaban2017convergence, dutta2018slow, cohen2021asynchronous, koloskova2022sharper, glasgow2022asynchronous, fedbuff}. We note here that stochastic optimization is a special case of SA, where the noise samples that enter into the gradients are generated in an i.i.d. manner, i.e., they exhibit no temporal correlations. However, for applications centered around asynchronous multi-agent RL (MARL)~\citep{bouteiller2020reinforcement, chen2023efficient, mnih2016asynchronous}, one needs to contend with two key challenges simultaneously: (i) the randomness in the observations comes from a \emph{temporally correlated} Markov process, and (ii) the noisy SA operator is evaluated at potentially stale iterates. \textit{We are unaware of any work that provides a finite-time analysis of SA schemes that feature both the above challenges.}  

\subsection{Contributions}
In light of the above discussion, the main contributions of this work are summarized below. 

\begin{enumerate}
    \item \textbf{Novel Problem Formulation:} Our work provides the first comprehensive finite-time convergence analysis for delayed SA schemes under Markovian sampling. 

    \item \textbf{Tight Bound in the Constant Delay Case:} We start by analyzing the constant-delay setting in Section~\ref{sec:constantDelay}, and prove exponential convergence of the delayed SA update rule to a noise ball around the fixed point of the SA operator; see Theorem~\ref{thm1}. Our analysis reveals that the exponent of convergence gets scaled down by $\max\{\tau, \tau_{mix}\}$, where $\tau$ represents the constant delay, and $\tau_{mix}$ represents the mixing time of the underlying Markov process. Notably, our result (i) complies with existing finite-time bounds for non-delayed SA schemes~\citep{srikant2019finite}; and (ii) exhibits a \textit{tight} dependence on the delay $\tau$; see Section~\ref{sec:constantDelay}
 for more discussion on this aspect. 

\item \textbf{Tight Bound for  Time-Varying Delays:} One of the main contributions of this work is to generalize our result on constant delays to \textit{arbitrary time-varying} delays that are bounded. We provide such a result in Theorem~\ref{thm2} of Section~\ref{sec:vanillaDelayed}, where we show that the exponent of convergence gets scaled down this time by $\max\{\tau_{max}, \tau_{mix}\}$. Here, $\tau_{max}$ is the maximum possible delay. An interesting observation stemming from our analysis is that for slowly mixing Markov chains where $\tau_{mix} > \tau_{max}$, the effect of delays gets subsumed by the natural mixing of the underlying Markov process. \textit{As such, slowly mixing Markov chains tend to be more robust to delays.}  

 \item \textbf{Novel Proof Technique:} As we explain in detail in Section~\ref{sec:vanillaDelayed}, achieving a tight dependence on $\tau_{max}$ in Theorem~\ref{thm2} requires a significantly different proof technique than those existing. In particular, our proof neither employs generating function techniques as in~\citep{arjevani2020tight}, nor the popular error-feedback framework as in~\citep{stich2020error}. Instead, it relies on a new inductive argument to establish uniform boundedness of the iterates generated by the delayed SA scheme. This allows us to treat the joint effect of delays and Markovian sampling as a bounded perturbation, thereby considerably simplifying the subsequent analysis. The resulting approach is novel, and may be of independent interest to both RL and optimization. 

\item \textbf{Introduction and Analysis of a Delay-Adaptive Algorithm:} In Section~\ref{sec:delayAdaptive}, 
we propose and analyze an intuitive delay-adaptive SA scheme, where an update is made only when the staleness in the update direction falls below a carefully chosen threshold. In Theorem~\ref{thm:picky}, we show that the convergence rate of this scheme depends on the average delay $\tau_{avg}$, as opposed to the maximum delay $\tau_{max}$. Furthermore, the step-size for this adaptive rule \textit{requires no knowledge at all of the delay sequence.}  

\item \textbf{Applications of Our Results:} As we explain in Section~\ref{sec:applications}, our theoretical findings find applications in a variety of  settings, including the effects of delays in TD learning, Q-learning, and SGD under Markovian sampling. 
\end{enumerate}

\textit{Motivation and Scope.} One of the main practical motivations for the present study lies in multi-agent RL where asynchronous communication naturally leads to delays. That said, the scope of this paper is limited to the single-agent case since even this basic setting poses major technical challenges that remain completely unexplored. As such, by laying the foundations for this single-agent setting, our work opens up avenues for understanding and designing algorithms in more complex MARL environments as future work. We also note that this type of \emph{single-agent} configuration has been the key enabler for the fundamental understanding of finite-time rates for SGD with delayed updates (under i.i.d. sampling)~\citep{feyzmahdavian2014delayed, arjevani2020tight, stich2020error}.
\section{PROBLEM FORMULATION}
\label{sec:model}
The objective of general SA is to solve a root finding problem of the following form:
\begin{equation}\label{eq:SA_problem}
    \textrm{Find } \xs \in \mathbb{R}^m \textrm{ such that } \barg(\xs) = 0, \hspace{0.2cm} 
\end{equation}
where, for a given approximation parameter $\thet \in \mathbb{R}^m$, the deterministic function $\barg(\thet)$ is the expectation of a noisy operator $\mathbf{g}(\thet, \ok)$ taken over a distribution $\pi$, and $\{\ok\}$ denotes a stochastic \emph{observation process}, which is typically assumed to converge in distribution to $\pi$~\citep{borkar2009stochastic, meyn2023stability}. 
\subsection{SA under Markovian Sampling}
In this paper, we consider SA under Markovian sampling, i.e., the observations $\{\ok\}$ \emph{are temporally correlated and form  a Markov chain}. We define
\begin{equation}
    \bar{\mathbf{g}}(\thet) \triangleq \mathbb{E}_{o_t\sim \pi}[\bfg(\thet,o_t)],
\end{equation}
where $\pi$ is the \emph{stationary distribution} of the Markov chain $\{\ok\}$. SA consists in finding an approximate solution to~\eqref{eq:SA_problem} while having access only to sampled instances $\mathbf{g}(\thet, \ok)$ of $\barg(\thet)$. The typical iterative SA update rule with a constant step size $\alpha$ is as follows~\citep{srikant2019finite, chen2022finite},
\begin{equation}\label{eq:SA_update}
	\xkp = \xk + \alpha\mathbf{g}(\xk, \ok).
\end{equation}
The asymptotic convergence of SA update rules of the form in Eq.~\eqref{eq:SA_update} under Markovian sampling was first investigated by~\cite{tsitsiklis1997analysis}. On the other hand, 
\emph{finite-time} convergence results under Markovian sampling have only recently been established for linear~\citep{srikant2019finite, bhandari2018} and non-linear~\citep{chen2022finite} SA, requiring fairly involved analyses. Indeed, temporal correlations between the samples of $\{\ok\}$, which is also inherited by the iterate sequence $\{\xk\}$, prevents the use of techniques commonly adopted for the finite-time study of SA under i.i.d. sampling, triggering the need for a more elaborate analysis. 





\subsection{Exemplar Applications}
\label{sec:applications}
In this subsection, we provide some examples of the considered SA setting under Markovian sampling.  

\textbf{TD learning.}
TD learning with linear function approximation is an SA algorithm that can be used to learn an approximation of the value function $V_{\mu}(s)$ of a Markov decision process (MDP) associated to a given policy $\mu$. In this example, we denote by $s\in\mathcal{S}$ the MDP state from a finite $n$-dimensional state space~$\mathcal{S}$. The algorithm works by iteratively updating a linear function approximation parameter $\thet\in\mathbb{R}^m$ (with $m<n$) of the (approximated) value function using the following update rule:
\eqal{eq:TD}{
&\thet_{t+1} = \thet_t + \alpha (R_t + \gamma \hat{V}(s_{t+1}; \thet_t) - \hat{V}(s_t; \thet_t)) \boldsymbol{\phi}(s_t), \\
&\hat{V}(s; \thet) = \thet^\top \boldsymbol{\phi}(s)
,}
where $\alpha$ is the learning rate and $0<\gamma<1$ a discount factor. At iteration $t$, $\thet_t$ is the parameter vector of the linear function approximation, $s_t$ is the state of the MDP that is part of a single Markovian trajectory, $R_t$ is the received reward, $s_{t+1}$ is the next state, $\hat{V}(s_t; \thet_t)$ is the estimated value of the state $s_t$, and $\boldsymbol{\phi}(s)\in\mathbb{R}^m$ is a feature vector that maps the state $s$ to a vector in $\mathbb{R}^m$. The noise in the update rule is introduced by the random reward $R_t$, and the random next state $s_{t+1}$ whose distribution is dictated by the transition probabilities of the underlying Markov chain. Specifically, the update in~\eqref{eq:TD} is an instance of SA with  
\eqal{}{\bfg(\thet_t,o_t)=(R_t + \gamma \hat{V}(s_{t+1}; \thet_t) - \hat{V}(s_t; \thet_t)) \boldsymbol{\phi}(s_t),}
where $o_t = (R_t, s_t, s_{t+1})$ is the correlated sampling process inducing the randomness. For more formal details on TD learning with linear function approximation, see \cite{srikant2019finite, bhandari2018}.

\textbf{Q-learning.} Q-learning is a reinforcement learning algorithm that can be used to learn an optimal policy for a Markov decision process (MDP). In Q-learning, the goal is to learn a Q-function that maps state-action pairs to their corresponding expected rewards. When the state and action spaces are large, one typically resorts to a parameterized approximation of the Q-function for each state-action pair. The special case of linear function approximation takes the form: $\hat{Q}(s,a; \thet) = \thet^\top \boldsymbol{\phi}(s,a)$, where $\boldsymbol{\phi}(s,a)$ is a fixed feature vector for the state-action pair $(s,a) \in \mathcal{S} \times \mathcal{A}$, and $\mathcal{S}$ and $\mathcal{A}$ represent the (finite) state and action spaces, respectively. The Q-learning update rule with linear function approximation is another instance of SA that takes the following form:
\begin{equation}\label{eq:Qlearnig}
 \resizebox{1\hsize}{!}{$
\begin{aligned}
\thet_{t+1} &= \thet_t + \alpha g(\thet_t,o_t),\\
 g(\thet_t,o_t)& =(R_t + \gamma \max_{a'\in \mathcal{A}}  \hat{Q}(s_{t+1},a'; \thet_t)-\hat{Q}(s_{t+1},a_t; \thet_t)) \boldsymbol{\phi}(s_t,a_t).
\end{aligned}
$}
\nonumber
\end{equation}

The notation in the above rule is consistent with what we used to describe the TD update rule in~\eqref{eq:TD}. The information used in each time-step is encapsulated in $o_t = (R_t, s_t, a_t, s_{t+1})$. For more details, we refer the reader to~\citep{chen2022finite}. 

\textbf{SGD with Markovian noise.} Stochastic Gradient Descent (SGD) is a widely used optimization method for minimizing a function $F(\thet)=\mathbb{E}_{o\sim \pi}{[f(\thet,o)]}$ with respect to a parameter vector 
$\thet\in\mathbb{R}^m$ and a distribution $\pi$. SGD operates by updating the parameter vector iteratively in the direction of the function's noisy negative gradient. In this context, SGD can be regarded as an instance of SA with the update rule defined as in \eqref{eq:SA_update}, with $\bfg(\thet_t,o_t) = -\nabla f(\thet_t,o_t)$. In online, dynamic settings, the samples $\{o_t\}$ used in SGD can exhibit temporal correlations. In this respect, our results are applicable to the SGD framework in which the samples $\{o_t\}$ are correlated and form a Markov chain - a setting studied for example by~\cite{doan2022finite} and~\cite{even2023stochastic}.




\subsection{SA with delayed updates}
In many real-world applications, the SA operator $\mathbf{g}(\cdot)$ is only available when computed with delayed iterates and/or observations. The main \textbf{objective} of this paper is to provide a unified framework to analyse the finite-time convergence of SA under the joint effects of Markovian sampling and delayed updates. We proceed by formally introducing the setting. We consider the following stochastic recursion with delayed updates:
\begin{equation}\label{eq:delayedSA}
	{\xkp = \xk + \alpha\mathbf{g}(\xktau, \oktau), \ \ 0\leq\tau_t \leq t,}
\end{equation}
where $\alpha$ is a constant step size and $\tau_t$ is the delay with which the operator $\mathbf{g}(\cdot)$ is available to be used at iteration $t$. This specific update rule is motivated by many scenarios of practical interest. For instance, in distributed machine learning and reinforcement learning, it is often the case that the agents' updates are performed in an asynchronous manner~\citep{bouteiller2020reinforcement, chen2023efficient}, leading naturally to update rules of the form~\eqref{eq:delayedSA}. 

Update rules of the form~\eqref{eq:delayedSA} have been recently studied in the context of SA but with i.i.d. observations (see for example the works by~\cite{koloskova2022sharper} and \cite{fedbuff} for SGD updates with delays). However, to the best of our knowledge, nothing is known about the finite-time convergence behaviour of such update rules under Markovian observations.  
Compared with i.i.d. settings, the Markovian setting introduces major technical challenges, including dealing with (i) the use of a delayed operator $\mathbf{g}(\xktau, \oktau)$, and (ii) sequences of correlated observation samples $\{\ok\}$, which, in turn, induce temporal correlations in the iterates $\{\xk\}$. The interplay between such delays and temporal correlations requires a notably careful analysis, one which we provide as a main contribution of this paper. The key features and challenges of the analysis relative to previous works and other settings are provided with more details in sections~\ref{sec:constantDelay} and~\ref{sec:vanillaDelayed}.

\subsection{Assumptions and Definitions}
We proceed by describing a few assumptions needed for our analysis.  First, we make the following natural assumption on the underlying Markov chain $\{\ok\}$ ~\citep{bhandari2018, srikant2019finite, chen2022finite}.

\begin{assumption}
	The Markov chain $\{\ok\}$ is aperiodic and irreducible.
	\label{assump:mixing}
\end{assumption}

Next, we state three further assumptions that are common in the analysis of SA algorithms.

\begin{assumption}\label{ass:strongconvex}
Problem~\eqref{eq:SA_problem} admits a solution $\xs$, and $\exists \,\mu > 0$ such that for all $\thet\in {\mathbb{R}}^{m}$, we have
 \eqal{}{\langle\thet -\thet^*, \bar{\bfg}(\thet)-\bar{\bfg}(\thet^*)\rangle\leq -\mu \|\thet-\thet^*\|^2.}
\end{assumption}
\begin{assumption}\label{ass:Lipschitz}
	There exists $L >0$ such that for any $\thet_1, \thet_2\in\mathbb{R}^m$ and $o\in \{\ok\}$, we have
 \eqal{}{\|\bfg(\thet_1,o)-\bfg(\thet_2,o)\|\leq L\|\thet_1-\thet_2\|.
 }
 Furthermore, there exists $\sigma > 0$ such that for any $\thet\in\mathbb{R}^m$, we have 
 \eqal{}{\|\bfg(\thet,o)\|\leq L(\|\thet\| + \sigma).}
\end{assumption}
Assumption~\ref{ass:strongconvex} is a strong monotone property of the map $-\barg(\thet)$ that guarantees
that the iterates generated by a ``mean-path'' (steady-state) version of Eq.~\eqref{eq:SA_problem}, $\xkp = \xk + \alpha\barg(\xk)$, converge exponentially fast to $\xs$. Assumption~\ref{ass:Lipschitz} states that $\mathbf{g}(\thet, \ok)$ is globally uniformly (w.r.t. $\ok$) Lipschitz in the
parameter $\thet$.

Finally, we introduce an assumption on the time-varying delay sequence $\{\tau_t\}$.
\begin{assumption}\label{ass:delayBound}
    There exists an integer $\tau_{max} \geq 0$ such that $\tau_t \leq \tau_{max}, \hspace{0.2cm} \forall t\geq 0.$
\end{assumption}

\begin{remark}
Assumption~\ref{ass:strongconvex} holds for TD learning (Lemma 1 and Lemma 3 in~\cite{bhandari2018}) with linear function approximation, variants of Q-learning with linear function approximation~\citep{chen2022finite}, and for strongly convex functions in the context of optimization. Similarly, Assumption~\ref{ass:Lipschitz} holds for TD learning~\citep{bhandari2018} and Q-learning with linear function approximation~\citep{chen2022finite}, and is typically made in the analysis of SGD~\citep{doan2022finite}. 
\end{remark}
We now introduce the following notion of \emph{mixing time} $\tau_{mix}$, that plays a crucial role in our analysis, as in the analysis of all existing finite-time convergence studies of SA under Markovian sampling.

\begin{definition} \label{def:mix} 
	{Let $\tmix$ be such that, for every $\boldsymbol{\theta} \in \mathbb{R}^m$ and ${ o_{0}}$, we have} 
   \hspace{-10mm}
 \begin{equation}
    \begin{aligned}
 \Vert \mathbb{E}\left[\mathbf{g}(\boldsymbol{\theta}, \ok)|o_{0}\right]-\barg(\boldsymbol{\theta})\Vert \leq \alpha\left(\Vert \boldsymbol{\theta} \Vert +1 \right), {\forall t \geq \tmix}. 
    \end{aligned}
\label{eqn:mix}
 \end{equation}
  \hspace{-4mm}
\end{definition}
\begin{remark}
{
 Note that Assumption~\ref{assump:mixing} implies that the Markov chain $\{\ok\}$ {mixes at a geometric rate}. This, in turn, implies the existence of some $K \geq 1$ such that $\tmix$ in Definition~\ref{def:mix} satisfies $\tmix \leq K \log(\frac{1}{\alpha})$. In words, this means that for a fixed $\boldsymbol{\theta}$, if we want the noisy operator $\mathbf{g}(\thet, \ok)$ to be $\alpha$-close (relative to $\boldsymbol{\theta}$) to the expected operator $\barg(\thet)$, then the amount of time we need to wait for this to happen scales logarithmically in the precision $\alpha$}.
\end{remark}

\section{WARM UP: STOCHASTIC APPROXIMATION WITH CONSTANT DELAYS}
\label{sec:constantDelay}
\begin{table*}[t]
\vspace{-1mm}
 \caption{Summary of results. Here, we do not show problem specific constants other than $\tmix$ and $\tmax$.}
 \vspace{-1mm}
   \centering
 \begin{tabular}{c|cc}
   \hline
    Algorithm  &  Variance  Bound & Bias Bound\\
     \hline
        \hline
     Constant Delay~\eqref{eq:SA_constD} &  ${O}(\sigma^2)$& $O\left(\exp\left(\frac{-T}{\max\{\tau, \tau_{mix}\}}\right)\right)$\\
     \hline
    Time-Varying Delays~\eqref{eqdelaymain} &  ${O}(\sigma^2)$& $O\left(\exp\left(\frac{-T}{\max\{\tau_{mix},\tau_{max}\}}\right)\right)$\\\hline
    
    Time-Varying Delays\\
    Delay-Adaptive Update~\eqref{eq:algo}&${O}(\sigma^2)$& $O\left(\exp\left(\frac{-T}{\tau_{mix}.\tau_{avg}}\right)\right)$\\
         \hline
    
  \end{tabular}
  \vspace{-5mm}
  \label{tab:1}
\end{table*}
In this section, we present the first finite-time convergence analysis of a SA scheme with a constant delay under Markovian sampling. With respect to the SA scheme with delayed updates introduced in~\eqref{eq:delayedSA}, we fix $\tau_t = \tau$, where $\tau$ represents the constant delay. This leads to the following update rule:
\eqal{eq:SA_constD}{
&\textbf{\textit{SA with Constant Delay:} }\\\quad &\thet_{t+1} = \begin{cases}
\thet_0 &\textrm{if  }\hspace{0.1cm} 0\leq t < \tau ,\\
\thet_t +\alpha \bfg(\thet_{t-\tau},o_{t-\tau})  &\textrm{if }\hspace{0.1cm} t \geq \tau.
\end{cases}
}
We set $\thet_{t+1}$ to $\thet_0$ for the first $\tau$ time-steps to simplify the exposition of the analysis; such a choice is not necessary, and one can use other natural initial conditions here without affecting the final form of our result. We now state our first main result that provides a finite-time convergence bound for the update rule in~\eqref{eq:SA_constD}.
\begin{theorem} 
\label{thm1} Suppose Assumptions 1-3 hold. Let \(w_t \triangleq (1 - 0.5\alpha\mu)^{-(t+1)}\) and \(W_T = \sum_{t=0}^{T} w_t\). Let \(\thet_{out}\) be an iterate chosen randomly from \(\{\thet_t\}_{t=0}^{T}\), such that \(\thet_{out} = \thet_t\)  with probability \(\frac{w_t}{W_T}\). Define \(r_{out} \triangleq \|\thet_{out} - \xs\|\) and \(\bar{\tau} \triangleq \max\{\tau, \tmix\}\). There exists a universal constant \(C_1 \geq 2\), such that, for \(\alpha\leq \frac{\mu}{C_1L^2\bar{\tau}}\), the following holds for the rule in~\eqref{eq:SA_constD} $\forall T \geq 0$: 
\eqal{eq:thm1main}{
   {\E{r_{out}^2}} &\leq C_{\alpha}\exp\left(-0.5\alpha\mu T\right) + O(\sigma^2)\frac{\alpha L^2\bar\tau}{\mu},
}
with $C_{\alpha}= O\left(\frac{1}{\alpha\mu} + \frac{\bar{\tau}\sigma^2}{\mu}\right)$.
Setting \(\alpha = \frac{\mu}{C_1L^2\bar{\tau}}\), we get
\eqal{}{
   {\E{r_{out}^2}} &\leq C_{\bar{\tau}}\exp\left(-0.5\frac{\mu^2}{C_1L^2\bar\tau} T\right) + O(\sigma^2),
\label{eqn:Constbnd}
}
with $C_{\bar{\tau}} = O\left(\frac{2C_1L^2\bar{\tau}}{\mu^2} + \frac{\sigma^2\bar{\tau}}{\mu}\right)$.
\end{theorem}

\textbf{Main Takeaways:} 
We now outline the key takeaways from the above Theorem. From~\eqref{eqn:Constbnd}, we note that the delayed SA scheme in~\eqref{eq:SA_constD} guarantees exponentially fast convergence (in mean-squared sense) to a ball around the fixed point $\thet^*$, where the size of the ball is proportional to the noise-variance level $\sigma^2.$ For SA problems satisfying Assumptions 1-3, a convergence bound of this flavor is consistent with what is observed in the non-delayed setting as well under Markovian sampling, when one employs a constant step-size. See, for instance, the bounds for TD learning in~\citep{srikant2019finite, bhandari2018}, and for Q-learning in~\citep{chen2022finite}.

\textit{The Effect of Delays.} A close look at~\eqref{eqn:Constbnd} reveals that the exponent of convergence in the first term of~\eqref{eqn:Constbnd} scales inversely with $\bar\tau = \max\{\tau, \tau_{mix}\}$. Hence, if $\tau \geq \tmix$, the constant delay $\tau$ can slacken the rate of exponential convergence to the noise ball. Is such an effect of the delay $\tau$ inevitable, or just an artifact of our analysis? We elaborate on this point below. 

\textit{On the Tightness of our Bounds.} It turns out that the inverse scaling of the exponent with $\tau$ shows up even for SGD with constant delays under i.i.d. sampling - a special case of the general SA scheme we study here. Furthermore, this dependency has been shown to be tight for SGD in~\citep{arjevani2020tight}. As such, our rate in Theorem~\ref{thm1} is tight in its dependence on the delay $\tau$. We also note that the inverse scaling of the exponent with the mixing time $\tau_{mix}$ is consistent with prior work~\citep{srikant2019finite, bhandari2018}, and is, in fact, \textit{unavoidable}~\citep{nagaraj2020least}. To sum up, in Theorem~\ref{thm1} we provide the \textit{first finite-time convergence bound for a SA scheme with updates subject to constant delays under Markovian sampling}, and establish a convergence rate that has \emph{tight} dependencies on both the delay $\tau$ and the mixing time $\tmix$. 

\textit{Slowly-mixing Markov Chains are Robust to Delays.} An interesting conclusion from our analysis is that if the underlying Markov chain mixes slowly, i.e., if $\tau_{mix}$ is large, and in particular, if $\tau_{mix} > \tau$, then the convergence rate will be dictated by $\tau_{mix}$, since then $\bar{\tau}=\max\{\tau, \tau_{mix}\}=\tau_{mix}$. In other words, the effect of the delays will be dominated by the natural mixing effect of the underlying Markov process. This observation is novel to our setting and analysis, and absent in optimization under delays with i.i.d. sampling.

\textit{Comments on the Analysis.} Our proof for Theorem \ref{thm1} - provided in Appendix~\ref{sec:proof of thm1} - is partially inspired by the analysis in~\citep{stich2020error}. However, the key technical challenge relative to~\citep{stich2020error} arises from the need to simultaneously contend with delays and temporal correlations introduced by Markovian sampling; notably, the existing literature on optimization under delays~\citep{stich2020error, dutta2018slow,
koloskova2022sharper} \textit{does not} need to deal with the latter aspect. This dictates the need for new ingredients in the analysis that we outline in Appendix ~\ref{sec:proof of thm1}. However, as we explain in Appendix~\ref{sec:proof of thm2}, the technique we employ to prove Theorem \ref{thm1} falls apart in the face of time-varying delays. In the next section, we will explain in some detail how to overcome this challenge. 

\section{STOCHASTIC APPROXIMATION WITH TIME-VARYING DELAYS}
\label{sec:vanillaDelayed}
The goal of this section is to analyze~\eqref{eq:delayedSA} in its full generality by accounting for Markovian sampling and arbitrary time-varying delays that are only required to be bounded, as per Assumption~\ref{ass:delayBound}. In particular, we will study the following SA update rule:
\begin{equation}
\label{eqdelaymain}
\begin{aligned}   
&\textbf{\textit{SA with Time-Varying Delays:} }\\\,&\xkp = \xk + \alpha\mathbf{g}(\thet_{t-\tau_t}, o_{t-\tau_t}), \tau_t\leq \min \{t, \tau_{max}\}. 
\end{aligned}
\end{equation}

The main contribution of this paper - as stated below - is a finite-time convergence result for the above rule. 
\begin{theorem} 
\label{thm2} Suppose Assumptions 1-4 hold. Let $r_t \triangleq \|\thet_t - \xs\|$, and $\bar{\tau} \triangleq \max\{\tmix, \tau_{max}\}$. There exists a universal constant $C\geq 2$, such that, for $\alpha \leq \frac{\mu}{CL^2\bar{\tau}}$, the iterates generated by the update rule~\eqref{eqdelaymain} satisfy the following $\forall T\geq 3 {\bar{\tau}}$: 
\eqal{}{
\E{r_T^2}\leq & \left(\exp\left({-2\alpha\mu T}\right) + \frac{\alpha L^2(\tmix + \tau_{max})}{\mu}\right)O(\sigma^2).
}
 Setting $\alpha = \frac{\mu}{CL^2\bar{\tau}}$, we get
\eqal{eq:boundVanNoAlpha}{
\E{r_T^2}\leq \left( \exp\left({-\frac{2\mu^2T}{CL^2\bar{\tau}}}\right) +1\right) O(\sigma^2).
}

\end{theorem}

\textbf{Main Takeaways:} Comparing~\eqref{eq:boundVanNoAlpha} with~\eqref{eqn:Constbnd}, we note that our bound in Theorem~\ref{thm2} for time-varying delays mirrors that for the constant delay setting in Theorem~\ref{thm1}, with $\tau_{\max}$ appearing in the exponent in~\eqref{eq:boundVanNoAlpha} exactly in the same manner as $\tau$ shows up in~\eqref{eqn:Constbnd}. Thus, all the conclusions we drew in Section~\ref{sec:constantDelay} after stating Theorem~\ref{thm1} carry over to the setting we study here as well. In particular, since the constant-delay model is a special case of the arbitrary-delay model (with bounded delays), and we have already argued the tightness of our bound for the constant-delay setting, the dependencies of our bound (in~\eqref{eq:boundVanNoAlpha}) on the maximum delay $\tau_{max}$, and the mixing time $\tau_{mix}$, are optimal. At this stage, one might contemplate the need for studying the constant-delay case in Section~\ref{sec:constantDelay}, given that our results in this section clearly subsume the former. To explain our rationale, it is instructive to consider what is already known about delays in optimization under i.i.d. sampling. 

Some of the early works analyzing the effect of time-varying bounded delays on gradient descent and variants thereof, show that for smooth, strongly-convex objectives, the iterates generated by the delayed rule converge exponentially fast~\citep{assran2020advances, feyzmahdavian2016asynchronous, gurbuzbalaban2017convergence}. However, the exponent of convergence in these works is \textit{sub-optimal} in that it scales inversely with $\tau^2_{\max}$, as opposed to $\tau_{\max}$ in our bound in Theorem~\ref{thm2}. More recently, under a constant delay $\tau$, the authors in~\citep{arjevani2020tight} established the optimal dependence on $\tau$ in the exponent for quadratics; the same result was generalized to smooth, strongly-convex functions in \citep{stich2020error}. While our analysis for the constant-delay case in Theorem~\ref{thm1} borrows some ideas from that in~\citep{stich2020error}, such a technique appears to be insufficient for achieving the tight dependence on $\tau_{\max}$ in Theorem~\ref{thm2}. 

The proof of Theorem~\ref{thm2} departs from previous proof techniques for optimization under delays, avoiding the need for using generating functions as in~\citep{arjevani2020tight}, or the error-feedback framework with carefully crafted weighted averages of iterates as in~\citep{stich2020error}. The purpose of studying the constant-delay case first is precisely to highlight the critical points that make it difficult to adapt the aforementioned existing proof techniques to suit our needs - we provide such a discussion in Appendix~\ref{sec:proof of thm2}. Relative to the above works that \emph{only provide bounds for a constant delay under i.i.d. sampling}, our proof of Theorem~\ref{thm2} adopts a conceptually simpler route: it relies on a novel inductive argument to demonstrate the uniform boundedness of iterates. Surprisingly, this simpler approach - which we explain in some detail in the next section - provides a tight bound for the more challenging general case of time-varying delays under Markovian sampling. 

\subsection{Overview of Our Proof Technique}
In this section, we begin by outlining the main challenges that make it difficult to adapt existing optimization and RL proofs to our setting. We then explain the key novel technical ingredients that allow us to overcome such challenges. To proceed, let us define the error (at time $t$) introduced by the delay as follows: 
\begin{equation}
	\mathbf{e}_t \triangleq \mathbf{g}(\xk, o_t) - \mathbf{g}(\thet_{t-\tau_t}, o_{t-\tau_t}).
\end{equation}
Using the above in~\eqref{eqdelaymain} yields
\begin{equation}\label{eq:newExpressxkp}
	\xkp = \xk + \alpha\mathbf{g}(\xk, o_t) - \alpha \mathbf{e}_t.
\end{equation}
We examine $\|\xkp - \xs\|^2$ using~\eqref{eq:newExpressxkp}, which leads to
\begin{equation}\label{eq:analTauMax}
	\begin{aligned}
		\|\xkp - \xs\|^2 = J_{t,1} + \alpha^2J_{t,2} - 2\alpha J_{t,3}, \hspace{1mm} \textrm{with}
	\end{aligned}
\end{equation}
\begin{equation}\label{eq:3TsVanilla}
\begin{aligned}
	J_{t,1} &\triangleq \|\xk - \xs + \alpha\mathbf{g}(\xk, o_t)\|^2,\\
	J_{t,2} &\triangleq \|\mathbf{e}_t\|^2,\\
	J_{t,3} &\triangleq \langle \mathbf{e}_t, \xk - \xs\rangle +\alpha \langle\mathbf{e}_t, \mathbf{g}(\xk, o_t) \rangle.
\end{aligned}
\end{equation}
Note that the presence of $J_{t,2}$ and $J_{t,3}$ in~\eqref{eq:analTauMax} is a consequence of the delay, and does not show up in the analysis of vanilla non-delayed SA schemes. Our convergence analysis is built upon providing bounds in expectation on each of the three terms defined in~\eqref{eq:3TsVanilla}. We now elaborate on why this task is non-trivial. 

$\bullet$ \textbf{Challenge 1: Bounding the Drift.} Bounding $J_{t,1}$ in \eqref{eq:3TsVanilla} requires exploiting the geometric mixing property of the underlying Markov chain in~\eqref{eqn:mix}. To do so, the standard technique in SA schemes with Markovian sampling is to condition on the state of the system sufficiently into the past. This creates the need to bound a drift term of the form $\Vert \boldsymbol{\theta}_t - \boldsymbol{\theta}_{t-\tau_{mix}} \Vert $. In the absence of delays, there are two main approaches to handling this drift term. In~\citep{bhandari2018}, the authors are able to provide a uniform bound on this term by assuming a projection step. However, such a projection step may be computationally expensive, and requires knowledge of the radius of the ball that contains $\xs$. In~\citep{srikant2019finite}, the authors were able to bypass the need for a projection step via a finer analysis; in particular, Lemma 3 in~\citep{srikant2019finite} provides a bound of the following form: 
\begin{equation}\label{eq:srikantlemma}
    \|\xk - \boldsymbol{\theta}_{t-\tau_{mix}}\| \leq O(\alpha\tau_{mix})(\|\xk\| + \sigma), \forall t\geq \tau_{mix}. 
\end{equation}

The key feature of the above result is that it relates the drift to the \textit{current iterate} $\xk$. However, for the delayed SA scheme we study in~\eqref{eqdelaymain}, a bound of the form in~\eqref{eq:srikantlemma} no longer applies. The reason for this follows from the fact that the presence of delays in our setting causes the drift $\|\xk - \boldsymbol{\theta}_{t-\tau_{mix}}\|$ \textit{to be a function of not just the current iterate, but several other iterates from the past.} Thus, we need a new technical result that is able to explicitly relate the drift to all past iterates over a certain time window. We provide such a result in Lemma~\ref{th:thm2_l1}. 

$\bullet$ \textbf{Challenge 2: Handling Delay-induced Errors and Temporal Correlations.} Bounding $J_{t,3}$ in~\eqref{eq:3TsVanilla} requires handling the term $\langle \et, \xk - \xs \rangle$. This step 
is much more challenging compared to the i.i.d. sampling setting considered in the optimization literature with delays~\citep{zhou2018distributed, koloskova2022sharper, arjovsky,cohen2021asynchronous}. The difficulty here arises due to the statistical correlation among the terms $\boldsymbol{e}_t=\mathbf{g}(\xk, o_t) - \mathbf{g}(\thet_{t-\tau_t}, o_{t-\tau_t})$ and $\xk - \xs $.  In particular, there are two issues to contend with: (i) due to the correlated nature of the Markovian samples, $\E{\bfg(\xk, \ok)} \neq \barg(\xk)$, and (ii) the observation $o_{t-\tau_t}$ influences the iterate $\xk$. In view of the above issues, we depart from the standard routes employed in the optimization literature to handle similar delay-induced error terms, and employ a careful mixing time-argument instead to bound $\langle \et, \xk - \xs \rangle$. 

$\bullet$ \textbf{Challenge 3: Time-Varying Delays.} As we alluded to earlier in Section~\ref{sec:constantDelay}, the extension from the constant-delay setting to the time-varying setting appears to be quite non-trivial. In Appendix~\ref{sec:proof of thm2}, we consider a few potential candidate proof strategies for the setting with time-varying delays; these include: (i) a natural extension of the technique we use to prove Theorem~\ref{thm1}, and (ii) an adaptation of the proof in~\citep{feyzmahdavian2014delayed}. Unfortunately, while these strategies do end up providing rates, the exponent of convergence exhibits a \emph{sub-optimal} inverse scaling with $\tau^2_{max}$.

\subsection{Auxiliary Lemmas} The proof of Theorem~\ref{thm2} hinges on three new technical results. We outline them below. Our first result provides bounds in expectation on drift terms of the form $\|\thet_t - \xktaua\|^2$. To state this result, we define $\tau' \triangleq 2\tmax + \tmix$, and $
		{r}_{t,2}\triangleq \max_{t-\tau'\leq l \leq t}{\E{r_l^2}}.$ 
	
\begin{lemma}\label{th:thm2_l1} For any $t \geq \tmix$, we have
    \begin{alignat*}{5}
      (i)&\hspace{0.1cm}\E{\|\xk - \thet_{t-\tmix}\|^2}&&\leq 2\alpha^2\tmix^2 L^2(2r_{t,2} + 3\sigma^2).
\end{alignat*}
Similarly, for any $t\geq 0$ and $\tau_t \leq t$,
\begin{alignat*}{5}
      (ii)&\hspace{0.1cm}\E{\|\xk - \thet_{t-\tau_t}\|^2}&&\leq 2\alpha^2\tau_{max}^2 L^2(2r_{t,2} + 3\sigma^2).
\end{alignat*}
\end{lemma}

Unlike the analogous drift bounds in the RL literature~\citep{srikant2019finite}, Lemma~\ref{th:thm2_l1} bounds the drift as a function of the maximum iterate-suboptimality $r_{t,2}$ over a horizon whose length scales as $O(\tau_{mix}+\tau_{max}).$ Exploiting Lemma~\ref{th:thm2_l1}, we can provide bounds on $\E{J_{t,1}}$, $\E{J_{t,2}}$, and $\E{J_{t,3}}$ in terms of $r_{t,2}$. This is the subject of the next result. 

\begin{lemma}\label{th:thm2_l2}
	Let $t \geq \tau' = 2\tau_{max} + \tmix$. Then 
\begin{alignat*}{3}
(i)&\hspace{0.5cm}\E{J_{t,1}}&&\leq (1-2\alpha\mu)\E{r_t^2} + O(\alpha^2\tmix L^2 r_{t,2} )\\& &&\quad+ O(\alpha^2\tmix L^2\sigma^2).
\\
(ii)&\hspace{0.5cm}\E{J_{t,2}}&&\leq O(L^2(2r_{t, 2} + 3\sigma^2)).
\\
(iii)&\hspace{0.5cm}\E{-J_{t,3}}&&\leq O(\alpha L^2(\tmix + \tau_{max})(r_{t,2} + \sigma^2)).
\end{alignat*}
\end{lemma}
The most challenging part of proving the above lemma is part $(iii)$, where mixing time arguments need to be carefully applied to deal with temporal correlations. Plugging the above bounds in~\eqref{eq:analTauMax} yields:  
\eqal{eq:mainRecSec5}{\E{r^2_{t+1}}\leq (1-2\alpha\mu)\E{r^2_t}+ O(\alpha^2L^2 \bar{\tau}) (r_{t,2} + \sigma^2),}
where $\bar{\tau} \triangleq \max\{\tmix, \tau_{max}\}$. The above inequality suggests that the one-step progress can be captured by a contractive term, and an additive perturbation that jointly captures the effects of delays and Markovian sampling. The crucial step now is to handle this perturbation term, while achieving the optimal dependence on $\tau_{max}.$ Our key technical innovation here is to use a novel inductive argument to prove that the iterates generated by \eqref{eqdelaymain} remain uniformly bounded in expectation. We have the following result. 

\begin{lemma}\label{th:thm2_l3}
	Suppose $\alpha\leq \frac{\mu}{C\bar{\tau}L^2}$. There exists a universal constant $C \geq 1$ such that for all $t\geq 0$, it holds that $\E{r_t^2}\leq O(\sigma^2).$
\end{lemma}

The above result immediately implies that the perturbation term appearing in \eqref{eq:mainRecSec5} can be \textit{uniformly} bounded by $O(\alpha^2L^2 \bar{\tau} \sigma^2).$ From this point onward, the analysis is straightforward. It is worth emphasizing here that the \textit{idea of establishing a uniform bound on the perturbations induced by delays and Markovian sampling - without a projection step - is the main departure from all existing analyses.} 

\section{DELAY-ADAPTIVE STOCHASTIC APPROXIMATION}\label{sec:delayAdaptive}
 
In the previous section, we analyzed the vanilla delayed SA rule under time-varying delays. Notably, the resulting convergence rate was dependent on the maximum delay, $\tau_{max}$, and selecting the proper step size required a priori knowledge of $\tau_{max}$. In practice, the worst-case delay $\tau_{max}$ may be very large, leading to slow convergence of~\eqref{eqdelaymain}; furthermore, $\tau_{max}$ may be unknown.
Here, we introduce a new \emph{delay-adaptive} update rule whose convergence rate only depends on the \emph{average} delay, $\tau_{avg} = \frac{1}{T}\sum\limits_{t=1}^{T}\tau_{t}$, for the $T$ iterations over which the algorithm is executed. Moreover, the proposed \emph{delay-adaptive} algorithm does not require any knowledge of the delay sequence at all for tuning the step size. Our proposed rule is as follows. 
\begin{equation}
 \resizebox{1\hsize}{!}{$
\begin{aligned}\label{eq:algo}
&\textbf{\textit{Delay-adaptive SA:}}\\&\qquad\thet_{t+1}=
\begin{cases}
\thet_{t}+\alpha \mathbf{g}(\thet_{t-\tau_t},o_{t-\tau_t}) & \|\thet_{t}-\thet_{t-\tau_t}\|\leq {\epsilon}, \\
\thet_{t}  &\textrm{otherwise.}
\end{cases}  
\end{aligned}
   $}
\end{equation}
The rationale behind the above rule is quite simple: it makes an update only if the pseudo-gradient $\mathbf{g}(\thet_{t-\tau_t},o_{t-\tau_t})$ available at time $t$ is not too stale in the sense that it is evaluated at an iterate that is at most $\epsilon$ away from the current iterate $\thet_{t}$. Unlike the vanilla SA update rule, we can control the effect of delays on the convergence rate by carefully picking the threshold $\epsilon$. This is reflected in the next result.  

\begin{theorem} 
\label{thm:picky} Suppose Assumptions 1-4 hold. Let $r_t \triangleq \|\xk - \xs\|$. There exists a universal constant $C_2\geq 1$ such that for $\alpha\leq \frac{\mu}{C_2L^2\tmix}$, and $\epsilon=\alpha$, the iterates of~\eqref{eq:algo} satisfy the following for $T\geq \tmix$: 
\begin{equation}
 \resizebox{1\hsize}{!}{$
\begin{aligned}
	\E{r_T^2}\leq \left(\exp\left(\frac{-\alpha\mu T}{4(\tau_{avg}+1)}\right)+\frac{L^2\alpha\tau_{mix}}{\mu}\right) O(\sigma^2).
 \end{aligned}$}
 \nonumber
 \end{equation}
Additionally, if we set $\alpha=\frac{\mu}{C_2L^2\tmix}$, we obtain
\begin{equation}
 \resizebox{1\hsize}{!}{$
\begin{aligned}
	\E{r_T^2}\leq \left(\exp\left(\frac{-{\mu}^2T}{4C_2{L}^2\tau_{mix}(\tau_{avg}+1)}\right)+1\right)O(\sigma^2).
 \end{aligned}$} 
 \nonumber
 \end{equation}
\end{theorem}

\textbf{Discussion and Insights.} Relative to Theorem~\ref{thm2}, there are two main takeaways from the above result: (i) the choice of the step-size $\alpha$ requires no information about the delay sequence, and (ii) the exponent of convergence gets scaled down by $\tau_{avg}$, and not $\tau_{max}$. This result is significant because \textit{it is the first to provide a finite-time result for a delay-adaptive SA scheme under Markovian sampling}. 

We now provide some intuition as to what makes the above result possible. The vanilla SA rule in~\eqref{eqdelaymain} always makes an update, \textit{no matter how delayed the pseudo-gradients are}. As a result, to counter the effect of potentially large delays, one needs to necessarily use a \emph{conservative}  step-size $\alpha$ that scales inversely with $\tau_{max}$. This is precisely what ends up slackening the final convergence rate. In sharp contrast, our proposed delay adaptive SA rule in \eqref{eq:algo}, by design, rejects overly stale pseudo-gradients. This allows us to analyze \eqref{eq:algo} as a variant of the non-delayed SA rule in \eqref{eq:SA_update} with at most $O(\epsilon)$ error. To see this, observe that $\|\thet_{t}-\thet_{t-\tau_t}\|\leq {\epsilon}$ implies $\|\bfg(\thet_{t},o)-\bfg(\thet_{t-\tau_t},o)\|\leq L{\epsilon}$. Thus, every time we update, we move in the correct direction up to only a small amount of error. We formalize this key insight in Appendix~\ref{sec:proof of thm:picky}, where we provide the full proof of Theorem~\ref{thm:picky}. In Appendix~\ref{app:sim}, we also provide simulation results comparing non-delayed SA with the vanilla delayed SA update~\eqref{eq:delayedSA}, and with the delay-adaptive algorithm~\eqref{eq:algo}.
\section{CONCLUSIONS AND FUTURE WORK}
We studied the interplay between delays and Markovian sampling in the context of general stochastic approximation with a contractive operator. Our analysis allowed for arbitrary time-varying (potentially random) delays that are uniformly bounded. Leveraging a novel inductive proof technique, we provided the first non-asymptotic convergence result for this setting, obtaining a rate that has a \emph{tight} dependence on both the maximum delay $\tau_{max}$ and the mixing time of the underlying Markov chain, $\tau_{mix}$. Furthermore, we proposed the first delay-adaptive SA scheme that features two distinct advantages relative to a vanilla delayed SA protocol: (i) the rates for the former depend on the average delay $\tau_{avg}$ as opposed to the maximum delay $\tau_{max}$; and (ii) implementing the delay-adaptive rule requires no knowledge whatsoever of the delay sequence. The insights and novel analysis techniques from our work pave the way for various interesting future research avenues. We discuss some of them below.

\begin{enumerate}
\item The most natural next step would be to consider the study of asynchronous multi-agent RL algorithms where delays are inevitable. In this context, a recent line of work on federated RL has revealed the benefits of cooperation among the agents by establishing \emph{linear speedups} in the sample complexity of learning, despite Markovian sampling~\citep{khodadadian, dal2023federated, han, zhang2024}. Can we continue to expect such speedups in the presence of delays? What if we have stragglers in the system that slow down the pace of computation? Do the ideas~\citep{dutta2018slow, reisizadehstraggler} that apply in the context of distributed optimization/supervised learning to tackle stragglers carry over to the MARL setting? This remains an open direction worth exploring. 

\item At a high level, the findings in this paper contribute to a robustness theory for iterative RL algorithms. While we specifically focused on robustness to delays, we believe that the tools that were developed in the process should apply to the study of robustness to other types of structured perturbations. For instance, in a recent paper, \cite{mitra2023temporal} showed that TD learning protocols (with linear function approximation) can be just as robust to extreme compression/sparsification as their SGD counterparts, e.g., the \texttt{SignSGD} algorithm~\citep{bernstein2018signsgd}. We conjecture that the inductive proof used to study delays here can simplify and sharpen the results in \cite{mitra2023temporal}. We plan to verify this conjecture as part of future work.

\item Yet another avenue is to consider RL settings more general than the ones studied here. In particular, all our results relied on the strong monotone property in Assumption~\ref{ass:strongconvex}. Deriving tight rates under arbitrary delays in the absence of such an assumption will likely require some work. One could also seek to generalize our results to settings that involve nonlinear function approximation schemes (e.g., neural nets) as in~\cite{tian2023, cayci2023}, or to two-time-scale SA protocols that capture actor-critic methods. 

\item Finally, delays in physical systems often adhere to some structure. For instance, one can imagine such delays being generated randomly from some distribution. If so, if the delays are generated in an i.i.d. manner, one should be able to compute estimates of the mean and higher-order moments (if they exist) of the delay distribution with high probability, given enough samples. Can we exploit such information to design algorithms with better bounds than in this paper? Intuition dictates that this should be possible. However, to our knowledge, this remains a relatively unexplored direction.

\end{enumerate}

\newpage 
\section*{Acknowledgements}
This work was supported, in part, by the Italian Ministry of Education, University and Research through the PRIN Project under Grant 2017NS9FEY. It was also partially supported by NSF Award 1837253 and by the ARL grant DCIST CRA W911NF-17-2-0181. The research was also supported, in part, by NSF Grant ECCS-2335876 and NSF CAREER Grant 1943064.
\bibliographystyle{apalike}
\bibliography{main}


\newpage
\newpage

\section*{Checklist}

 \begin{enumerate}

 \item For all models and algorithms presented, check if you include:
 \begin{enumerate}
   \item A clear description of the mathematical setting, assumptions, algorithm, and/or model. Yes
   \item An analysis of the properties and complexity (time, space, sample size) of any algorithm. Yes
   \item (Optional) Anonymized source code, with specification of all dependencies, including external libraries. Not Applicable
 \end{enumerate}

 \item For any theoretical claim, check if you include:
 \begin{enumerate}
   \item Statements of the full set of assumptions of all theoretical results. Yes
   \item Complete proofs of all theoretical results. Yes
   \item Clear explanations of any assumptions. Yes
 \end{enumerate}

 \item For all figures and tables that present empirical results, check if you include:
 \begin{enumerate}
   \item The code, data, and instructions needed to reproduce the main experimental results (either in the supplemental material or as a URL). Not Applicable
   \item All the training details (e.g., data splits, hyperparameters, how they were chosen). Not Applicable
         \item A clear definition of the specific measure or statistics and error bars (e.g., with respect to the random seed after running experiments multiple times). Not Applicable
         \item A description of the computing infrastructure used. (e.g., type of GPUs, internal cluster, or cloud provider). Not Applicable
 \end{enumerate}

 \item If you are using existing assets (e.g., code, data, models) or curating/releasing new assets, check if you include:
 \begin{enumerate}
   \item Citations of the creator If your work uses existing assets. Not Applicable
   \item The license information of the assets, if applicable. Not Applicable
   \item New assets either in the supplemental material or as a URL, if applicable. Not Applicable
   \item Information about consent from data providers/curators. Not Applicable
   \item Discussion of sensible content if applicable, e.g., personally identifiable information or offensive content. Not Applicable
 \end{enumerate}

 \item If you used crowdsourcing or conducted research with human subjects, check if you include:
 \begin{enumerate}
   \item The full text of instructions given to participants and screenshots.Not Applicable
   \item Descriptions of potential participant risks, with links to Institutional Review Board (IRB) approvals if applicable. Not Applicable
   \item The estimated hourly wage paid to participants and the total amount spent on participant compensation. Not Applicable
 \end{enumerate}

 \end{enumerate}

\appendix
\onecolumn

\section*{SUPPLEMENTARY MATERIALS}

\section{Related Work}
\label{sec:related}

In this section, we review selected works related to the existing literature on delays in optimization, bandits, and reinforcement learning (RL).

\subsection{Delays in Optimization}

The study of delays and asynchrony in optimization has been a topic of interest since the seminal work by~\cite{bertsekas1989convergence}, which investigates convergence rates of asynchronous iterative algorithms in parallel or distributed computing systems. Subsequently, many researchers have explored the effects of delay and asynchrony on various learning and optimization methods. We summarize some of the significant works in this area below. 

\cite{agarwal2011distributed} focus on distributed delayed stochastic optimization, specifically gradient-based optimization algorithms that rely on delayed stochastic gradient information. They analyze the convergence of such algorithms and propose procedures to overcome communication bottlenecks and synchronization requirements. Their work demonstrates that delays are asymptotically negligible, achieving order-optimal convergence results for smooth stochastic problems in distributed optimization settings.

\cite{stich2020error} introduce an error-feedback framework, which examines stochastic gradient descent (SGD) with delayed updates on smooth quasi-convex and non-convex functions. They derive non-asymptotic convergence rates and show that the delay only linearly slows down the higher-order deterministic term, while the stochastic term remains unaffected. This result illustrates the robustness of SGD to delayed stochastic gradient updates, improving upon previous rates for different forms of delayed gradients. Notably, this work provides the best-known rate for SGD with i.i.d. noise. It is worth mentioning that most existing literature has focused on bounds depending only on the maximum delay. However, the recent works of~\cite{cohen2021asynchronous} and~\cite{koloskova2022sharper} have explored convergence rates that depend on the average delay sequence.

The aforementioned studies on delays in optimization contribute to understanding the impact of delays and asynchrony in various optimization algorithms. They provide insights into the convergence properties and shed light on the robustness of these methods to different forms of delay.  Nevertheless, there is still a gap in the literature regarding the finite-time convergence rates of delayed stochastic approximation schemes under Markovian sampling/noise.

\subsection{Delays in Bandits}

There has been significant research efforts on the impact of delays in bandits. Some of the key works in this area are summarized in the following.

Non-stochastic multi-armed bandits with unrestricted delays were studied in~\cite{thune2019nonstochastic}. The authors prove that the ``delayed" Exp3 algorithm achieves the $\mathcal{O}(\sqrt{(K T+D) \ln K})$ regret bound for variable but bounded delays. They also introduce a new algorithm that handles delays without prior knowledge of the total delay, achieving the same regret bound. The paper provides insights into the regret bounds for bandit problems with delays.

The challenges of stochastic linear bandits with delayed feedback, where the feedback is randomly delayed and delays are only partially observable, were addressed in~\cite{vernade2020linear}. The authors propose computationally efficient algorithms, OTFLinUCB and OTFLinTS, capable of integrating new information as it becomes available and handling permanently censored feedback. The authors prove optimal regret bounds for the proposed algorithms and validate their findings through experiments on simulated and real data.

Another paper investigates a variant of the stochastic $K$-armed bandit problem called ``bandits with delayed, aggregated anonymous feedback"~\citep{pike2018bandits}. In this setting, the player observes only the sum of a number of previously generated rewards that arrive in each round, and the information of which arm led to a particular reward is lost. The authors provide an algorithm that achieves the same worst-case regret as in the non-anonymous problem when the delays are bounded.

The above papers demonstrate that it is possible to design algorithms that can achieve good performance in the presence of delays in bandits. However, the performance metric of interest 
in these papers is typically cumulative regret. It is important to note here that the conclusions drawn for such a regret metric do not necessarily have implications for the sample-complexity bounds we care about in this work in the context of delayed stochastic approximation. 

\subsection{Delays in RL}

Until recently, the field of reinforcement learning had not thoroughly explored the impact of delays. In what follows, we highlight some key research works in this area.

\cite{bouteiller2020reinforcement} conducted a study on reinforcement learning with random delays, specifically focusing on environments with delays in actions and observations. They introduced the Delay-Correcting Actor-Critic (DCAC) algorithm, which incorporates off-policy multi-step value estimation to accommodate delays. Through theoretical analysis and practical experiments using a delay-augmented version of the MuJoCo continuous control benchmark, the authors demonstrated that DCAC outperforms other algorithms in delayed environments. In this work, however, no finite-time convergence analysis is provided for the analyzed algorithms.

\cite{mnih2016asynchronous} introduced asynchronous methods for deep reinforcement learning. They presented a lightweight framework that utilizes asynchronous gradient descent to optimize deep neural network controllers. The authors showed that parallel actor-learners have a stabilizing effect on training and achieve superior performance compared to state-of-the-art methods in domains such as Atari games and continuous motor control problems. Note that, however, although large measurement campaigns are conducted, no finite-time convergence analysis is performed in this work.

\cite{chen2023efficient} studied the problem of policy learning in environments with delayed or missing observations. They showed that it is possible to learn a near-optimal policy in this setting, even though the agent does not have access to the most recent state of the system. They established near-optimal regret bounds for this case. Note that this work focuses on regret analysis, while ours is focused on the impact of the joint effect of delays and Markovian sampling on the finite-time rate of convergence to the SA solution, which we study through the lenses of analyzing the interplay between maximum/average delay and mixing time on the convergence rates we provide.

\newpage

\section{Proofs of the Theorems}
In this Appendix, we provide the proofs for the theoretical results stated in the paper. We start by recalling some implications of the Assumptions of Section~\ref{sec:model} in the following.
\subsection*{Preliminaries}
First, recall that from Assumption~\ref{ass:strongconvex} we have, $\forall \boldsymbol{\theta} \in \mathbb{R}^d$: 
\begin{equation}\label{lemma:bhand}
			\begin{aligned}
				\langle \xs - \thet, \barg(\thet) \rangle \geq \mu\|\xs -\thet\|^2.
			\end{aligned}	
\end{equation}
Throughout the proof, we will often invoke the mixing property (see Definition \ref{def:mix}), which implies that, for a fixed $\thet$, the following is true:
\begin{equation}\label{eq:mixingEq}
\Vert \mathbb{E}\left[\mathbf{g}(\boldsymbol{\theta}, \ok)|o_{t-\tmix}\right]-\barg(\boldsymbol{\theta})\Vert \leq \alpha\left(\Vert \boldsymbol{\theta} \Vert +\sigma \right).
\end{equation}
We will also use the fact that the SA update directions and their steady-state versions are $L$-Lipschitz (Assumption \ref{ass:Lipschitz}), i.e., $\forall o \in \{o_t\}_{t\in \mathbb{N}}$, and $\forall \thet, \thet' \in \mathbb{R}^d$, we have: 
	\begin{equation}\label{eq:Lipschitz}
		\begin{aligned}
			&\|\bar{\mathbf{g}}(\thet) - \bar{\mathbf{g}}(\thet')\| \leq L\|\thet - \thet'\|, \hspace{1mm} \textrm{and} \\ 
			&\|\bfg(\thet, o_t) - \bfg(\thet', o_t)\| \leq L\|\thet - \thet'\|. 
		\end{aligned}
	\end{equation}
	We further have 
	\begin{equation}\label{eq:boundGradNorm}
		\|\mathbf{g}(\thet, o)\|\leq L(\|\thet\| + \sigma), \forall o \in \{o_t\}_{t\in \mathbb{N}},\forall \thet\in \mathbb{R}^d. 
	\end{equation}
	Given that $(x + y)^2 \leq 2(x^2 + y^2), \forall x, y \in \mathbb{R}$, we will often use the following inequality:
	\begin{equation}\label{eq:boundGsquared}
		\|\mathbf{g}(\thet, o_t)\|^2 \leq L^2(\|\thet\| + \sigma)^2 \leq 2L^2(\|\thet\|^2 + \sigma^2). 
	\end{equation}\vspace{0.2cm}
\noindent Without loss of generality, we assume that 
 \eqal{}{
 L \geq 1,\hspace{0.5cm}
 \sigma \geq \max\{\|\thet_0\|, \|\thet^*\|\},\hspace{0.5cm}\mu < 1.
 }
 We will often use the fact that, for any $x, y \in \mathbb{R}$, we have
 \begin{equation}\label{eq:halfSq}
     xy \leq \frac{1}{2}(x^2 + y^2).
\end{equation}
In addition, we will often use the fact that, for $t\geq 2$, $a_i \in \mathbb{R}, i = 0, ..., t-1$, it holds 
\begin{equation}\label{eq:sumSquared}
\left(\sum_{i = 0}^{t-1}a_i\right)^2 \leq t\sum_{i = 0}^{t-1}a_i^2
\end{equation} 
\subsection{Proof of Theorem \ref{thm1} and Related Lemmas}\label{sec:proof of thm1}

First, we recall the definition of the SA recursion with constant delay:

 \eqal{}{
\thet_{t+1} = \begin{cases}
\thet_0 &\textrm{if  }\hspace{0.1cm} 0\leq t < \tau \\
\thet_t +\alpha \bfg(\thet_{t-\tau},o_{t-\tau}).  &\textrm{if }\hspace{0.1cm} t \geq \tau
\end{cases}
}

For analysis purposes, we define a virtual iterate, $\tdxt$. This virtual iterate is updated with the SA update direction without delays, and it is defined as follows:
\eqal{eq:thTilde}{
\tilde\thet_{t+1} = 
\tilde\thet_t +\alpha \bfg(\thet_{t},o_{t}), \ \tilde{\thet}_0 = \thet_0. 
}
We also introduce the related error term $\e_t$, which is the gap between the virtual iterate and the actual iterate.
	\begin{alignat}{3}
		\tilde\thet_t &= \thet_t + \e_t, \hspace{0.25cm}\textrm{ with }\e_0 = \mathbf{0}.
	\end{alignat}
From the definition of $\tdxt$, we can write the following recursions for $\e_t$, for $t\geq 0$:
\eqal{}{\e_{t+1} &= 
\e_t + \alpha(\bfg(\thet_t,o_t) - \bfg(\thet_{t-\tau},o_{t-\tau}))  
.\label{eq:defRecursions}
}
We define $\bfg(\thet_l,o_l)=\thet_l=\e_l= \mathbf{0}$ for $l<0$. We also define $\tdrt = \|\tdxt - \xs\|$. For convenience, we define $\tdrt = 0$ for $t<0$.

\subsubsection{Auxiliary Lemmas}
Here, we present the main Lemmas needed to prove Theorem~\ref{thm1}. We start with three bounds on $\|\e_t\|$, $\|\e_t\|^2$ and $\sum_{t=0}^{T}w_t\|\e_t\|^2$, as follows.
\begin{lemma}\label{th:thm1_l1}
The  following three inequalities hold:

\begin{alignat}{3}
      (i)&\hspace{1cm}\|\e_t\|&&\leq \alpha\tau L\sigma + \alpha L\sum_{l = t-\tau}^{t-1}\|\thet_l\|,\label{eq:et}
      \\
      (ii)&\hspace{1cm}\|\e_t\|^2&&\leq 2\alpha^2\tau^2 L^2\sigma^2 + 2\alpha^2\tau L^2\sum_{l = t-\tau}^{t-1}\|\thet_l\|^2,\label{eq:etSq}
      \\
      (iii)&\hspace{0.4cm}\sum_{t=0}^{T}w_t\|\e_t\|^2&&\leq 4W_T\alpha^2\tau^2 L^2\sigma^2+ 16\alpha^2\tau^2 L^2\sum_{t = 0}^{T}{w_t}\|\tilde{\thet}_t\|^2,
\end{alignat}
where $(iii)$ holds for $\alpha\leq\frac{1}{4\tau L}$.
\end{lemma}
Part $(iii)$ of this Lemma is key to obtain the bound in~\eqref{eq:71}.
In the next Lemma, we provide bounds on the terms $\|\tilde{\thet}_t - \tilde{\thet}_{t-\tmix}\|$ and $\|\tilde{\thet}_t - \tilde{\thet}_{t-\tmix}\|^2$.
\begin{lemma}\label{th:thm1_l3}
For any $t\geq \tmix$, we have
\begin{alignat}{3}
  (i)&\hspace{1cm}\|\tilde\thet_{t-\tmix} - \tilde\xk\|&&\leq L\alpha\sigma\tau_{mix}+ L\alpha\sum_{l = t-\tmix}^{t-1}\|\thet_{l}\|.
  \\
  (ii)&\hspace{1cm}\|\tilde\thet_{t-\tmix} - \tilde\xk\|^2&&\leq 2L^2\alpha^2\tau_{mix}^2\sigma^2 + 2L^2\alpha^2\tau_{mix}\sum_{l = t-\tmix}^{t-1}\|\thet_{l}\|^2.\label{eq:gapSq}
\end{alignat}
\end{lemma}
Note that this Lemma is a variation of Lemma 3 in~\cite{srikant2019finite}, which is key to invoke mixing time arguments to get finite-time convergence bounds in existing non-delayed SA analysis. Let us define 
\eqal{}{
&n_t \triangleq \|\bfg(\thet_t, o_t)\|^2,\\
 &m_t  \triangleq \langle \bfg(\thet_{t},o_{t})-\bfg(\tilde\thet_{t},o_{t}), \tilde\thet_t - \thet^\star\rangle, \\
 &h_t \triangleq \langle \tdxt - \xs, \bfg(\tdxt, o_t) - \bar{\bfg}(\tdxt)\rangle.
}
To obtain a bound in the form~\eqref{eq: mainbound}, we need to bound $\E{h_t}$ properly, for which, in turn, we need Lemma~\ref{th:thm1_l3}. Furthermore, note that, in contrast to~\cite{srikant2019finite}, the bound is obtained for the sequence of \emph{virtual iterates}. In the next lemma, we provide bounds for the three key terms of the bound in~\eqref{eq:mainboundthm1}, i.e.,  $\|\bfg(\xk, o_t)\|^2$, $m_t$, and $\E{h_t}$.
\begin{lemma}\label{th:thm1_l4}
For all $t\geq 0$, we have
\begin{alignat}{3}
(i)&\hspace{0.4cm} n_t&&\leq4L^2\|\e_t\|^2 + 8L^2\tdrt^2 + 10L^2\sigma^2,
\\
(ii)&\hspace{0.4cm}m_t&&\leq 6\alpha\tau L^2\sigma^2 + 3\alpha\tau L^2\tdrt^2 + 2\alpha L^2\sum_{l = t-\tau}^{t-1}\left(\|\e_l\|^2+2\tilde{r}_l^2\right),
\\
(iii)&\hspace{0.3cm}\E{h_t}&&\leq \begin{cases}111\sigma^2L^2, \text{  for } 0\leq t\leq \tmix
\\
4\alpha\tmix L^2\left(8\sigma^2 + 3\E{\tdrt^2}\right)
+ 8\alpha L^2\sum_{l = t-\tmix}^{t-1}\E{\|\e_l\|^2 + 2\tilde{r}_l^2}, \text{  for } t\geq \tmix
\end{cases}
\end{alignat}
where $(iii)$ holds for $\alpha \leq \frac{1}{36L^2\tmix}$.
\end{lemma}
The proof of this last Lemma relies on the bound on $\|\e_t\|$ established in Lemma~\ref{th:thm1_l1}. The proof of $(iii)$ relies on the mixing properties of the Markov chain $\{o_t\}$ and on the bounds on $\|\thet_t - \thet_{t-\tmix}\|$ and $\|\thet_t - \thet_{t-\tmix}\|^2$ established in Lemma~\ref{th:thm1_l3}. Part $(iii)$ is the key and most challenging part of the proof, which allows us to get to the bound in~\eqref{eq: mainbound}. Using this last Lemma, in combination with Lemma~\ref{th:thm1_l1}, we are able to get the bound in~\eqref{eq:thm1main}. The conclusion of the proof is enabled by using $\E{ r_{t}^2}\leq 2\E{ \tilde r_{t}^2}+2 \E{  \|\e_{t}\|^2}$ and some further manipulations.

\subsubsection{Proofs of Auxiliary Lemmas}
We first state and prove the following lemma, which we will use later in the proof of Theorem~\ref{thm1}.
\noindent
\begin{lemma}\label{lem:weights}
    For $w_t \triangleq (1 - 0.5\mu\alpha)^{-(t+1)}$ with $\alpha\leq \frac{\mu}{C\bar\tau}$, $C\geq 2$, the following inequality holds for $0\leq i \leq 2\bar\tau$, and for any $t$,
    \eqal{}{w_t\leq 2 w_{t-i}.}
\begin{proof}

\eqal{eq:slowing}{
w_t &= w_{t-i} \left(1 - \frac{\mu\alpha}{2} \right)^{-i}
\\&
\overset{(a)}\leq w_{t-i} \left(1 - \frac{\mu^2}{2C\bar\tau} \right)^{-i} 
\\&
\overset{(b)}\leq w_{t-i} \left(1 - \frac{\mu^2}{2C\bar\tau} \right)^{-\bar\tau}
\\&
\overset{(c)}\leq w_{t-i} \left(1 - \frac{1}{4\bar\tau} \right)^{-\bar\tau} 
\\&
\overset{(d)}\leq w_{t-i} \left(1 +\frac{1}{2\bar\tau} \right)^{\bar\tau}  
\\&
\overset{(e)}\leq w_{t-i} \exp\left(\frac{1}{2}\right)\\&\leq 2w_{t-i}.
}
In $(a)$, we used the bound on $\alpha$; in $(b)$, we used the bound on $i$; in $(c)$, we used $\mu<1$ and $C\geq 2$; in $(d)$, we used
\eqal{eq:ineqInv}{\left(1 - x\right)^{-1}\leq \left(1 + 2x \right) \,\text{for}\quad 0\leq x\leq \frac{1}{2},}
and for $(e)$, we used $(1+x)^k\leq \exp(xk)$ for $k\geq 0$.
\end{proof}
\end{lemma}
Note that we defined  $\bfg(\thet_i,o_i)=\mathbf{0}$ for $i<0 $, $\thet_t= \mathbf{0}$ for $t< 0$, and $\e_t=\mathbf{0}$ for $t< 0$. First, note that, starting from the definition of $\e_t$ in~\eqref{eq:defRecursions}, 
\begin{equation}\label{eq:etUnfold}
\begin{alignedat}{3}
\e_{t+1} &= \e_t + &&\alpha\left(\bfg(\xk, o_t) - \bfg(\thet_{t-\tau}, o_{t-\tau}\right)\\
& = \e_{t-1} &&+ \alpha\left(\bfg(\thet_{t-1}, o_{t-1}) - \bfg(\thet_{t-1-\tau}, o_{t-1-\tau})\right)  \\& &&+\alpha\left(\bfg(\xk, o_t) - \bfg(\thet_{t-\tau}, o_{t-\tau})\right)
\\& = \e_{0} &&+ \alpha\sum_{l = 0}^{t}\left(\bfg(\thet_l, o_l) - \bfg(\thet_{l-\tau}, o_{l-\tau})\right)
\\& \overset{(*)}=\mathbf{0} &&+ \alpha\sum_{l = t-\tau+1}^t\bfg(\thet_l, o_l) ,
\end{alignedat}
\end{equation}
where $(*)$ follows because the overlapping terms in the sum cancel out. So, we obtain, for all $t\geq 0$,
\eqal{eq:etdefExt}{
\e_t &= \alpha\sum_{l = t-\tau}^{t-1}\bfg(\thet_l, o_l). }
We can now prove Lemma~\ref{th:thm1_l1}, which is key to proving  Theorem~\ref{thm1}.\vspace{0.3cm}

\textbf{Proof of Lemma~\ref{th:thm1_l1} - (i), (ii).}
From~\eqref{eq:etdefExt}, using the triangle inequality and the bound on the update direction~\eqref{eq:boundGradNorm}, we get, recalling that $\sigma\geq \|\thet_0\|$,
\eqal{}{
\|\e_t\| &= \| \alpha\sum_{l = t-\tau}^{t-1}\bfg(\thet_l, o_l)\|
\\& 
\overset{\eqref{eq:boundGradNorm}}\leq  \alpha L\sum_{l = t-\tau}^{t-1}(\|\thet_l\| + \sigma)
\\& 
\leq \alpha\tau L\sigma + \alpha L\sum_{l = t-\tau}^{t-1}\|\thet_l\|,
}
which proves (i). We now prove (ii). Using the triangle inequality and~\eqref{eq:sumSquared},
\eqal{}{
\|\e_t\|^2 &= \| \alpha\sum_{l = t-\tau}^{t-1}\bfg(\thet_l, o_l)\|^2
\\& \overset{\eqref{eq:sumSquared}}\leq  \alpha^2\tau\sum_{l = t-\tau}^{t-1}\|\bfg(\thet_l, o_l)\|^2.
}
Now, using the upper bound on the squared gradient norm~\eqref{eq:boundGsquared}, 
\eqal{}{
\|\e_t\|^2&\leq  \alpha^2\tau\sum_{l = t-\tau}^{t-1}\|\bfg(\thet_l, o_l)\|^2
\\& \leq 2\alpha^2\tau L^2\sum_{l = t-\tau}^{t-1}(\|\thet_l\|^2 + \sigma^2)
\\&\leq 2\alpha^2\tau^2 L^2\sigma^2 + 2\alpha^2\tau L^2\sum_{l = t-\tau}^{t-1}\|\thet_l\|^2,
}
which concludes the proof.\hspace*{\fill}~$\square$\vspace{0.3cm}

\noindent Using the above inequalities, we can now prove part $(iii)$ of  Lemma~\ref{th:thm1_l1}.\vspace{0.3cm}

\textbf{Proof of Lemma~\ref{th:thm1_l1} - (iii).} First, recall that, from Lemma~\ref{th:thm1_l1}, we have 
\begin{equation}\label{eq:recall3}
\begin{aligned}
    &\|\e_t\|^2\leq 2\alpha^2\tau^2 L^2\sigma^2 + 2\alpha^2\tau L^2\sum_{l = t-\tau}^{t-1}\|\thet_l\|^2.
\end{aligned}
\end{equation}
Based on Lemma~\ref{lem:weights}, for $0\leq i\leq 2\bar\tau$, we have $w_t\leq 2w_{t-i}$ (see~\eqref{eq:slowing}). Using~\eqref{eq:recall3},
\eqal{}{
\sum_{t=0}^{T}w_t\|\e_t\|^2&\leq \sum_{t=0}^{T}w_t\left(2\alpha^2\tau^2 L^2\sigma^2 + 2\alpha^2\tau L^2\sum_{l = t-\tau}^{t-1}\|\thet_l\|^2\right)
\\&
\leq 2W_T\alpha^2\tau^2 L^2\sigma^2 +2\alpha^2\tau L^2 \sum_{t =0}^{T}{w_t}\sum_{l = t-\tau}^{t-1}\|\thet_l\|^2
\\&
\overset{(*)}{\leq} 2W_T\alpha^2\tau^2 L^2\sigma^2 + 4\alpha^2\tau L^2\sum_{t = 0}^{T}\sum_{l = t-\tau}^{t-1}w_l\|\thet_l\|^2
\\&
\overset{(**)}{\leq} 2W_T\alpha^2\tau^2 L^2\sigma^2+ 4\alpha^2\tau^2 L^2\sum_{t = 0}^{T}w_t\|\thet_t\|^2
\\&
\leq 2W_T\alpha^2\tau^2 L^2\sigma^2+ 8\alpha^2\tau^2 L^2\sum_{t = 0}^{T}{w_t}\left(\|\tilde{\thet}_t\|^2 +\|\e_t\|^2\right)
\\&
\leq 2W_T\alpha^2\tau^2 L^2\sigma^2+ 8\alpha^2\tau^2 L^2\sum_{t = 0}^{T}{w_t}\|\tilde{\thet}_t\|^2 +\frac{1}{2}\sum_{t = 0}^{T}w_t\|\e_t\|^2,
}
where for $(*)$ we used the fact that $w_t \leq 2w_l$ for $t-2\bar\tau\leq l\leq t-1$, and for $(**)$ we used the fact that each element $w_l\|\thet_l\|^2$ appears at most $\tau$ times in the sum, for $l = 0, ..., T-1$ (note that, by definition, $\thet_l = 0$ for $l<0$). In the last inequality, we used $\alpha\leq \frac{1}{4\tau L}$. We can conclude getting
\begin{equation}\label{eq:interm2}
    \sum_{t=0}^{T}w_t\|\e_t\|^2\leq 4W_T\alpha^2\tau^2 L^2\sigma^2+ 16\alpha^2\tau^2 L^2\sum_{t = 0}^{T}{w_t}\|\tilde{\thet}_t\|^2.
\end{equation}\hspace*{\fill}~$\square$\vspace{0.3cm}

We now prove Lemma~\ref{th:thm1_l3}, that provides a bound on the norm of the gap $\|\tilde\thet_{t-\tmix} - \tilde\xk\|$ and its squared version $\|\tilde\thet_{t-\tmix} - \tilde\xk\|^2$.\vspace{0.3cm}

\textbf{Proof of Lemma~\ref{th:thm1_l3}.}
Inequality (i) of the Lemma can be easily proved by applying the definition of the recursion~\eqref{eq:thTilde}:
\eqal{eq:tildeTaumix}{\|\tilde\thet_{t-\tmix} - \tilde\xk\| &\leq \sum_{l = t-\tmix}^{t-1}\|\tilde\thet_{l+1} - \tilde\thet_{l}\|
\\&\leq \alpha\sum_{l = t-\tmix}^{t-1}\|\mathbf{g}(\thet_{l}, o_{l})\|
\\&\leq  L\alpha\sum_{l = t-\tmix}^{t-1}(\|\thet_{l}\| + \sigma) \\&= L\alpha\sigma\tau_{mix}+ L\alpha\sum_{l = t-\tmix}^{t-1}\|\thet_{l}\|.
}
Similarly, for inequality (ii), note that, squaring equation~\eqref{eq:tildeTaumix},
\eqal{}{\|\tilde\thet_{t-\tmix} - \tilde\xk\|^2& \leq 2L^2\alpha^2\tau_{mix}^2\sigma^2 + 2L^2\alpha^2\tau_{mix}\sum_{l = t-\tmix}^{t-1}\|\thet_{l}\|^2
.}\hspace*{\fill}~$\square$\vspace{0.3cm}

We now prove Lemma~\ref{th:thm1_l4}, which provide bounds for $\|\bfg(\xk, o_t)\|^2$, $m_t$, and $\E{h_t}$.\vspace{0.3cm}

\textbf{Proof of Lemma~\ref{th:thm1_l4} - (i).}
From~\eqref{eq:boundGsquared}, we have $\|\bfg(\thet_t, o_t)\|^2\leq 2L^2(\|\thet_t\|^2 + \sigma^2)$, and so
\eqal{}{
n_t = \|\bfg(\thet_t, o_t)\|^2&\leq 2L^2(\|\thet_t\|^2 + \sigma^2)
\\&
\leq 2L^2\|\thet_t - \tdxt + \tdxt\|^2 + 2L^2\sigma^2
\\&
\leq 4L^2\|\e_t\|^2 + 4L^2\|\tdxt\|^2 + 2L^2\sigma^2
\\&
\leq 4L^2\|\e_t\|^2 + 4L^2\|\tdxt- \xs + \xs\|^2 + 2L^2\sigma^2
\\&
\leq 4L^2\|\e_t\|^2 + 8L^2\tdrt^2 + 8L^2\|\xs\|^2 + 2L^2\sigma^2
\\&
\leq 4L^2\|\e_t\|^2 + 8L^2\tdrt^2 + 10L^2\sigma^2,
}
where we used $\|\thet^*\|\leq \sigma$. This concludes the proof. \hspace*{\fill}~$\square$\vspace{0.3cm}\vspace{0.3cm}

\textbf{Proof of Lemma~\ref{th:thm1_l4} - (ii).}
By the Cauchy-Schwarz inequality, Lipschitz continuity of $\bfg(\thet, o_t)$ in $\thet$ (see~\eqref{eq:Lipschitz}), and from the definition of $\e_t$, we get
\eqal{}{
 m_t &= \langle \bfg(\thet_{t},o_{t})-\bfg(\tilde\thet_{t},o_{t}), \tilde\thet_t - \thet^\star\rangle 
\\&\leq  \|\bfg(\thet_{t},o_{t})-\bfg(\tilde\thet_{t},o_{t})\|\|\tdxt - \xs\|
\\&
\leq L\|\tdxt - \xk\|\|\tdxt - \xs\|
\\&= L\|\e_t\|\tdrt.
}
Applying Lemma~\ref{th:thm1_l1} to bound $\|\e_t\|$, we get
\eqal{}{m_t&\leq L\left(\alpha\tau L\sigma + \alpha L\sum_{l = t-\tau}^{t-1}\|\thet_l\|\right)\tdrt 
\\&
= \alpha\tau L^2\sigma\tdrt + \alpha L^2\sum_{l = t-\tau}^{t-1}\|\thet_l\|\tdrt
\\&
\overset{\eqref{eq:halfSq}}\leq 2\alpha\tau L^2\sigma^2 + 2\alpha\tau L^2\tdrt^2 + \alpha L^2\sum_{l = t-\tau}^{t-1}\left(\|\thet_l\|^2 + \tdrt^2\right)
\\&
= 2\alpha\tau L^2\sigma^2 + 3\alpha\tau L^2\tdrt^2 + \alpha L^2\sum_{l = t-\tau}^{t-1}\|\thet_l\|^2
\\&
\overset{\eqref{eq:sumSquared}}\leq 2\alpha\tau L^2\sigma^2 + 3\alpha\tau L^2\tdrt^2 + 2\alpha L^2\sum_{l = t-\tau}^{t-1}\left(\|\e_l\|^2 + \|\tilde{\thet}_l\|^2\right)
\\&
\leq 6\alpha\tau L^2\sigma^2 + 3\alpha\tau L^2\tdrt^2 + 2\alpha L^2\sum_{l = t-\tau}^{t-1}\|\e_l\|^2 + 4\alpha L^2\sum_{l = t-\tau}^{t-1}\tilde{r}_l^2
.}
\hspace*{\fill}~$\square$\vspace{0.3cm}

Next, we provide the proof of Lemma~\ref{th:thm1_l4}, which, in turn,  provides a bound for $\E{h_t}$ - the term related to Markovian sampling whose analysis requires special care and mixing time arguments.\vspace{0.3cm}

\textbf{Proof of Lemma~\ref{th:thm1_l4} - (iii).}	We start with the case $0\leq t\leq \tmix$. Note that, using~\eqref{eq:boundGsquared}, 
\eqal{}{
h_t &= \langle \tdxt - \xs, \bfg(\tdxt, o_t) - \bar{\bfg}(\tdxt)\rangle
\\&
\leq \|\tdxt - \xs\|\|\bfg(\tdxt, o_t) - \bar{\bfg}(\tdxt)\|
\\&
\overset{\eqref{eq:halfSq}}\leq \frac{1}{2}\tdrt^2 + \frac{1}{2}\|\bfg(\tdxt, o_t) - \bar{\bfg}(\tdxt)\|^2
\\&
\overset{\eqref{eq:sumSquared}}\leq \frac{1}{2}\tdrt^2 + \|\bfg(\tdxt, o_t)\|^2 + \|\bar{\bfg}(\tdxt)\|^2
\\&
\overset{\eqref{eq:boundGsquared}}{\leq} \frac{\tdrt^2}{2} + 2L^2\|\tdxt\|^2 + 2L^2\sigma^2 + 2L^2\|\tdxt\|^2 + 2L^2\sigma^2
\\&
\leq \frac{\tdrt^2}{2} + 8L^2\tdrt^2 + 12L^2\sigma^2
\\&
\leq 9L^2\tdrt^2 + 12L^2\sigma^2.
}
Recall that 
\eqal{}{
\thet_{t+1} = \begin{cases}
\thet_0 &\textrm{if  }\hspace{0.1cm} 0\leq t < \tau \\
\thet_t +\alpha \bfg(\thet_{t-\tau},o_{t-\tau})  &\textrm{if }\hspace{0.1cm} t \geq \tau
\end{cases},
}
from which we can write, for $t\geq \tau$,
\eqal{eq:initial}{
r_{t+1}^2 &= r_t^2 + 2\alpha\langle \thet_t - \xs, \bfg(\thet_{t-\tau}, o_{t-\tau})\rangle + \alpha^2\|\bfg(\thet_{t-\tau}, o_{t-\tau})\|^2
\\&
\overset{\eqref{eq:halfSq}}\leq r_t^2 + \alpha r_t^2 + \alpha \|\bfg(\thet_{t-\tau}, o_{t-\tau})\|^2 + \alpha^2\|\bfg(\thet_{t-\tau}, o_{t-\tau})\|^2
\\&
\overset{\alpha<1}\leq (1+\alpha)r_t^2 + 2\alpha \|\bfg(\thet_{t-\tau}, o_{t-\tau})\|^2,
}
and note that, for $t<\tau$, $r_{t+1}^2 = r_t^2$, and hence~\eqref{eq:initial} holds true for all $t\geq 0$. Now note that 
\eqal{}{
\|\bfg(\thet_{t-\tau}, o_{t-\tau})\|^2 \leq 2L^2(\|\thet_{t-\tau}\|^2 + \sigma^2)\leq 4L^2r_{t-\tau}^2 + 6L^2\sigma^2.
}
Therefore, we can write 
\eqal{}{
r_{t+1}^2 \leq (1+\alpha)r_t^2 + 8\alpha L^2r_{t-\tau}^2 + 12\alpha L^2\sigma^2.
}
Now, we show that, for $k < \tmix$, 
\eqal{eq:prop}{
r_k^2 \leq \rho^kr_0^2 + \epsilon_k,
}
with $\epsilon_k = \rho\epsilon_{k-1} + \beta$, $\epsilon_0 = 0$, where $\rho = 1+\alpha + 8\alpha L^2 > 1$, and $\beta = 12\alpha L^2\sigma^2$. We show it by induction. The base case $k = 0$ is trivially true. Now suppose that inequality~\eqref{eq:prop} is true up to some $k\geq 0$, thus
\eqal{}{
r_s^2 \leq \rho^sr_0^2 + \epsilon_s, \ \ \forall s \leq k.
}
We can get, noting that, for all $k$, $0\leq \epsilon_k \leq \epsilon_{k+1}$,
\eqal{}{
r_{k+1}^2 &\leq (1+\alpha)r_k^2 + 8\alpha L^2r_{k-\tau} + 12\alpha L^2\sigma^2
\\&
\leq (1+\alpha)(\rho^kr_0^2 
 + \epsilon_k) + 8\alpha L^2(\rho^kr_0^2 
 + \epsilon_k) + 12\alpha L^2\sigma^2
 \\&
= (1+\alpha + 8\alpha L^2)\rho^kr_0^2 + (1+\alpha + 8\alpha L^2)\epsilon_k + 12\alpha L^2\sigma^2
\\&
= \rho^{k+1}r_0^2 + \rho\epsilon_k + \beta
\\&
= \rho^{k+1}r_0^2 + \epsilon_{k+1},
}
which concludes the induction proof of~\eqref{eq:prop}. Now note that, given that $L \geq 1$, $\rho \leq 1+9\alpha L^2$, and, for $\alpha \leq \frac{1}{36L^2\tmix}$, 
\eqal{}{
\rho^k \leq (1+9\alpha L^2)^k \leq (1+9\alpha L^2)^\tmix\leq  e^{9\alpha L^2\tmix}\leq e^{0.25}\leq 2.
}
Also note that, for all $k\leq \tmix$,
\eqal{}{
\epsilon_k = \beta\sum_{j = 0}^{k-1}(1+9\alpha L^2)^j\leq \beta\sum_{j = 0}^{\tmix-1}(1+9\alpha L^2)^\tmix\leq 2\beta\tmix,
}
and we can get, for all $k\leq \tmix$, noting that $r_0^2\leq 4\sigma^2$,
\eqal{eq:established}{
r_k^2 \leq 2r_0^2 + 2\beta\tmix = 2r_0^2 + 24\alpha L^2\sigma^2\tmix\leq 9\sigma^2.
}
Now note that, similarly to the calculations performed above, for $t<\tmix$,
\eqal{}{
\tdrp^2 &= \tdrt^2 + 2\alpha\langle \thet_t - \xs, \bfg(\thet_{t}, o_{t})\rangle + \alpha^2\|\bfg(\thet_{t}, o_{t})\|^2
\\&
\leq (1+\alpha)\tdrt^2 + 2\alpha\|\bfg(\thet_{t}, o_{t})\|^2
\\&
\overset{\eqref{eq:boundGsquared}}\leq (1+\alpha)\tdrt^2 + 4\alpha L^2(\|\thet_t\|^2 + \sigma^2)
\\&
\leq (1+\alpha)\tdrt^2 + 4\alpha L^2(2r_t^2 + 3\sigma^2).
}
Using the bound established in~\eqref{eq:established}, we can get
\eqal{}{
\tdrp^2&\leq (1+\alpha)\tdrt^2 + 8\alpha L^2r_t^2 + 12\alpha L^2\sigma^2
\\&
\leq (1+\alpha)\tdrt^2 + 84\alpha L^2\sigma^2.
}
From this, we can proceed as follows:
\eqal{}{
\tdrp^2&\leq (1+\alpha)\tdrt^2 + 84\alpha L^2\sigma^2
\\&
\leq (1 + \alpha)^2\tilde{r}_{t-1}^2 + (1+\alpha)84\alpha L^2\sigma^2 + 84\alpha L^2\sigma^2
\\&
\leq (1+\alpha)^{t+1}\tilde{r}_0^2 + 84\alpha L^2\sigma^2\sum_{j = 0}^{t}(1 + \alpha)^j.
}
So, for $0\leq t<\tmix$, 
\eqal{}{
\tdrp^2 \leq (1+\alpha)^{\tmix}\tilde{r}_0^2 + 84\alpha L^2\sigma^2\sum_{j = 0}^{\tmix}(1 + \alpha)^j.
}
Now, given that $L\geq 1$, note that, for $\alpha\tmix\leq\frac{1}{36L^2}$ and $j = 0, ..., \tmix-1$, we have $(1+\alpha)^j\leq(1+\alpha)^\tmix\leq e^{\alpha\tmix}\leq e^{0.25}\leq 2$. Thus, we get
\eqal{eq:boundLessTau}{
\tdrt^2
\leq 2\tilde{r}_0^2 + 84\alpha L^2\sigma^2\tmix\leq 11\sigma^2.
}
Finally,
\eqal{}{
h_t &\leq 9L^2\tdrt^2 + 12L^2\sigma^2
\\&
\leq 9L^2(11\sigma^2) + 12L^2\sigma^2
\\&
\leq 111L^2\sigma^2.}

We now analyze the case in which $t\geq \tmix$. Adding and subtracting $\tilde\thet_{t-\tau_{mix}}$ in the left hand side of the inner product, we have
\begin{equation}
\begin{aligned}
h_t &= \langle \tdxt - \xs, \bfg(\tdxt, o_t) - \bar{\bfg}(\tdxt)\rangle
\\&
= \underbrace{\langle \tilde\xk - \tilde\thet_{t-\tmix},  \mathbf{g}(\tilde\xk, o_t)-\bar{\mathbf{g}}(\tilde\xk) \rangle}_{T_{1}} + \underbrace{\langle \tilde\thet_{t- \tmix} - \xs,  \mathbf{g}(\tilde\xk, o_t)-\bar{\mathbf{g}}(\tilde\xk) \rangle}_{T_{2}},
\end{aligned}
\end{equation}
where, using~\eqref{eq:boundGradNorm}, Cauchy-Schwarz inequality and Lemma~\ref{th:thm1_l3},
\begin{equation}\label{eq:deltaThet}
\begin{aligned}
T_{1}&\leq \|\tilde\xk - \tilde\thet_{t-\tmix}\|(\|\mathbf{g}(\tilde\xk, o_t)\| + \|\bar{\mathbf{g}}(\tilde\xk)\|)
\\&
\overset{\eqref{eq:boundGradNorm}}{\leq} \|\tilde\xk - \tilde\thet_{t-\tmix}\|2L(\|\tilde\xk\| + \sigma)
\\&
\leq 2\alpha L^2\left(\sigma\tau_{mix}+ \sum_{l = t-\tmix}^{t-1}\|\thet_{l}\|\right)(\|\tilde\xk\| + \sigma)
\\&
\leq 2\alpha L^2\sigma\tau_{mix}(\|\tilde\xk\| + \sigma)+ 2\alpha L^2\sum_{l = t-\tmix}^{t-1}\|\thet_{l}\|(\|\tilde\xk\| + \sigma)
\\&
\overset{\eqref{eq:halfSq}}\leq2\alpha L^2\sigma^2\tau_{mix} + 2\alpha L^2\tau_{mix}\sigma\|\tilde\xk\|
\\&
+ 2\alpha L^2\sum_{l = t-\tmix}^{t-1}\left(\frac{1}{2}\|\thet_{l}\|^2 + \frac{1}{2}(\|\tilde\xk\| + \sigma)^2\right)
\\&
\overset{\eqref{eq:sumSquared}}\leq2\alpha L^2\sigma^2\tau_{mix} + \alpha L^2\tau_{mix}\sigma^2 + \alpha L^2\tau_{mix}\|\tilde\xk\|^2
\\&
+ 2\alpha L^2\sum_{l = t-\tmix}^{t-1}\left(\frac{1}{2}\|\thet_{l}\|^2 + \|\tilde\xk\|^2 + \sigma^2\right)
\\&
\leq 11\alpha L^2\sigma^2\tau_{mix} + 6\alpha L^2\tau_{mix}\tdrt^2 + \alpha L^2\sum_{l = t-\tmix}^{t-1}\|\thet_{l}\|^2.
\end{aligned}
\end{equation}
So, taking the expectation,
\begin{equation}\label{eq:expect_t11}
    \E{T_{1}}\leq 11\alpha L^2\sigma^2\tau_{mix} + 6\alpha L^2\tau_{mix}\E{\tdrt^2} + \alpha L^2\sum_{l = t-\tmix}^{t-1}\E{\|\thet_{l}\|^2}.
\end{equation}
Now, we focus on $T_{2}$. Note that, adding and subtracting $\mathbf{g}(\tilde\thet_{t-\tmix}, o_t)$ and $\bar{\mathbf{g}}(\tilde\thet_{t-\tmix})$ to the right hand side of the inner product, we can write
\begin{equation}
    \begin{aligned}
        T_{2} &= \langle \tilde\thet_{t-\tmix} - \xs,  \mathbf{g}(\tilde\xk, o_t)-\bar{\mathbf{g}}(\tilde\xk)\rangle\\&
= \bar{T}_1 + \bar{T}_2 + \bar{T}_3
\end{aligned}
\end{equation}
with
\begin{equation}
\begin{aligned}
\bar{T}_1 &= \langle \tilde\thet_{t-\tmix} - \xs,  \mathbf{g}(\tilde\thet_{t-\tmix}, o_t)-\bar{\mathbf{g}}(\tilde\thet_{t-\tmix})\rangle
\\
\bar{T}_2 &= \langle \tilde\thet_{t-\tmix} - \xs,  \mathbf{g}(\tilde\thet_{t}, o_t)-{\mathbf{g}}(\tilde\thet_{t-\tmix}, o_t)\rangle
\\
\bar{T}_3 &= \langle \tilde \thet_{t-\tmix} - \xs,  \bar{\mathbf{g}}(\tilde\thet_{t-\tmix})-\bar{\mathbf{g}}(\tilde\thet_{t})\rangle.
\end{aligned}
\end{equation}
We first bound $\bar{T}_2$ and $\bar{T}_3$. Note that, using the Lipschitz property of the TD update direction~\eqref{eq:Lipschitz} and Lemma~\ref{th:thm1_l3},
\begin{equation}
\begin{aligned}
\bar{T}_2 &\leq \|\tilde\thet_{t-\tmix} - \xs\|\|\mathbf{g}(\tilde\xk, o_t) - \mathbf{g}(\tilde\thet_{t-\tmix}, o_t)\|
\\&
\leq L\|\tilde\thet_{t-\tmix} - \xs\|\|\tilde\thet_{t-\tmix} - \tilde\xk\|
\\&
\leq L\|\tilde\thet_{t-\tmix} -\tilde\thet_{t}+\tilde\thet_{t} - \xs\|\|\tilde\thet_{t-\tmix} - \tilde\xk\|
\\&
\leq L\tdrt\|\tilde\thet_{t-\tmix} - \tilde\xk\| + L\|\tilde\thet_{t-\tmix} - \tilde\xk\|^2
\\&
\leq L^2\alpha\left(\sigma\tau_{mix}+ \sum_{l = t-\tmix}^{t-1}\|\thet_{l}\|\right)\tdrt + L\left(2L^2\alpha^2\tau_{mix}^2\sigma^2 + 2L^2\alpha^2\tau_{mix}\sum_{l = t-\tmix}^{t-1}\|\thet_{l}\|^2\right)
\\&
= L^2\alpha\tmix\frac{1}{2}\left(\sigma^2 + \tdrt^2 \right) + \frac{1}{2}L^2\alpha  \sum_{l = t-\tmix}^{t-1}\left(\|\thet_{l}\|^2+ \tdrt^2\right)
\\&
+ 2L^3\alpha^2\tmix^2\sigma^2 + 2L^3\alpha^2\tmix\sum_{l = t-\tmix}^{t-1}\|\thet_{l}\|^2
\\&
\leq \alpha\tmix L^2\sigma^2 + \alpha\tmix L^2\tdrt^2 + \alpha L^2\sum_{l = t-\tmix}^{t-1}\|\thet_{l}\|^2,
\end{aligned}
\end{equation}
where in the last inequality we used $\alpha\leq \frac{1}{8\tmix L}$. Taking the expectation,
\begin{equation}
\E{\bar{T}_2}\leq \alpha\tmix L^2\sigma^2 + \alpha\tmix L^2\E{\tdrt^2} + \alpha L^2\sum_{l = t-\tmix}^{t-1}\E{\|\thet_{l}\|^2}.
\end{equation}
With the same calculations, we can get 
\begin{equation}
\E{\bar{T}_3}\leq \alpha\tmix L^2\sigma^2 + \alpha\tmix L^2\E{\tdrt^2} + \alpha L^2\sum_{l = t-\tmix}^{t-1}\E{\|\thet_{l}\|^2}.
\end{equation}
We now proceed to bound $\bar{T}_1$.
\begin{equation}
\begin{aligned}
\E{\bar{T}_1} &= \E{\langle \tilde\thet_{t-\tmix} - \xs, \mathbf{g}(\tilde\thet_{t-\tmix}, o_t) - \bar{\mathbf{g}}(\tilde\thet_{t-\tmix}) \rangle}
\\&
= \E{\langle \tilde \thet_{t-\tmix} - \xs, \E{\mathbf{g}(\tilde\thet_{t-\tmix}, o_t)|o_{t-\tmix},\tilde \thet_{t-\tmix}} - \bar{\mathbf{g}}(\tilde\thet_{t-\tmix}) \rangle}
\\&
\leq \E{\|\tilde\thet_{t-\tmix} - \xs\| \|\E{\mathbf{g}(\tilde\thet_{t-\tmix}, o_t)|o_{t-\tmix}, \tilde\thet_{t-\tmix}} - \bar{\mathbf{g}}(\tilde\thet_{t-\tmix})\|}
\\&
\overset{(*)}\leq \alpha\E{\|\tilde\thet_{t-\tmix} - \xs\|(\|\tilde\thet_{t-\tmix}\| + \sigma)}
\\&
\leq \alpha\E{\|\tilde\thet_{t-\tmix} - \xs\|(\|\tilde\thet_{t-\tmix} - \xs\| + 2\sigma)}
\\&
\leq \alpha\E{\frac{1}{2}\|\tilde\thet_{t-\tmix} - \xs\|^2 + \frac{1}{2}(\|\tilde\thet_{t-\tmix} - \xs\| + 2\sigma)^2}
\\&
\leq \alpha\E{\frac{1}{2}\|\tilde\thet_{t-\tmix} - \xs\|^2 + \|\tilde\thet_{t-\tmix} - \xs\|^2 + 2\sigma^2}
\\&
\leq 2\alpha\E{\|\tilde\thet_{t-\tmix} - \xs\|^2 + \sigma^2}
\\&
\leq 2\alpha\E{2\|\tilde\thet_t - \xs\|^2 +2\|\tilde\thet_t-\tilde\thet_{t-\tmix} \|^2+ \sigma^2}
\\&
\leq 2\alpha\E{2\tdrt^2 +2 ( 2L^2\alpha^2\tau_{mix}^2\sigma^2 + 2L^2\alpha^2\tau_{mix}\sum_{l = t-\tmix}^{t-1}\|\thet_{l}\|^2)+ \sigma^2}
\\&
\leq 4\alpha\E{\tdrt^2} + 3\alpha\sigma^2 + \alpha\sum_{l = t-\tmix}^{t-1}\E{\|\thet_{l}\|^2},
\end{aligned}
\end{equation}
where for $(*)$ we used Definition~\ref{def:mix} of mixing time and the fact that $\sigma \geq 1$, and in the last inequality we used $\alpha \leq \frac{1}{8\tmix L}$. So, we get 
\begin{equation}
\begin{aligned}
\E{T_{2}}& = \E{\bar{T}_1} + \E{\bar{T}_2} + \E{\bar{T}_3}  
\\&
\leq 6\alpha\tmix L^2\E{\tdrt^2} + 5\alpha\tmix L^2\sigma^2 + 3\alpha L^2\sum_{l = t-\tmix}^{t-1}\E{\|\thet_{l}\|^2}.
\end{aligned}
\end{equation}
Finally, we get
\begin{equation}
\begin{aligned}
\E{h_t} &= \E{T_{1}} + \E{T_2}
\\&
\leq 16\alpha\tmix L^2\sigma^2 + 12\alpha\tmix L^2\E{\tdrt^2} + 4\alpha L^2\sum_{l = t-\tmix}^{t-1}\E{\|\thet_{l}\|^2}
\\&
\leq 16\alpha\tmix L^2\sigma^2 + 12\alpha\tmix L^2\E{\tdrt^2}
\\&
+ 8\alpha L^2\sum_{l = t-\tmix}^{t-1}\left(\E{\|\e_l\|^2} + \E{\|\tilde{\thet}_l\|^2}\right)
\\&
\leq 32\alpha\tmix L^2\sigma^2 + 12\alpha\tmix L^2\E{\tdrt^2}
+ 8\alpha L^2\sum_{l = t-\tmix}^{t-1}\E{\|\e_l\|^2 + 2\tilde{r}_l^2}.
\end{aligned}
\end{equation}   \hspace*{\fill}~$\square$\vspace{0.3cm}

\subsubsection{Proof of Theorem~\ref{thm1}}
First, we have 
\eqal{}{\tdrp^2 = \tdrt^2 + 2 \alpha \langle \bfg(\xk, o_t), \tilde\thet_t - \thet^\star\rangle + \alpha^2 \| \bfg(\xk, o_t) \|^2.}
Then, using~\eqref{lemma:bhand}, i.e.,  $\langle \bar{\bfg}(\tilde\thet_t), \tilde\thet_t - \thet^\star\rangle\leq -\mu\tdrt^2$, we have 
\eqal{eq:mainboundthm1}{
\tdrp^2 &=\tdrt^2 + 2\alpha\langle\bfg(\xk, o_t), \tdxt - \xs\rangle + \alpha^2\|\bfg(\xk, o_t)\|^2
\\&=\tdrt^2 + 2\alpha \langle \bar{\bfg}(\tilde\thet_t), \tilde\thet_t - \thet^\star\rangle + 2\alpha h_t + 2\alpha m_t + \alpha^2\|\bfg(\xk, o_t)\|^2
\\&\leq (1 - 2\alpha\mu)\tdrt^2 + 2\alpha h_t + 2\alpha m_t + \alpha^2n_t.
}
We now apply the inequalities obtained in Lemma \ref{th:thm1_l4} to bound $\E{h_t}$, $m_t$, and $n_t$. Recall that $\bar{\tau} = \max \{\tau, \tau_{mix}\}$. Note that, from Lemma~\ref{th:thm1_l4} - (iii), we can write $\E{h_t}\leq \bar{h}_t$, defining
\eqal{}{
\bar{h}_t = \begin{cases}B  &\textrm{if  }\hspace{0.1cm} 0\leq t < \tmix \\
q_t &\textrm{if  }\hspace{0.1cm} t \geq \tmix
\end{cases},
}
with $B = 111\sigma^2$, and
\eqal{}{
q_t = \alpha\tmix L^2\left(32\sigma^2 + 12\E{\tdrt^2}\right)
+ 8\alpha L^2\sum_{l = t-\tmix}^{t-1}\E{\|\e_l\|^2 + 2\tilde{r}_l^2}.
}
As a consequence, we can write, for every $t\geq 0$,
\eqal{}{
\E{h_t} \leq q_t + \bar{B}_t,
}
where, in turn,
\eqal{}{
\bar{B}_t = \begin{cases}
    B&\textrm{if  }\hspace{0.1cm} 0\leq t < \tmix \\
    0& \textrm{otherwise}
\end{cases}.
}
Also, recall that, from Lemma~\ref{th:thm1_l4}, we have
\eqal{}{
n_t &\leq 4L^2\|\e_t\|^2 + 8L^2\tdrt^2 + 10L^2\sigma^2,
\\
m_t&\leq 6\alpha\tau L^2\sigma^2 + 3\alpha\tau L^2\tdrt^2 + 2\alpha L^2\sum_{l = t-\tau}^{t-1}\left(\|\e_l\|^2+2\tilde{r}_l^2\right).
}
Combining these inequalities together, we have, for $t\geq 0$,

\eqal{eq: mainbound}{
\E{\tdrp^2} &\leq (1 - 2\alpha\mu)\E{\tdrt^2} + 2\alpha \E{h_t} + 2\alpha \E{m_t} + \alpha^2\E{\|\bfg(\xk, o_t)\|^2}
\\&
\leq(1 - 2\alpha\mu)\E{\tdrt^2} + 2\alpha^2\tmix L^2\left(32\sigma^2 + 12\E{\tdrt^2}\right)
\\&
+ 16\alpha^2 L^2\sum_{l = t-\tmix}^{t-1}\E{\|\e_l\|^2 + 2\tilde{r}_l^2}
\\&
+12\alpha^2\tau L^2\sigma^2 + 6\alpha^2\tau L^2\E{\tdrt^2} + 4\alpha^2 L^2\sum_{l = t-\tau}^{t-1}\E{\|\e_l\|^2 +2\tilde{r}_l^2}
\\&
+ 4\alpha^2L^2\E{\|\e_t\|^2 + 2\tdrt^2} + 10\alpha^2L^2\sigma^2 + 2\alpha\bar{B}_t.
}
Combining terms, we can get
\eqal{eq:rewriteMe}{
\E{\tdrp^2} &\leq (1 - 2\alpha\mu+48\alpha^2L^2\bar{\tau})\E{\tdrt^2} + 128\alpha^2L^2\bar{\tau}\sigma^2
\\&
+ 4\alpha^2L^2\E{\|\e_t\|^2} + 20\alpha^2 L^2\sum_{l = t-\bar{\tau}}^{t-1}\E{\|\e_l\|^2 + 2\tilde{r}_l^2}+ 2\alpha\bar{B}_t,
}
where we have used $\tau + \tau_{mix} \leq 2\bar{\tau}$. 
Now, using the fact that $r_t^2\leq 2\tdrt^2 + 2\|\e_t\|^2$, which implies $-\tdrt^2\leq -\frac{r_t^2}{2} + \|\e_t\|^2$, we have 
\eqal{}{
(1 - 2\alpha\mu+48\alpha^2L^2\bar{\tau})\E{\tdrt^2}&=(1 - \alpha\mu+48\alpha^2L^2\bar{\tau})\E{\tdrt^2} -\alpha\mu\E{\tdrt^2}
\\&
\leq (1 - \alpha\mu+48\alpha^2L^2\bar{\tau})\E{\tdrt^2} -\alpha\mu\frac{\E{r_t^2}}{2} + \alpha\E{\|\e_t\|^2},
}
and, using $\mu \leq 1$, we can re-write~\eqref{eq:rewriteMe} as
\eqal{}{
\E{\tdrp^2} &\leq (1 - \alpha\mu+48\alpha^2L^2\bar{\tau})\E{\tdrt^2} -\alpha\mu\frac{\E{r_t^2}}{2} + 128\alpha^2L^2\bar{\tau}\sigma^2 \\&+ \alpha(1 + 4\alpha L^2)\E{\|\e_t\|^2} + 20\alpha^2L^2\sum_{l = t-\bar{\tau}}^{t-1}\E{\|\e_l\|^2 + 2\tilde{r}_l^2}+ 2\alpha\bar{B}_t. 
}
Multiplying both sides by $w_t$, we have 
\eqal{}{
w_t\E{\tdrp^2} &\leq (1 - \alpha\mu+48\alpha^2L^2\bar{\tau})w_t\E{\tdrt^2} -\alpha\mu\frac{w_t\E{r_t^2}}{2} + 128w_t\alpha^2L^2\bar{\tau}\sigma^2 \\&+ \alpha(1 + 4\alpha L^2)w_t\E{\|\e_t\|^2} + 20\alpha^2L^2w_t\sum_{l = t-\bar{\tau}}^{t-1}\E{\|\e_l\|^2 + 2\tilde{r}_l^2}+ 2\alpha w_t\bar{B}_t.}
By summing over $t=0, ..., T$, we get, with $W_T = \sum_{t=0}^Tw_t$: 
\eqal{eq:p1p2}{
\sum_{t=0}^T w_t\E{\tdrp^2} &\leq (1 - \alpha\mu+48\alpha^2L^2\bar{\tau})\sum_{t=0}^Tw_t\E{\tdrt^2} -\frac{\alpha\mu}{2}\sum_{t=0}^Tw_t\E{r_t^2} \\&+ 128W_T\alpha^2L^2\bar{\tau}\sigma^2
+ \alpha(1+4\alpha L^2)\underbrace{\sum_{t=0}^Tw_t\E{\|\e_t\|^2}}_{p_1} \\&+ 20\alpha^2 L^2\underbrace{\sum_{t=0}^Tw_t\sum_{l = t-\bar{\tau}}^{t-1}\E{\|\e_l\|^2 + 2\tilde{r}_l^2}}_{p_2} + 2W_{\tmix-1}\alpha B.}
Note that, from Lemma~\ref{th:thm1_l1} - (iii), we have, picking $\alpha\leq \frac{1}{72\tau L^2}$,
\begin{equation}\label{eq:p1_bound}
\begin{aligned}
p_1 = \sum_{t=0}^{T}w_t\E{\|\e_t\|^2} &\leq 4W_T\alpha^2\tau^2 L^2\sigma^2+ 16\alpha^2\tau^2 L^2\sum_{t = 0}^{T}{w_t}\E{\|\tilde{\thet}_t\|^2}
\\&
\leq 36W_T\alpha^2\tau^2 L^2\sigma^2+ 32\alpha^2\tau^2 L^2\sum_{t = 0}^{T}{w_t}\E{\tilde{r}_t^2}
\\&
\leq \frac{\alpha\tau W_T\sigma^2}{2}+ \frac{\alpha\tau}{2}\sum_{t = 0}^{T}{w_t}\E{\tilde{r}_t^2}.
\end{aligned}
\end{equation}
Furthermore, using the fact that $w_t \leq 2w_l$ for $l= t-\bar{\tau}, ..., t-1$, the above bound on $p_1$, and picking $\alpha \leq \frac{1}{72\tau L^2}$, we can bound $p_2$ as follows: 
\eqal{}{
p_2 &= \sum_{t=0}^Tw_t\sum_{l = t-\bar{\tau}}^{t-1}\E{\|\e_l\|^2 + 2\tilde{r}_l^2}
\\&
\overset{(a)}{\leq} 2\sum_{t=0}^T\sum_{l = t-\bar{\tau}}^{t-1}w_l\E{\|\e_l\|^2 + 2\tilde{r}_l^2}
\\&
\overset{(b)}{\leq} 
2\bar{\tau}\sum_{t=0}^Tw_t\E{\|\e_t\|^2 + 2\tilde{r}_t^2}.
\\&
\leq 2\bar\tau\sum_{t=0}^Tw_t\E{\|\e_t\|^2} + 4\bar\tau\sum_{t=0}^Tw_t\E{\tilde{r}_t^2}
\\&
\overset{(c)}\leq 2\bar\tau \left(\frac{\alpha\tau W_T\sigma^2}{2}+ \frac{\alpha\tau}{2}\sum_{t = 0}^{T}{w_t}\E{\tilde{r}_t^2}\right)+4\bar\tau\sum_{t=0}^Tw_t\E{\tilde{r}_t^2}
\\&
\leq 5\bar{\tau}\sum_{t=0}^Tw_t\E{\tdrt^2} + W_T\sigma^2\bar{\tau},
}
where for $(a)$ we used Lemma~\ref{lem:weights}, for $(b)$ we used the fact that each element $w_l\|\thet_l\|^2$ appears at most $\tau$ times in the sum, for $l = 0, ..., T-1$ (note that, by definition, $\|\e_l\| = \tilde{r}_l = 0$ for $l<0$), and for $(c)$, we used the bound on $p_1$. In the last inequality we simply used $\alpha\tau\leq 1$. Plugging the two bounds on $p_1$ and $p_2$ in~\eqref{eq:p1p2}, {we get}
\eqal{eq:71}{
\sum_{t=0}^T w_t\E{\tdrp^2} &\leq (1 - \alpha\mu+150\alpha^2L^2\bar{\tau})\sum_{t=0}^Tw_t\E{\tdrt^2}-\frac{\alpha\mu}{2}\sum_{t=0}^Tw_t\E{r_t^2}
\\&
+ 150W_T\alpha^2L^2\bar{\tau}\sigma^2 + 2W_{\tmix-1}\alpha B.
}
Now, note that for $\alpha \leq \frac{\mu}{100L^2\bar{\tau}}$, it holds  that $(1 - 2\alpha\mu+150\alpha^2L^2\bar{\tau})\leq (1 - 0.5\alpha\mu)$.  We can then re-write~\eqref{eq:71} as
\eqal{eq:71bis}{
\sum_{t=0}^T w_t\E{\tdrp^2} &\leq (1 - 0.5\alpha\mu)\sum_{t=0}^Tw_t\E{\tdrt^2}-\frac{\alpha\mu}{2}\sum_{t=0}^Tw_t\E{r_t^2}
\\&
+ 150W_T\alpha^2L^2\bar{\tau}\sigma^2
+ 2W_{\tmix-1}\alpha B.
}

Now, dividing by $W_T$ both sides of~\eqref{eq:71bis}, bringing $\sum_{t=0}^Tw_t\E{\tdrp^2}$ to the right hand side of the inequality and $-\frac{\alpha\mu}{2}\sum_{t=0}^Tw_t\E{r_t^2}$ to the left side, we get
\eqal{eq:rewrAg}{
\frac{\alpha\mu}{2}\sum_{t=0}^T\frac{w_t}{W_T}\E{r_t^2}\leq \frac{1}{W_T}\sum_{t=0}^T\left(w_t(1 - 0.5\alpha\mu)\E{\tdrt^2} - w_t\E{\tdrp^2}\right)& \\+150\alpha^2L^2\bar{\tau}\sigma^2 + \frac{2W_{\tmix-1}\alpha B}{W_T}.&}
Now, recalling that $w_t = (1-0.5\alpha\mu)^{-(t+1)}$, note that $w_t(1-0.5\alpha\mu) = w_{t-1}$, and we can get, noting that $w_{-1} = 1$,
\eqal{}{
\sum_{t=0}^T\left(w_t(1 - 0.5\alpha\mu)\E{\tdrt^2} - w_t\E{\tdrp^2}\right) &= \sum_{t=0}^T\left(w_{t-1}\E{\tdrt^2} - w_t\E{\tdrp^2}\right)
\\&
\leq \E{\tilde{r}_0^2} - w_{T}\E{\tilde{r}_{T+1}^2}\leq \tilde{r}_0^2.
}
Hence, we can write~\eqref{eq:rewrAg} as
\eqal{eq:againRW}{
\frac{\alpha\mu}{2}\sum_{t=0}^T\frac{w_t}{W_T}\E{r_t^2}\leq \frac{{\tilde{r}_{0}^2}}{W_T}+150\alpha^2L^2\bar{\tau}\sigma^2+\frac{2W_{\tmix-1}\alpha B}{W_T}.
}
Now note that 
\eqal{}{
W_{\tmix-1} = \sum_{t = 0}^{\tmix-1}w_t = \sum_{t = 0}^{\tmix-1}(1-0.5\alpha\mu)^{-(t+1)}\leq \sum_{t = 0}^{\tmix-1}(1+\alpha\mu)^{t+1}  \leq 2\tmix
}
and that 
\eqal{}{
\frac{1}{W_T}\leq \frac{1}{w_T} = (1-0.5\alpha\mu)^{T+1},
}
from which we can obtain, re-arranging the different terms in~\eqref{eq:againRW},
\eqal{}{
\frac{1}{W_T}\sum_{t = 0}^{T}w_t\E{ r_{t+1}^2}&\leq (1-0.5\alpha\mu)^{T+1}{\tilde{r}_{0}^2}\left(\frac{2}{\alpha\mu} + \frac{4\bar{\tau}B}{\mu}\right) +300\frac{\alpha L^2\bar{\tau}\sigma^2}{\mu}
\\&
= C_{\alpha}(1-0.5\alpha\mu)^{T+1}\tilde{r}_0^2 + C_2\frac{\alpha L^2\bar{\tau}\sigma^2}{\mu},
}
where we define $C_{\alpha} = \left(\frac{2}{\alpha\mu} + \frac{4\bar{\tau}B}{\mu}\right)$ and $C_{2} = 300$.
By plugging the maximum value for the step size $\alpha = \frac{\mu}{150L^2\bar{\tau}}$, we can get, defining $C_{\bar{\tau}} = \frac{\bar{\tau}}{\mu}\left(\frac{2C_1L^2}{\mu} + {4B}\right)$,
\eqal{}{
\frac{1}{W_T}\sum_{t = 0}^{T}w_t\E{ r_{t+1}^2}&\leq C_2(1-0.5\alpha\mu)^{T+1}\tilde{r}_0^2 + 2{\sigma^2}.
}
Indeed, for $\alpha = \frac{\mu}{150L^2\bar{\tau}}$, it holds $C_2\frac{\alpha L^2\bar{\tau}}{\mu} = 2$. Finally, by definition of $\thet_{out}$ in Theorem~\ref{thm1}, note that we have 
\eqal{}{
\E{\|\thet_{out}-\thet^*\|^2}=\frac{1}{W_T}\sum_{t=0}^{T} w_t \E{ r_{t}^2},} and we can conclude the proof of Theorem~\ref{thm1}.
\hspace*{\fill}~$\square$\vspace{0.3cm}
\newpage

\subsection{Proof of Theorem \ref{thm2} and Related Lemmas}\label{sec:proof of thm2}

Let $r_t \triangleq \|\xk - \xs\|$. Define $\tau' = 2\tau_{max} + \tmix$, and recall
\begin{equation}
	\begin{aligned}
		{r}_{t, 2} &\triangleq \max_{t-\tau'\leq l \leq t}{\E{r_l^2}}.
	\end{aligned}
\end{equation}
As we mentioned in Section~\ref{sec:constantDelay}, the proof technique of Theorem~\ref{thm1} cannot be directly extended to a time-varying delay setting. To see why this is the case, please take a look back at~\eqref{eq:etdefExt} in the proof of Theorem~\ref{thm1}, and note that the specific way in which we can write $\e_t$, which is crucial in several key steps of the proof, \emph{relies on the fact that the delay is constant}. Accordingly, we do not see how we could generalize that type of identity to a time-varying delay setting in a way that leads to the convergence rate we are aiming to obtain. In the rest of this section, we provide the details and the proofs of the auxiliary lemmas introduced in Section~\ref{sec:vanillaDelayed}, and the proof of Theorem~\ref{thm2}, whose structure departs completely from the proof of Theorem~\ref{thm1} and whose outline has been illustrated in Section~\ref{sec:vanillaDelayed}.
\subsubsection{Proofs of Auxiliary Lemmas}
We start by proving Lemma \ref{th:thm2_l1}, i.e., the bounds on terms of the form $\|\thet_t - \thet_{t-\tau}\|^2$, for some $0\leq \tau\leq t$. \vspace{0.4cm}

\noindent\textbf{Proof of Lemma~\ref{th:thm2_l1}.} To prove (i), note that we can get
\eqal{}{
{\|\thet_t - \thet_{t-\tmix}\|^2} &\leq \left(\sum_{l = t-\tmix}^{t-1}\|\thet_{l+1} - \thet_l\|\right)^2
\\&
\overset{\eqref{eq:sumSquared}}\leq \tmix\sum_{l = t-\tmix}^{t-1}\|\thet_{l+1} - \thet_l\|^2
\\&
= \tmix \alpha^2\sum_{l = t-\tmix}^{t-1}\|\bfg(\thet_{l-\tau_l}, o_{l-\tau_l})\|^2
\\&
\overset{\eqref{eq:boundGsquared}}\leq 2\alpha^2\tmix L^2\sum_{l = t-\tmix}^{t-1}(\|\thet_{l-\tau_l}\|^2 + \sigma^2)
\\&
\leq 2\alpha^2\tmix L^2\sum_{l = t-\tmix}^{t-1}(2r_{l-\tau_l}^2 + 3\sigma^2).
}
Taking the expectation on both sides of the inequality, we get
\eqal{}{
\E{\|\thet_t - \thet_{t-\tmix}\|^2}&\leq 2\alpha^2\tmix L^2\sum_{l = t-\tmix}^{t-1}(2\E{r_{l-\tau_l}^2} + 3\sigma^2)
\\&
\leq 2\alpha^2\tmix L^2\sum_{l = t-\tmix}^{t-1}(2\max_{t-\tmix-\tau_{max}\leq j\leq t}\E{r_j^2} + 3\sigma^2)
\\&
\leq 4\tmix^2\alpha^2L^2r_{t,2} + 6\alpha^2\tmix^2 L^2\sigma^2
\\&
= 2\alpha^2\tmix^2L^2(2r_{t,2} + 3\sigma^2).
}
With analogous computations, we can get part (ii) of the Lemma, i.e.
\eqal{}{
\E{\|\thet_t - \thet_{t-\tau_t}\|^2}\leq 2\alpha^2\tau_{max}^2L^2(2r_{t,2} + 3\sigma^2).
}\hspace*{\fill}~$\square$\vspace{0.3cm}

\noindent Recall the definition of $\mathbf{e}_t$: 
\begin{equation}
	\mathbf{e}_t \triangleq \mathbf{g}(\xk, o_t) - \mathbf{g}(\thet_{t-\tau_t}, o_{t-\tau_t}).
\end{equation}
As illustrated in the outline of the analysis in Section~\ref{sec:vanillaDelayed}, for the purpose of analysis, we write the update rule as follows: 
\begin{equation}
	\xkp = \xk + \alpha\mathbf{g}(\xk, o_t) - \alpha \mathbf{e}_t,
\end{equation}
from which we can write
\begin{equation}\label{eq:analTauMaxApp}
	\begin{aligned}
		\|\xkp - \xs\|^2 = J_{t,1} + \alpha^2J_{t,2} - 2\alpha J_{t,3},
	\end{aligned}
\end{equation}
with 
\begin{equation}\vspace{0.4cm}
\begin{aligned}
	J_{t, 1} &\triangleq \|\xk - \xs + \alpha\mathbf{g}(\xk, o_t)\|^2\\
	J_{t, 2} &\triangleq \|\mathbf{e}_t\|^2\\
	J_{t, 3} &\triangleq \langle \mathbf{e}_t, \xk - \xs + \alpha\mathbf{g}(\xk, o_t) \rangle.
\end{aligned}
\end{equation}
\textbf{Proof of Lemma~\ref{th:thm2_l2} - (i).}  
Note that
\begin{equation}
\begin{aligned}
	J_{t,1} = \|\xk - \xs + \alpha\mathbf{g}(\xk, o_t)\|^2 &= r_t^2 + 2\alpha\underbrace{\langle \xk - \xs,\mathbf{g}(\xk, o_t) \rangle}_{J_{t,11}}
	\\& + \alpha^2\underbrace{\|\mathbf{g}(\xk, o_t)\|^2}_{J_{t,12}}.
\end{aligned}
\end{equation}
We also observe that 
\begin{equation}
\begin{aligned}
\E{J_{t, 12}} &= \E{\|\mathbf{g}(\xk, o_t)\|^2} 
\\&
\leq \E{2L^2\left(\|\xk\|^2 + \sigma^2\right)}
\\&
\leq 2L^2\left(2\E{r_t^2} + 3\sigma^2\right)
\\&
\leq 2L^2\left(2r_{t,2} + 3\sigma^2\right). 
\end{aligned}
\end{equation}
Now note that, using~\eqref{lemma:bhand},
\begin{equation}
\begin{aligned}
J_{t,11} &= \langle \xk  - \xs, \mathbf{g}(\xk, o_t)\rangle = -\langle \xs - \xk, \bar{\mathbf{g}}(\xk)\rangle\\& + \langle \xk  - \xs, \mathbf{g}(\xk, o_t)-\bar{\mathbf{g}}(\xk)\rangle \\&
\leq -\mu r_{t}^2 + \underbrace{\langle \xk  - \xs, \mathbf{g}(\xk, o_t)-\bar{\mathbf{g}}(\xk)\rangle}_{T_1'},
\end{aligned}
\end{equation}
where we now omit the dependence on the iterate $t$ in the terms we bound, for notational convenience. Now, note that
\begin{equation}
\begin{aligned}
T_1' = \underbrace{\langle \xk - \thet_{t-\tmix},  \mathbf{g}(\xk, o_t)-\bar{\mathbf{g}}(\xk) \rangle}_{T_{11}'} + \underbrace{\langle \thet_{t- \tmix} - \xs,  \mathbf{g}(\xk, o_t)-\bar{\mathbf{g}}(\xk) \rangle}_{T_{12}'},
\end{aligned}
\end{equation}
where, using the Cauchy-Schwarz inequality and the triangle inequality, 
\begin{equation}\label{eq:deltaThet}
\begin{aligned}
T_{11}'&\leq \|\xk - \thet_{t-\tmix}\|(\|\mathbf{g}(\xk, o_t)\| + \|\bar{\mathbf{g}}(\xk)\|)
\\&
\overset{\eqref{eq:boundGradNorm}}\leq 2L(\|\xk - \thet_{t-\tmix}\|(\|\xk\| + \sigma))
\\&
\overset{(*)}\leq L\left(\frac{1}{\alpha\tmix L}\|\xk - \thet_{t-\tmix}\|^2 + \alpha\tmix L(\|\xk\| + \sigma)^2\right)
\\&
\overset{\eqref{eq:sumSquared}}\leq L\left(\frac{1}{\alpha\tmix L}\|\xk - \thet_{t-\tmix}\|^2 + 2\alpha\tmix L(\|\xk\|^2 + \sigma^2)\right),
\end{aligned}
\end{equation}
where for $(*)$ we used the fact that,  from~\eqref{eq:halfSq}, we have 
\begin{equation}\label{eq:halfVar}
    ab = (\frac{1}{\sqrt{c}}a)(\sqrt{c}b)\leq \frac{1}{2c}a^2 + \frac{cb^2}{2},
\end{equation} 
specifically with $c = \alpha\tmix L$.
Taking the expectation on both sides and applying (ii) of Lemma~\ref{th:thm2_l1}, we get
\begin{equation}
\begin{aligned}
\E{T_{11}'}&\leq L\left(\frac{1}{2\alpha\tmix L}\E{\|\xk - \thet_{t-\tmix}\|^2} + \alpha\tmix L(2\E{r_t^2} +3 \sigma^2)\right)
\\&
\leq L\left(\frac{2\alpha^2\tmix^2L^2}{2\alpha\tmix L}(2r_{t,2} + 3\sigma^2) +\alpha\tmix L(2r_{t,2} + 3\sigma^2)\right)
\\&
= 4\alpha\tmix L^2 r_{t,2} + 6\alpha\tmix L^2\sigma^2.
\end{aligned}
\end{equation}
Now, we proceed to bound $\E{T_{12}'}$. Note that
\begin{equation}
\begin{aligned}
T_{12}' &= \langle \thet_{t-\tmix} - \xs,  \mathbf{g}(\xk, o_t)-\bar{\mathbf{g}}(\xk)\rangle\\&
 = \bar{T}_1 + \bar{T}_2 + \bar{T}_3,
\end{aligned}
\end{equation}
with
\begin{equation}
\begin{aligned}
\bar{T}_1 &= \langle \thet_{t-\tmix} - \xs,  \mathbf{g}(\thet_{t-\tmix}, o_t)-\bar{\mathbf{g}}(\thet_{t-\tau})\rangle
\\
\bar{T}_2 &= \langle \thet_{t-\tmix} - \xs,  \mathbf{g}(\thet_{t}, o_t)-{\mathbf{g}}(\thet_{t-\tmix}, o_t)\rangle
\\
\bar{T}_3 &= \langle \thet_{t-\tmix} - \xs,  \bar{\mathbf{g}}(\thet_{t-\tmix})-\bar{\mathbf{g}}(\thet_{t})\rangle.
\end{aligned}
\end{equation}
We first bound $\bar{T}_2$ and $\bar{T}_3$. Note that, using the Lipschitz property of the TD update direction~\eqref{eq:Lipschitz}, and calculations similar to the ones used to bound $\E{T_{11}'}$, we get
\begin{equation}\label{eq:(16)1}
\begin{aligned}
\bar{T}_2 &\leq \|\thet_{t-\tmix} - \xs\|\|\mathbf{g}(\xk, o_t) - \mathbf{g}(\thet_{t-\tmix}, o_t)\|
\\&
\leq L\|\thet_{t-\tmix} - \xs\|\|\thet_{t-\tmix} - \xk\|
\\&
\overset{\eqref{eq:halfVar}}\leq \frac{L^2\alpha\tmix}{2}r_{t-\tmix}^2 + \frac{\|\thet_t - \thet_{t-\tmix}\|^2}{2\alpha\tmix}.
\end{aligned}
\end{equation}
Taking the expectation and applying (ii) of Lemma~\ref{th:thm2_l1}, we can get
\begin{equation}\label{eq:(16)2}
\begin{aligned}
\E{\bar{T}_2}&\leq \frac{\alpha\tmix L^2 r_{t,2}}{2} + 2\alpha\tmix L^2r_{t,2} + 3\alpha\tmix L^2 \sigma^2
\\&
\leq 3\alpha\tmix L^2(r_{t,2} + \sigma^2).
\end{aligned}
\end{equation}
With the same calculations, we can get 
\begin{equation}
	\E{\bar{T}_3}\leq 3\alpha\tmix L^2(r_{t,2}^2 + \sigma^2).
\end{equation}
We now proceed to bound $\bar{T}_1$.
\begin{equation}\label{eq:T1bar}
\begin{aligned}
\E{\bar{T}_1} &= \E{\langle \thet_{t-\tmix} - \xs, \mathbf{g}(\thet_{t-\tmix}, o_t) - \bar{\mathbf{g}}(\thet_{t-\tmix}) \rangle}
\\&
= \E{\langle \thet_{t-\tmix} - \xs, \E{\mathbf{g}(\thet_{t-\tmix}, o_t)|o_{t-\tau}, \thet_{t-\tmix}} - \bar{\mathbf{g}}(\thet_{t-\tmix}) \rangle}
\\&
\leq \E{\|\thet_{t-\tmix} - \xs\| \|\E{\mathbf{g}(\thet_{t-\tmix}, o_t)|o_{t-\tmix}, \thet_{t-\tmix}} - \bar{\mathbf{g}}(\thet_{t-\tmix})\|}
\\&
\overset{(*)}\leq \alpha\E{\|\thet_{t-\tmix} - \xs\|(\|\thet_{t-\tmix}\| + \sigma)}
\\&
\leq \alpha\E{\|\thet_{t-\tmix} - \xs\|(\|\thet_{t-\tmix} - \xs\| + 2\sigma)}
\\&
\leq \alpha\E{\frac{1}{2}\left(r_{t-\tmix}^2 + 2r_{t-\tmix}^2 + 4\sigma^2\right)}
\\&
\leq 2\alpha(r_{t,2} + \sigma^2),
\end{aligned}
\end{equation}
where for $(*)$ we used Definition~\ref{def:mix} of the mixing time and the fact that $\sigma \geq 1$.
So, putting the above bounds together, we get 
\begin{equation}
\begin{aligned}
	\E{T_{12}'} = \E{\bar{T}_1} + \E{\bar{T}_2} + \E{\bar{T}_3} \leq 8\alpha\tmix L^2(r_{t,2}^2 + \sigma^2).
\end{aligned}
\end{equation}
This then implies
\begin{equation}
\begin{aligned}
\E{T_1'} &= \E{T_{11}'} + \E{T_{12}'}
\\&
\leq 4\alpha\tmix L^2 r_{t,2} + 6\alpha\tmix L^2\sigma^2 + 8\alpha\tmix L^2(r_{t,2} + \sigma^2)
\\&
\leq 12\alpha\tmix L^2 r_{t,2} + 14\alpha\tmix L^2 \sigma^2.
\end{aligned}
\end{equation}
We also have
\begin{equation}
\E{J_{t, 11}} \leq -\mu\E{r_{t}^2} + \E{T_1'}. 
\end{equation}
Hence, 
\begin{equation}
\begin{aligned}
\E{J_{t, 1}} &= \E{r_t^2} + 2\alpha\E{J_{t,11}} + \alpha^2\E{J_{t, 12}}
\\&
\leq (1-2\alpha\mu)\E{r_t^2} + 28\alpha^2\tmix L^2 r_{t,2} + 34\alpha^2\tmix L^2\sigma^2,
\end{aligned}
\end{equation}
which concludes the proof of the Lemma.\hspace*{\fill}~$\square$\vspace{0.4cm}

\noindent\textbf{Proof of Lemma~\ref{th:thm2_l2} - (ii).} Note that
\eqal{}{
J_{t,2} = \|\mathbf{e}_t\|^2 &= \|\bfg(\thet_t, o_t) - \bfg(\thet_{t-\tau_t}, o_{t-\tau_t})\|^2
\\&
\overset{\eqref{eq:sumSquared}}\leq 2\left(\|\bfg(\thet_t, o_t)\|^2 + \|\bfg(\thet_{t-\tau_t}, o_{t-\tau_t})\|^2\right)
\\&
\overset{\eqref{eq:boundGsquared}}\leq 2\left(2L^2(\|\thet_t\|^2 + \sigma^2) + 2L^2(\|\thet_{t-\tau_t}\|^2 + \sigma^2)\right)
\\&
\leq 4L^2(2r_t^2 + 3\sigma^2 + 2r_{t-\tau_t} + 3\sigma^2).
}
Taking the expectation, we conclude getting 
\eqal{eq:etSquaredExp}{
\E{J_{t,2}} = \E{\|\mathbf{e}_t\|^2}\leq 8L^2(2r_{t, 2} + 3\sigma^2).
}

\noindent\textbf{Proof of Lemma~\ref{th:thm2_l2} - (iii).}
In the following, we drop the dependence on the iteration $t$ in most of the terms we bound. We write 
\begin{equation}
\begin{aligned}
-J_{t,3} &= \langle \et, \xk - \xs + \alpha\bfgofxo \rangle
\\& 
=  \underbrace{\langle -\et, \xk - \xs \rangle}_{\Delta} + \underbrace{\alpha\langle -\et, \bfgofxo \rangle}_{\bar{\Delta}}.
\end{aligned}
\end{equation}
Note that 
\eqal{}{
\bar{\Delta} &= \alpha \langle -\et, \bfgofxo \rangle \leq \alpha\|\et\|\|\bfgofxo\|
\\&
\leq \frac{\alpha}{2}\left(\|\mathbf{e}_t\|^2 + \|\bfg(\thet_t, o_t)\|^2\right).
}
Using~\eqref{eq:etSquaredExp} and~\eqref{eq:boundGsquared} to bound $\E{\|\mathbf{e}_t\|^2}$ and $\E{\|\bfg(\thet_t, o_t)\|^2}$, respectively, we get
\eqal{}{
\E{\bar{\Delta}}&\leq \frac{\alpha}{2}\left(8L^2\left(2r_{t,2} + 3\sigma^2\right) + 2L^2\left(2r_{t,2} + 3\sigma^2\right)\right)
\\&
= 10\alpha L^2r_{t,2} + 15\alpha L^2\sigma^2.
}
We now proceed to bound $\Delta$ as follows: 
\eqal{}{
\Delta &= \langle -\et, \xk - \xs \rangle = \langle -\bfgofxo + \bfg(\thet_{t-\tau_t}, o_{t-\tau_t}), \xk - \xs\rangle
\\&
= \underbrace{\langle -\bfgofxo +\bfg(\xk, o_{t-\tau_t}), \xk - \xs \rangle}_{\Delta_1}
\\&
+ \underbrace{\langle-\bfg(\xk, o_{t-\tau_t}) + \bfg(\thet_{t-\tau_t}, o_{t-\tau_t}), \xk -\xs\rangle}_{\Delta_2}.
}
Note that, thanks to the Lipschitz property of the TD direction and with calculations analogous to the ones performed to obtain the bound on $\E{\bar{T}_2}$ (see~\eqref{eq:(16)1} and~\eqref{eq:(16)2}), we get
\eqal{}{
\E{\Delta_2} \leq \E{L\|\xk - \thet_{t-\tau_t}\|r_t}\leq 3\alpha\tmix L^2(r_{t,2} + \sigma^2).
}
We now bound $\Delta_1$: 
\eqal{}{
\Delta_1 &= \underbrace{\langle -\bfgofxo + \bfg(\thet_{t-\tmix}, o_t), \xk, -\xs \rangle}_{\Delta_{11}} 
\\&
+ \underbrace{\langle -\bfg(\thet_{t-\tmix}, o_t) + \bfg(\xk, o_{t-\tau_t}), \xk - \xs \rangle}_{\Delta_{12}}.
}
With calculations analogous to the ones performed to obtain the bound on $\E{\bar{T}_2}$ (see~\eqref{eq:(16)1} and~\eqref{eq:(16)2}), we get
\eqal{}{
\E{\Delta_{11}} \leq  \E{L\|\xk - \thet_{t-\tau_t}\|r_t}\leq 3\alpha\tmix L^2(r_{t,2} + \sigma^2).
}
We now proceed to bound $\Delta_{12}$:
\eqal{}{
\Delta_{12} &= \underbrace{\langle -\bfg(\thet_{t-\tmix}, o_t) + \bar{\bfg}(\thet_{t-\tmix}), \xk - \xs \rangle}_{\Delta_1'} 
\\&
+ \underbrace{\langle -\bar{\bfg}(\thet_{t-\tmix}) + \bfg(\xk, o_{t-\tau_t}), \xk - \xs \rangle.}_{\Delta_2'}
}
We have 
\eqal{}{
\Delta_{1}' &= \underbrace{\langle -\bfg(\thet_{t-\tmix}, o_t) +\bar{\bfg}(\thet_{t-\tmix}), \thet_{t-\tmix} - \xs \rangle}_{\Delta_{11}'} 
\\&
+ \underbrace{\langle -\bfg(\thet_{t-\tmix}, o_t) + \bar{\bfg}(\thet_{t-\tmix}), \xk - \thet_{t-\tmix} \rangle.}_{\Delta_{12}'}
}
Note that
\eqal{}{
\Delta_{12}'&\leq \|\bfg(\thet_{t-\tmix}, o_t) - \bar{\bfg}(\thet_{t-\tmix})\|\|\thet_t - \thet_{t-\tmix}\|
\\&
\leq \left(\|\bfg(\thet_{t-\tmix}, o_t)\| + \|\bar{\bfg}(\thet_{t-\tmix})\|\right)\|\thet_t - \thet_{t-\tmix}\|
\\&
\overset{\eqref{eq:boundGradNorm}}\leq 2L\left(\|\thet_{t-\tmix}\| + \sigma\right)\|\thet_t - \thet_{t-\tmix}\|
\\&
\leq
2L\left(r_{t-\tmix}^2 + 2\sigma\right)\|\thet_t - \thet_{t-\tmix}\|
\\&
\leq
2\alpha L^2\tmix(r_{t-\tmix}^2  +2\sigma^2) + \frac{1}{2\alpha\tmix}\|\thet_t-\thet_{t-\tmix}\|^2.
}
Taking expectation on both sides and applying Lemma~\ref{th:thm2_l1}, we get
\eqal{eq:delta12prime}{
\E{\Delta_{12}'} &\leq 2\alpha\tmix L^2\left(\E{r_{t-\tmix}^2} + 2\sigma^2\right) + \frac{1}{2\alpha\tmix}\E{\|\thet_t- \thet_{t-\tmix}\|^2}
\\&
\leq 2\alpha\tmix L^2\left(r_{t,2} + 2\sigma^2\right) + \alpha\tmix L^2\left(2r_{t,2} + 3\sigma^2\right)
\\&
= 4\alpha\tmix L^2r_{t,2} + 7\alpha\tmix L^2\sigma^2.
}
Now note that $\Delta_{11}'$ can be bounded using the same calculations used for $\bar{T}_1$ in~\eqref{eq:T1bar}: 
\eqal{}{
\E{\Delta_{11}'} &\leq 2\alpha(r_{t,2} + \sigma^2).
}
Next note that 
\eqal{}{
\Delta_2' &= \langle -\bar{\bfg}(\thet_{t-\tmix}) + \bfg(\xk, o_{t-\tau_t}), \xk - \xs \rangle 
\\&
= \underbrace{\langle  -\bar{\bfg}(\thet_{t-\tmix}) + \bfg(\thet_{t-\tmix}, o_{t-\tau_t}), \xk - \xs \rangle}_{\Delta_{21}'}
\\&
+ \underbrace{\langle -\bfg(\thet_{t-\tmix}, o_{t-\tau_t})  + \bfg(\xk, o_{t-\tau_t}), \xk - \xs \rangle.}_{\Delta_{22}'}
}
To bound $\E{\Delta_{22}'}$, we can proceed with calculations analogous to the ones performed to obtain the bound on $\E{\bar{T}_2}$ (see~\eqref{eq:(16)1} and~\eqref{eq:(16)2}), getting
\eqal{}{
\E{\Delta_{22}'} \leq 3\alpha\tmix L^2(r_{t,2} + \sigma^2).
}
Now, we write
\eqal{}{
\Delta_{21}' &= \underbrace{\langle  -\bar{\bfg}(\thet_{t-\tmix}) + \bar{\bfg}(\thet_{t-\tmix - \tau_t}), \xk - \xs \rangle}_{\bar{\Delta}_{1}}
\\&
+ \underbrace{\langle -\bar{\bfg}(\thet_{t-\tmix - \tau_t}) + \bfg(\thet_{t-\tmix}, o_{t-\tau_t}), \xk - \xs \rangle}_{\bar{\Delta}_{2}}.
}
We see that, as before, we can bound $\E{\bar{\Delta}_1}$ with the same procedure we used to bound $\E{\Delta_{22}'}$:
\eqal{}{
\E{\bar{\Delta}_1} \leq 3\alpha\tmax L^2(r_{t,2} + \sigma^2).
}
We write
\eqal{}{
\bar{\Delta}_2 &= \underbrace{\langle -\bar{\bfg}(\thet_{t-\tmix - \tau_t}) + \bfg(\thet_{t-\tmix - \tau_t}, o_{t-\tau_t}), \xk - \xs \rangle}_{\bar{\Delta}_{21}}
\\&
+ \underbrace{\langle -\bfg(\thet_{t-\tmix - \tau_t}, o_{t-\tau_t}) + \bfg(\thet_{t-\tmix}, o_{t-\tau_t}),\xk - \xs \rangle}_{\bar{\Delta}_{22}}.
}
Now note that $\E{\bar{\Delta}_{22}}$ can be bounded with calculations analogous to the ones performed to obtain the bound on $\bar{\Delta}_1$: 
\eqal{}{
\E{\bar{\Delta}_{22}} &\leq \E{L(\|\thet_{t-\tmix - \tau_t} - \thet_{t-\tmix}\|\|\xk - \xs\|)}
\\&
\leq 3\alpha\tmax L^2(r_{t,2} + \sigma^2).
}
Next, observe that 
\eqal{}{
\bar{\Delta}_{21} & =  \underbrace{\langle -\bar{\bfg}(\thet_{t-\tmix - \tau_t}) + \bfg(\thet_{t-\tmix - \tau_t}, o_{t-\tau_t}), \thet_{t-\tmix - \tau_t} - \xs \rangle}_{\bar{\Delta}_{211}}
\\&
+ \underbrace{\langle -\bar{\bfg}(\thet_{t-\tmix - \tau_t}) + \bfg(\thet_{t-\tmix - \tau_t}, o_{t-\tau_t}), \xk -  \thet_{t-\tmix - \tau_t} \rangle}_{\bar{\Delta}_{212}}.
}
With calculations analogous to the ones performed to obtain the bound on $\E{\Delta_{12}'}$ (see~\eqref{eq:delta12prime}), we get
\eqal{}{
\E{\bar{\Delta}_{212}} &\leq \E{\|\xk - \thet_{t-\tmix - \tau_t}\|L\left(\|\bar{\bfg}(\thet_{t-\tmix - \tau_t})\| + \|\bfg(\thet_{t-\tmix - \tau_t}, o_{t-\tau_t})\|\right)}
\\&
\leq 4\alpha(\tmix+\tmax) L^2r_{t,2} + 7\alpha(\tmix+\tmax) L^2\sigma^2.
}
Finally, note that $\bar{\Delta}_{211}$ can be bounded using the same calculations used for $\bar{T}_1$ in~\eqref{eq:T1bar}, yielding 
\eqal{}{
\E{\bar{\Delta}_{211}} \leq 2\alpha(r_{t,2} + \sigma^2).
}
So, $\E{T_3}$ can be upper-bounded by a sum of terms that are upper-bounded by either $O(\alpha)(r_{t,2}^2 + \sigma^2)$, $O(\alpha\tau_{max})(r_{t,2}^2 + \sigma^2)$, $O(\alpha\tmix)(r_{t,2}^2 + \sigma^2)$, or $O(\alpha)(\tmix + \tau_{max})(r_{t,2}^2 + \sigma^2)$. Putting all the terms together, we can get
\eqal{}{
\E{-J_{t,3}} \leq 28\alpha L^2(\tmix + \tau_{max})(r_{t,2} + \sigma^2),
}
which concludes the proof.\hspace*{\fill}~$\square$\vspace{0.4cm}

\noindent Now, recall the definition of the update rule for delayed SA with time-varying delay under Assumption~\ref{ass:delayBound}:
\eqal{ruleAppTV}{
\textbf{\textit{Delayed SA:} }\qquad\xkp = \xk + \alpha\mathbf{g}(\thet_{t-\tau_t}, o_{t-\tau_t}), \ \ \tau_t\leq \min \{t, \tau_{max}\}.}
Consider the mean squared error term $\E{r_t^2} = \E{\|\thet_t - \thet^*\|^2}$, and its expression derived in~\eqref{eq:analTauMaxApp}. Let us define ${\tau'} = 2\tau_{max} + \tau_{mix}.$ The bounds on $\E{J_{t,1}}$, $\E{J_{t,2}}$, and $\E{J_{t,3}}$ provided in the previous section are such that the update rule~\eqref{ruleAppTV} satisfies the following $\forall t\geq {\tau}' $: 
\eqal{eqmaindelayedApp}{\E{r_{t+1}^2}&=\E{\|\xkp - \xs\|^2}\\ &= \E{J_{t,1}} + \alpha^2\E{J_{t,2}} - 2\alpha \E{J_{t,3}}\\&\leq (1-2\alpha\mu)\E{r_t^2} + 98\alpha^2L^2(\tmix + \tau_{max})(r_{t,2} + \sigma^2),}
with $${r}_{t,2} = \max_{t-{\tau}'\leq l \leq t}{\E{r_l^2}}.$$
As mentioned in the outline of the analysis in Section~\ref{sec:vanillaDelayed}, the final part of the proof of Theorem~\ref{thm2} is based on a crucial argument that shows that, for a sufficiently small step size, the iterates generated by~\eqref{ruleAppTV} are uniformly bounded, which is shown in Lemma~\ref{th:thm2_l3}. Note that, starting from the above inequality \eqref{eqmaindelayedApp}, one could apply a technique similar to \cite{feyzmahdavian2014delayed} to handle the delayed optimality gaps in the bound. However, \emph{this technique would provide a suboptimal convergence rate with a $\tmax^2$ dependency on the maximum delay.} 
To prove Lemma~\ref{th:thm2_l3}, that enables the obtainment of the tight linear dependence on $\tmax$, we first provide the following result, which proves the base case of the induction proof on which the proof of Lemma~\ref{th:thm2_l3} relies on.
\begin{lemma}\label{th:baseCaseBounded}
Consider the update rule in~\eqref{ruleAppTV} and let $B = 9\sigma^2$. For $0\leq t\leq {\tau}' = 2\tmax + \tmix$ and $\alpha\leq \frac{1}{24L^2{\tau}'}$, we have 
\eqal{}{
\E{r_{t}^2}\leq B, \ \ \ 0\leq t\leq {\tau}'.
}
\begin{proof}
Note that 
\eqal{eq:plugging}{
r_{t+1}^2 &= r_t^2 +2\alpha \langle \thet_t - \xs,\bfg(\thet_{t-\tau_t}, o_{t-\tau_t}) \rangle + \alpha^2\|\bfg(\thet_{t-\tau_t}, o_{t-\tau_t})\|^2
\\&
\overset{\eqref{eq:halfSq}}{\leq }r_t^2 + \alpha r_t^2 + \alpha\|\bfg(\thet_{t-\tau_t}, o_{t-\tau_t})\|^2 + \alpha^2\|\bfg(\thet_{t-\tau_t}, o_{t-\tau_t})\|^2
\\&
\leq (1+\alpha)r_t^2 + 2\alpha\|\bfg(\thet_{t-\tau_t}, o_{t-\tau_t})\|^2
\\&
\overset{\eqref{eq:boundGsquared}}{\leq}(1+\alpha)r_t^2 + 4\alpha L^2(\|\thet_{t-\tau_t}\|^2 + \sigma^2)
\\&
\leq 
(1+\alpha)r_t^2 + 4\alpha L^2(2r_{t-\tau_t}^2 + 3\sigma^2)
\\&
= (1+\alpha)r_t^2 + 8\alpha L^2r_{t-\tau_t}^2 + 12\alpha L^2\sigma^2.
}
Taking the expectation on both sides, we get
\eqal{}{
\E{r_{t+1}^2}\leq (1 + \alpha)\E{r_t^2}+ 8\alpha L^2\E{r_{t-\tau_t}^2}+ 12\alpha L^2\sigma^2.}
Hence, we get an inequality of the following form:
\eqal{eq:formOf}{
V_{t+1}\leq pV_t + qV_{t-\tau_t} + \beta, \ \ 0 \leq \tau_t\leq \min\{t, \tau_{max}\},
}
with $V_t = \E{r_t^2}$, $p = 1+\alpha$, $q = 8\alpha L^2\tau_{max}$, and $\beta= 12\alpha L^2\sigma^2$. We define $\rho = p+q$, noting that $\rho >1$. We now prove by induction that, for all $t\geq 0$, 
\eqal{eq:statAppIndBase}{
V_t \leq \rho^tV_{0} + \epsilon_t,
}
where 
\eqal{}{
\epsilon_t = \begin{cases}
	\rho\epsilon_{t-1} + \beta \textrm{ for } t\geq 1\\
	0 \textrm{ for } t = 0
\end{cases}
}
The base case is trivially satisfied, because $V_0 \leq V_0$. As the induction hypothesis, suppose that~\eqref{eq:statAppIndBase} is true for $0\leq s\leq k$, for some $k\geq 0$, so that \eqal{}{
V_s\leq \rho^sV_0 +\epsilon_s, \ \ \ 0\leq s\leq k.
}
Now, we check the property for $k+1$, using~\eqref{eq:formOf}, and noting that $\epsilon_k$ is an increasing sequence: 
\eqal{}{
V_{k+1} &\leq pV_k + qV_{k-\tau_k} + \beta,
\\&
\leq p(\rho^kV_0 + \epsilon_k) + q(\rho^{k-\tau_k}V_0 + \epsilon_{k-\tau_k}) + \beta
\\&
\leq p(\rho^kV_0 + \epsilon_k) + q(\rho^kV_0 + \epsilon_k) + \beta
\\&
\leq (p+q)\rho^kV_0 + (p+q)\epsilon_k + \beta
\\&
= \rho^{k+1}V_0 + \epsilon_{k+1}.
}
From this, we can conclude the proof of~\eqref{eq:statAppIndBase}. Now, note that
\eqal{}{
\epsilon_t = \beta\sum_{j = 0}^{t-1}\rho^j,
}
and so we can write, for $0\leq t\leq {\tau}'$,
\eqal{}{
\rho^t\leq \rho^{{\tau}'}\leq (1+\alpha)^{{\tau}'}\leq e^{\alpha{\tau}'}\leq e^{0.25}\leq 2,
}
using the fact that $\alpha\leq \frac{1}{4{\tau}'}$. Hence, using the above results, we can get for $0\leq t\leq {\tau'}$: 
\eqal{}{
\E{r_t^2}\leq \rho^tr_0^2 + \epsilon_t\leq 2r_0^2 + \beta\sum_{j = 0}^{{\tau'}-1}\rho^j&\leq 2r_0^2 +2\beta {\tau'} = 2r_0^2 +2(12\alpha L^2\sigma^2){\tau'}
\\&
\leq 2r_0^2 + \sigma^2\leq 9\sigma^2,
}
where we used the fact that $\alpha\leq \frac{1}{24L^2{\tau}'}$ and that $r_0^2 = \|\thet_0 - \xs\|^2\leq 2\|\thet_0\|^2 + 2\|\xs\|^2\leq 4\sigma^2$.

We have just shown that, for $0\leq t\leq {\tau}'$, it holds
\eqal{}{
\E{r_t^2}\leq 9\sigma^2.
}
\end{proof}
\end{lemma}
\noindent Now, we provide the proof of Lemma~\ref{th:thm2_l3}, which relies on Lemma~\ref{th:baseCaseBounded}.\vspace{0.4cm}

\noindent\textbf{Proof of Lemma~\ref{th:thm2_l3}.}
We know from Lemma~\ref{th:baseCaseBounded} that for $t = 0, ..., {\tau}'$, with ${\tau}' = 2\tau_{max} + \tmix$, and $\alpha \leq \frac{1}{24L^2{\tau}'}$, we have 
\eqal{eq:inductionBaseproof}{
\E{r_t^2} \leq B.
}
We now proceed by induction to show that the bound holds true also for any $t\geq {\tau}'$, thus for all $t\geq 0$. We use~\eqref{eq:inductionBaseproof} as the base case for the induction proof. Fix any $ t \geq \tau'$, and as the induction hypothesis, assume that the property is true for all $s$ satisfying  ${\tau}'\leq s\leq t$, i.e., 
\eqal{}{
\E{r_s^2}\leq B \ \ \forall {\tau'}\leq s \leq t.
}

Now, from~\eqref{eqmaindelayedApp} we can write
\eqal{}{
\E{r_{t+1}^2} \leq  (1 - 2\alpha{\mu})\E{r_t^2}
+98\alpha^2L^2(\tmix + \tau_{max})({r_{t,2}} + \sigma^2).
}
Observe that from the induction hypothesis and the induction base case, the following is true: 
\eqal{}{
r_{t,2} = \max_{t-{\tau}'\leq l \leq t}{\E{r_l^2}} \leq B.
}
Hence, we can write, recalling that $B= 9\sigma^2 \geq \sigma^2$,
\eqal{}{
\E{r_{t+1}^2} &\leq  (1 - 2\alpha{\mu})\E{r_t^2}
+98\alpha^2L^2(\tmix + \tau_{max})({r_{t,2}} + \sigma^2)
\\&
\leq (1 - 2\alpha{\mu})B + 98\alpha^2L^2(\tmix + \tau_{max})(B + \sigma^2)
\\&
\leq (1 - 2\alpha{\mu})B + 2B98\alpha^2L^2(\tmix + \tau_{max})
\\&
\leq (1 - 2\alpha{\mu} + 196\alpha^2L^2(\tmix + \tau_{max}))B.
}
Thus, for $\alpha \leq \frac{\mu}{196L^2\bar{\tau}}\leq \frac{\mu}{98L^2{\tau}'} \leq \frac{\mu}{98L^2(\tmix + \tau_{max})}$, we get $1 - 2\alpha{\mu} + 196\alpha^2L^2(\tmix + \tau_{max}) \leq 1$, implying 
\eqal{}{
\E{r_{t+1}^2}\leq B.
}
This concludes the proof. \hspace*{\fill}~$\square$\vspace{0.4cm}

\noindent Using this last Lemma in conjunction with~\eqref{eqmaindelayedApp}, we can now prove Theorem~\ref{thm2}.\vspace{0.4cm}

\subsubsection{Proof of Theorem~\ref{thm2}} Note that, from~\eqref{eqmaindelayedApp}, we can write, for $T \geq \tau' = 2\tmax + \tmix$,
\eqal{}{
\E{r_{t+1}^2} &\leq (1 - 2\alpha{\mu})\E{r_t^2}
+98\alpha^2L^2(\tmix + \tau_{max})({r_{t,2}} + \sigma^2)
\\&
\overset{(*)}\leq (1 - 2\alpha{\mu})\E{r_t^2}
+2B98\alpha^2L^2(\tmix + \tau_{max}),
}
where for $(*)$ we used the fact that, for $\alpha \leq \frac{1}{196L^2\bar{\tau}}$, it holds $r_{t,2} \leq B = 9\sigma^2$, as established by Lemma~\ref{th:thm2_l3}. Iterating the inequality, we get
\eqal{}{
\E{r_{t+1}^2} &\leq (1-2\alpha\mu)^{t+1-{\tau'}}r_{{\tau'}}^2 + 98L^2\alpha^2(\tmix + \tau_{max})2B\sum_{j = 0}^{\infty}(1-2\alpha\mu)^j
\\&
\leq (1-2\alpha\mu)^{t+1-{\tau'}}r_{{\tau'}}^2 + \frac{98L^2\alpha(\tmix + \tau_{max})B}{\mu}
\\&
\leq (1-2\alpha\mu)^{t+1-{\tau'}}B + \frac{98L^2\alpha(\tmix + \tau_{max})B}{\mu}
\\&
\overset{(*)}{\leq} (1-2\alpha\mu)^{t+1}2B + \frac{98L^2\alpha(\tmix + \tau_{max})B}{\mu}
\\&
\leq e^{-2\alpha\mu(t+1)}2B + \frac{98L^2\alpha(\tmix + \tau_{max})B}{\mu}
,
}
where $(*)$ follows because  
\eqal{}{
(1-2\alpha\mu)^{-{\tau'}}\leq e^{2\alpha\mu{\tau'}}\leq e^{0.25}\leq 2,
}
and by noting that $\alpha\mu\leq \alpha\leq \frac{1}{196L^2\bar{\tau}}\leq \frac{1}{8{\tau'}}$. Hence, for $\alpha\leq \frac{1}{CL^2\bar{\tau}}$, with $C = 196$, we get the result. Setting $\alpha= \frac{1}{CL^2\bar{\tau}}$, with $C\geq 196$ and $C' = 98$, we can also get
\eqal{}{
\E{r_T^2}\leq \exp\left({-\frac{2\mu^2T}{CL^2\bar{\tau}}}\right)2B + \frac{C'B}{C}.
}\hspace*{\fill}~$\square$\vspace{0.4cm}

\newpage

\subsection{Proof of Theorem \ref{thm:picky}}\label{sec:proof of thm:picky}
For a precision threshold $\epsilon >0$ to be specified soon, recall that the update rule of delay-adaptive SA takes the form: 
\begin{align}\label{eq:algoApp}
\thet_{t+1}=
\begin{cases}
\thet_{t}+\alpha\left( \mathbf{g}(\thet_{t-\tau_t},o_{t-\tau_t})\right) \hspace{1cm} &\textrm{ if } \|\thet_{t}-\thet_{t-\tau_t}\|\leq {\epsilon} \\
\thet_{t} \hspace{1cm} &\textrm{otherwise}.
\end{cases}
\end{align}
Let us define $$\beta\triangleq\epsilon + \sigma, \hspace{2mm} \textrm{and} \hspace{2mm} c \triangleq L\beta,$$ where $L>0$ is the Lipschitz constant defined in Assumption \ref{ass:Lipschitz}.

\noindent We use the following way of writing the update rule:
\eqal{}{
\thet_{t+1} = \thet_t + \alpha\mathbf{g}(\thet_t, o_t) - \alpha\mathbf{e}_t,
}
where $\mathbf{e}_t = \mathbf{g}(\thet_t, o_t) - \mathbf{g}(\thet_{t-\tau_t}, o_{t-\tau_t})$. We also define the following indicator function $I_t$:
\eqal{}{
I_t=
\begin{cases}
1 \hspace{1cm} &\textrm{ if } \|\thet_{t}-\thet_{t-\tau_t}\|\leq {\epsilon} \\
0 \hspace{1cm} &\textrm{otherwise}.
\end{cases}
}

\subsubsection{Outline of the Analysis}
Recall that $I_t=1$ if $\|\thet_{t}-\thet_{t-\tau_t}\|\leq {\epsilon}$ and $I_t=0$ otherwise. Then, we can express  $\|\xkp - \xs\|^2$ as
\begin{equation}\label{eq:analTauavg}
	\begin{aligned}
		\|\xkp - \xs\|^2 = I_t\left(K_{t,1} + \alpha^2K_{t,2} - 2\alpha K_{t,3}\right)+(1-I_t)\|\xk - \xs\|^2, 
	\end{aligned}
\end{equation}
with 
\begin{equation}
\begin{aligned}
	K_{t,1} &\triangleq \|\xk - \xs + \alpha\mathbf{g}(\xk, o_t)\|^2,\\
	K_{t,2} &\triangleq \|\mathbf{e}_t\|^2,\\
	K_{t,3} &\triangleq \langle \mathbf{e}_t, \xk - \xs + \alpha\mathbf{g}(\xk, o_t) \rangle.
\end{aligned}
\end{equation}
In order to analyze the convergence of the delay-adaptive SA update rule, we first derive bounds for $K_{t,1}$, $K_{t,2}$, and $-K_{t,3}$ for iterations in which $I_t=1$. Then, we establish a lower bound on the number of iterations in which $I_t=1$. By using these two results, we are able to obtain the finite-time convergence rate of Theorem~\ref{thm:picky}.

\subsubsection{Auxiliary Lemmas} Similarly to Section~\ref{sec:vanillaDelayed}, we provide three lemmas that are instrumental to proving the main result of this section, i.e., Theorem~\ref{thm:picky}. As in the case of the vanilla delayed SA update, we start by providing a result that provides a bound on the norm of $\xk - \xktaua$. Note that here, unlike the non-adaptive vanilla case of Section~\ref{sec:vanillaDelayed}, we are able to get a bound on $\|\xk - \xktaua\|$ which is a function only of the \emph{current} iterate $\xk$. This aspect significantly simplifies the proof. Note that this step, which plays a crucial role in enabling us to remove the dependency of the rate on the maximum delay, is provided by the specific adaptive update rule, which only uses SA update directions $\mathbf{g}({\thet_{t-\tau_t}, o_{t-\tau_t}})$ when the distance between the delayed iterate $\thet_{t-\tau_t}$ - at which the direction was computed - and the current approximation parameter $\xk$ is smaller than the threshold $\epsilon$, with $\epsilon$ set to $\alpha$. In the proof of Theorem~\ref{thm:picky} and related lemmas, without loss of generality, we also use negative indices for iterates and observations. In particular, we define, for $j < 0$, $\thet_j = \thet_0$ and $o_{j}= o_{0}$.

\begin{lemma}\label{lemma:lemmaSrikantOurs}
    For any $\tau \geq 1$ and $t\geq 0$, we have
    \begin{align}
        \|\thet_t-\thet_{t-\tau}\|\leq 4L\alpha \tau(r_t+2\beta),
    \end{align}
    where $\beta=L\epsilon + L\sigma$.
    \end{lemma}

Using Lemma~\ref{lemma:lemmaSrikantOurs}, we can bound $K_{t,1}$, $K_{t,2}$, and $K_{t,3}$ in Lemma~\ref{lemma:tavg}.

\begin{lemma}\label{lemma:tavg}
For $t \geq \tmix$, if $I_t=1$, we have, for some absolute constants $C_{11}, C_{12}\geq 1$,
\begin{alignat*}{3}
(i)&\hspace{0.5cm}
\E{K_{t,1}}&&\leq \left(1 - 2\alpha\mu+C_{11}L^2(\alpha^2\tmix)\right)\E{r_t^2} + C_{11}L^2(\alpha^2\tmix) \beta^2
\\
(ii)&\hspace{0.5cm}\E{K_{t,2}}&&\leq 2L^2 \tau_{mix}(\E{r_{t}^2} + \beta^2)
\\
(iii)&\hspace{0.5cm}\E{-K_{t,3}}&&\leq C_{12}L^2\alpha\tau_{mix}( \E{r_{t}^2} + \beta^2). 
\end{alignat*}
\end{lemma}

With the help of the previous lemma, we can rewrite Equation~\eqref{eq:analTauavg} as follows, when $I_t=1$:
\eqal{eqmaindelayed}{\E{r_{t+1}^2}&=\E{\|\xkp - \xs\|^2}\\ &= \E{K_{t,1}} + \alpha^2\E{K_{t,2}} - 2\alpha \E{K_{t,3}}\\&\leq \left(1 - 2\alpha\mu+C_{11}L^2(\alpha^2\tmix)\right)\E{r_t^2} + C_1L^2(\alpha^2\tmix) \beta^2.}
To finish the proof, we only need to determine the number of times we update the parameter in $T$ iterations. This can be achieved through the following lemma, which we borrow from~\cite{cohen2021asynchronous}. By utilizing this lemma, we can complete the proof and obtain the bound presented in Theorem~\ref{thm:picky}.
\begin{lemma}\label{th:lemmaAtleast}
Let $\tau_{avg}$ be the average delay, i.e., $\tau_{avg}=\frac{1}{T}\sum\limits_{t=1}^{T}\tau_{t}$. The number of updates that the delay-adaptive SA algorithm makes is at least ${\frac{T}{4\tau_{avg}+4}}$.
\end{lemma}

In the remainder of the section, we first provide the proofs for all the auxiliary lemmas, and then conclude with the proof of Theorem~\ref{thm:picky}.

\subsubsection{Proofs of Auxiliary Lemmas}
\vspace{0.4cm}
\noindent\textbf{Proof of Lemma~\ref{lemma:lemmaSrikantOurs}.}
Let $t^'=t-\tau_t$, then
    \eqal{eq:bound207}{
        \|\thet_{t+1}-\thet_{t}\|&\leq \alpha I_t \|\mathbf{g}(\thet_{t^'},o_{t^'})\|
        \\&\leq \alpha LI_t (\|\thet_{t'}\| + \sigma)
        \\&
        \leq \alpha L I_t(\|\xk\| + \|\xk - \thet_{t'}\| + \sigma)
        \\&
        \leq \alpha L I_t(\|\xk\| + \epsilon + \sigma)
        \\&
        \leq \alpha L\left(\|\xk\| + \beta\right)
        ,
    }
    from which we get 
    \eqal{}{
    \|\thet_{t+1}\|\leq (1 + \alpha L)\|\thet_t\| + \alpha L\beta. 
    }
    In what follows, we will recursively use the above inequality in tandem with the following facts: (i) If $I_t=0$, then $\thet_t=\thet_{t-1}$ and our next steps hold trivially, and that (ii) $\thet_j = \thet_0$ for $j<0$. We then have that $\forall t \geq 0$: 
    \eqal{eq:bound209}{
        \|\thet_t\|&\leq (1+\alpha L)\|\thet_{t-1}\|+\alpha L\beta
        \\&
        \leq(1+ \alpha L)^2\|\thet_{t-2}\| + \alpha L\beta(1+\alpha L) + \alpha L\beta
        \\&
        \leq (1+\alpha L)^{\tau}\|\thet_{t-\tau}\| + \alpha L\beta\sum_{j = 0}^{\tau-1}(1+\alpha L)^j
        \\&
        \leq 2\|\thet_{t-\tau}\| + 2\alpha L\beta\tau,
    }
    where we used $\alpha\tau L \leq \frac{1}{4}$ and the fact that, for $0\leq j\leq \tau$, $(1+x)^j\leq(1+x)^{\tau}\leq e^{x\tau}\leq 2$ for $x\tau \leq \frac{1}{4}$. \newline
    Now, using the above inequality on $\|\thet_t\|$,
    \begin{equation}
    \begin{aligned}
        \|\thet_{t}-\thet_{t-\tau}\|&\leq \sum_{j=t-\tau}^{t-1}\|\thet_{j+1}-\thet_j\|\\
        &\overset{\eqref{eq:bound207}}\leq \sum_{j=t-\tau}^{t-1} \alpha L(\|\thet_j\| +\beta)
        \\
        &\overset{\eqref{eq:bound209}}\leq\alpha L\sum_{j=t-\tau}^{t-1} (2\|\thet_{t-\tau}\| + 2\alpha L\beta\tau) + \alpha\tau L\beta 
        \\
        &\leq \alpha\tau L (2\|\thet_{t-\tau}\|+2L\alpha \beta \tau)+\alpha L\beta\tau\\
        &\leq 2\alpha L\tau \|\thet_{t-\tau}\|+2L\alpha \beta\tau,
    \end{aligned}
    \end{equation}
    where in the last inequality we used the fact that $\alpha \tau \leq \frac{1}{4L}$.
Moreover, 
\eqal{}{
        \|\thet_{t}-\thet_{t-\tau}\|&\leq 2\alpha \tau L\|\thet_{t-\tau}\|+2L\alpha \beta\tau
        \\&
        \leq 2\alpha \tau L\|\thet_t\|+2L\alpha \beta\tau + 2\alpha \tau L\|\thet_{t}-\thet_{t-\tau}\|
        \\&
        \leq 2\alpha \tau L\|\thet_t\|+2L\alpha \beta\tau + \frac{1}{2} \|\thet_{t}-\thet_{t-\tau}\|,
}
which results in 
 \begin{align}
\|\thet_{t}-\thet_{t-\tau}\|\leq 4\alpha \tau L\|\thet_t\|+4L\alpha\beta\tau.
\end{align}
Recalling $c=L\beta,$ we can write 
\eqal{}{\label{eq:delta1}
\|\thet_{t}-\thet_{t-\tau}\|&\leq 4\alpha \tau L\|\thet_t\|+4\alpha c\tau
\\&
\leq 4\alpha \tau L(r_t + 2\beta).
}
Also note that (recall $c = L\beta$): 
\eqal{eq:4delta}{
\|\thet_{t}-\thet_{t-\tau}\|^2&\leq (4\alpha \tau L\|\thet_t\|+4\alpha c\tau)^2
\\&
\leq(4\alpha \tau Lr_t+8\alpha c\tau)^2
\\&
\leq 16(\alpha \tau Lr_t+2\alpha c\tau)^2
\\&
\overset{\eqref{eq:sumSquared}}\leq 32(\alpha^2\tau^2L^2r_t^2 + 4\alpha^2\tau^2 c^2)
\\&
\leq 32\alpha^2\tau^2L^2(r_t^2 + 4\beta^2). 
}
\hspace*{\fill}~$\square$\vspace{0.4cm}

Note that for an iteration $t$ in which an update is not made, i.e., $I_t = 0$, we have 
\eqal{}{
\|\thet_{t+1}-\thet^*\|^2 = \|\thet_{t}-\thet^*\|^2.
}
If $I_t=1$, we have:
\eqal{}{
    \|\thet_{t+1}-\thet^*\|^2=\underbrace{\|\thet_{t}+\alpha \mathbf{g}(\thet_{t},o_{t})- \thet^*\|^2}_{K_{t,1}}+\underbrace{\alpha^2 \|\mathbf{e}_t\|^2 }_{K_{t,2}}\underbrace{-2\alpha \langle \mathbf{e}_t, \thet_t-\thet^*+\alpha  {\mathbf{g}(\thet_{t},o_{t})}\rangle.}_{K_{t,3}}
}
\noindent\textbf{Proof of Lemma~\ref{lemma:tavg} - (i).} In the following, we drop the dependence on the iteration $t$ in the terms we bound. We write 
\begin{equation}
\begin{aligned}
	K_{t,1} = \|\xk - \xs + \alpha \mathbf{g}(\xk, o_t)\|^2 &= r_t^2 + 2\alpha\underbrace{\langle \thet_t - \xs,\mathbf{g}(\thet_t, o_t) \rangle}_{I_1}
	\\& + \alpha^2\underbrace{\|\mathbf{g}(\thet_t, o_t)\|^2}_{I_2}.
\end{aligned}
\end{equation}
Note that 
\begin{equation}
\begin{aligned}
\E{I_2} = \E{\|\mathbf{g}(\thet_t, o_t)\|^2} 
&\overset{\eqref{eq:boundGsquared}}\leq 2L^2\left( \E{\|\thet_t\|^2} + \sigma^2\right)
&\leq 2L^2 \left(2\E{r_t^2} + 3\sigma^2\right).
\end{aligned}
\end{equation}
Next, note that
\begin{equation}\label{eq:defTprime}
\begin{aligned}
I_1 &= \langle \thet_t  - \xs, \mathbf{g}(\thet_t, o_t)\rangle = -\langle \xs - \thet_t, \barg(\thet_t)\rangle\\& + \langle \thet_t  - \xs, \mathbf{g}(\thet_t, o_t)-\barg(\thet_t)\rangle \\&
\leq -\omega(1-\gamma)r_{t}^2 + \underbrace{\langle \xk  - \xs, \mathbf{g}(\xk, o_t)-\barg(\xk)\rangle}_{T_1'}.
\end{aligned}
\end{equation}
Now,
\begin{equation}
\begin{aligned}
T_1' = \underbrace{\langle \xk - \thet_{t-\tmix},  \mathbf{g}(\xk, o_t)-\barg(\xk) \rangle}_{T_{11}'} + \underbrace{\langle \thet_{t-\tmix} - \xs,  \mathbf{g}(\xk, o_t)-\barg(\xk) \rangle}_{T_{12}'}.
\end{aligned}
\end{equation}
Using the fact that $\beta \leq\sigma$ and $L\geq1$, we can bound $T_{11}'$ as follows:
\begin{equation}\label{eq:deltaThet}
\begin{aligned}
T_{11}'&\leq \|\thet_t-\thet_{t-\tau_{mix}}\|\|\bfg(\thet_{t},o_{t})-\barg(\thet_t)\| \\
&\overset{\eqref{eq:boundGradNorm}}\leq \|\thet_t-\thet_{t-\tau_{mix}}\|2L(\|\thet_t\|+\sigma)
\\&
\overset{\eqref{eq:4delta}}\leq (4\alpha \tau_{mix} L\|\thet_{t}\|+4 \alpha \tau_{mix} c)2L(\|\thet_t\|+\sigma)
\\& 
\leq 8\alpha \tau_{mix}L^2 \|\thet_t\|^2+8\alpha c \tau_{mix}L \sigma+(8\alpha\tau_{mix}L^2\sigma+8\alpha \tau_{mix}L c)\|\thet_t\|
\\&
\leq 8\alpha \tau_{mix}L^2 \|\thet_t\|^2+8\alpha L^2\beta \tau_{mix} \sigma+(8\alpha\tau_{mix}L^2\sigma+8\alpha \tau_{mix} L^2\beta)\|\thet_t\|
\\&
\leq 8\alpha \tau_{mix}L^2 \|\thet_t\|^2+8\alpha L^2 \tau_{mix}\beta^2+32\alpha\tau_{mix}L^2\beta^2+32\alpha \tau_{mix} L^2\|\thet_t\|^2
\\&
= 40\alpha\tmix L^2\|\thet_t\|^2 + 40\alpha\tmix L^2\beta^2
\\&
\leq 80\alpha\tmix L^2r_t^2 + 120\alpha\tmix L^2\beta^2
\\&
\leq 40L^2 \alpha \tau_{mix}(2r_t^2+3\beta^2).
\end{aligned}
\end{equation}
Taking expectations on both sides of the above display, we obtain
\begin{equation}\label{eq:expect_t11}
	\E{T_{11}'}\leq 40L^2\alpha\tmix(2\E{r_{t}^2} + 3\beta^2).
\end{equation}
Now, observe that 
\begin{equation}
\begin{aligned}
T_{12}' &= \langle \thet_{t-\tmix} - \xs,  \mathbf{g}(\xk, o_t)-\barg(\xk)\rangle\\&
 = \bar{T}_1 + \bar{T}_2 + \bar{T}_3,
\end{aligned}
\end{equation}
with
\begin{equation}
\begin{aligned}
\bar{T}_1 &= \langle \thet_{t-\tmix} - \xs,  \mathbf{g}(\thet_{t-\tmix}, o_t)-\barg(\thet_{t-\tau_{mix}})\rangle
\\
\bar{T}_2 &= \langle \thet_{t-\tmix} - \xs,  \mathbf{g}(\thet_{t}, o_t)-{g}(\thet_{t-\tmix}, o_t)\rangle
\\
\bar{T}_3 &= \langle \thet_{t-\tmix} - \xs,  \barg(\thet_{t-\tmix})-\barg(\thet_{t})\rangle.
\end{aligned}
\end{equation}
Note that, from \eqref{eq:4delta}, we have, recalling $c = L\beta$, 
\eqal{}{
\|\thet_t - \thet_{t-\tmix}\|^2\leq 32\alpha^2\tmix^2L^2\left(r_t^2 + 4\beta^2\right).
}
We now proceed to bound $\bar{T}_2$ and $\bar{T}_3$. Note that, using the Lipschitz property (Assumption \ref{ass:Lipschitz}) and the fact that $\alpha\leq \frac{1}{12\tmix L^2}$, we obtain 
\begin{equation}
\begin{aligned}
\bar{T}_2 &\leq \|\thet_{t-\tmix} - \xs\|\|\mathbf{g}(\xk, o_t) - \mathbf{g}(\thet_{t-\tmix}, o_t)\|
\\&
\leq L\|\thet_{t-\tmix} - \xs\|\|\thet_{t-\tmix} - \xk\|
\\&
\overset{\eqref{eq:halfSq}}\leq \frac{\alpha\tmix L^2}{2}\|\thet_{t-\tmix} - \thet^*\|^2 + \frac{1}{2\alpha\tmix}\|\thet_{t-\tmix} - \thet_t\|^2
\\&
\leq \alpha\tmix L^2\|\thet_t-\thet^*\|^2 + \alpha\tmix L^2\|\thet_{t-\tmix} - \thet_t\|^2 + \frac{1}{2\alpha\tmix}\|\thet_{t-\tmix} - \thet_t\|^2
\\&
\leq \alpha\tmix L^2r_t^2 + \alpha\tmix L^2\left( 32\alpha^2\tmix^2L^2\left(r_t^2 + 4\beta^2\right)\right)
\\&
+\frac{1}{2\alpha\tmix}\left( 32\alpha^2\tmix^2L^2\left(r_t^2 + 4\beta^2\right)\right)
\\&
\leq \alpha\tmix L^2r_t^2 + \alpha\tmix L^2(r_t^2 +\beta^2) + 16\alpha\tmix L^2r_t^2 + 64\alpha\tmix L^2\beta^2
\\&
\leq \alpha\tmix L^2(18r_t^2 + 65\beta^2). 
\end{aligned}
\end{equation}

Taking the expectation, we get
\begin{equation}
	\E{\bar{T}_2}\leq \alpha\tmix L^2(18\E{r_{t}^2} + 65\beta^2).
\end{equation}
With the same calculations, we can get 
\begin{equation}
	\E{\bar{T}_3}\leq \alpha\tmix L^2(18\E{r_{t}^2} + 65\beta^2).
\end{equation}
We now proceed to bound $\E{\bar{T}_1}$.
Using the fact that $\alpha\tmix \leq \frac{1}{12 L^2}$, we can write
\begin{equation}
\begin{aligned}
\E{\bar{T}_1} &= \E{\langle \thet_{t-\tmix} - \xs, \mathbf{g}(\thet_{t-\tmix}, o_t) - \barg(\thet_{t-\tmix}) \rangle}
\\&
= \E{\langle \thet_{t-\tmix} - \xs, \E{\mathbf{g}(\thet_{t-\tmix}, o_t)|o_{t-\tau_{mix}}, \thet_{t-\tmix}} - \barg(\thet_{t-\tmix}) \rangle}
\\&
\leq \E{\|\thet_{t-\tmix} - \xs\| \|\E{\mathbf{g}(\thet_{t-\tmix}, o_t)|o_{t-\tmix}, \thet_{t-\tmix}} - \barg(\thet_{t-\tmix})\|}
\\&
\overset{\eqref{eq:mixingEq}}\leq \alpha\E{\|\thet_{t-\tmix} - \xs\|(\|\thet_{t-\tmix}\| + \sigma)}
\\&
\overset{\eqref{eq:halfSq}}\leq \frac{\alpha}{2}\E{\|\thet_{t-\tmix} - \xs\|^2 + \left(\|\thet_{t-\tmix} - \xs\| + 2\sigma\right)^2}
\\&
\leq \frac{\alpha}{2}\E{2r_t^2 + 2\|\thet_t - \thet_{t-\tmix}\|^2 + 2\|\thet_{t-\tmix} - \xs\|^2 + 8\sigma^2}
\\&
\leq \frac{\alpha}{2} \E{2r_t^2 + 2\|\thet_t - \thet_{t-\tmix}\|^2 + 2\left(2\|\thet_t - \thet_{t-\tmix}\|^2 + 2\|\thet_{t} - \xs\|^2\right) + 8\sigma^2}
\\&
\leq 3\alpha\E{r_t^2} + 3\alpha\E{\|\thet_{t} - \thet_{t-\tmix}\|^2} + 4\alpha\sigma^2
\\&
\leq 3\alpha\E{r_t^2} + 96\alpha^3\tmix^2\left(\E{r_t^2} + 4 c^2\right) + 4\alpha\sigma^2
\\&
\leq 4\alpha\E{r_t^2} + 8\alpha c^2
.
\end{aligned}
\end{equation}
So, we obtain 
\eqal{}{
\E{T_{12}'} &= \E{\bar{T}_1} + \E{\bar{T}_2} + \E{\bar{T}_3} 
\\& 
\leq 4\alpha\E{r_t^2} + 8\alpha c^2+2\left(\alpha\tmix L^2\left(18\E{r_{t}^2} + 65\beta^2\right)\right)
\\&
\leq 46\alpha\tmix L^2\left(\E{r_{t}^2} + 3\beta^2\right).
}
Combining the bounds on $\E{T_{11}'}$ and $\E{T_{12}'}$, we obtain
\eqal{eq:T1prime}{
\E{T_1'} &= \E{T_{11}'} + \E{T_{12}'}
\\&
\leq 40\alpha\tmix L^2(2\E{r_{t}^2} + 3\beta^2) +46\alpha\tmix L^2\left(\E{r_{t}^2} + 3\beta^2\right)
\\&
\leq 130\alpha\tmix L^2(\E{r_{t}^2} + 2\beta^2).
} This immediately yields: 
\eqal{}{
\E{I_1} &\leq -\omega(1-\gamma)\E{r_{t}^2} + \E{T_1'}
\\&
\leq -\omega(1-\gamma)\E{r_{t}^2} + 130\alpha\tmix L^2(\E{r_{t}^2} + 2\beta^2)
. 
}
Hence, 
\begin{equation}
\begin{aligned}
\E{K_{t,1}} &= \E{r_t^2} + 2\alpha\E{I_1} + \alpha^2\E{I_2}
\\&
\leq \left(1- 2\alpha\mu + 264\alpha^2\tmix L^2\right)\E{r_t^2} + 526\alpha^2\tmix L^2\beta^2.
\end{aligned}
\end{equation}
\hspace*{\fill}~$\square$\vspace{0.4cm}\noindent\\
\noindent\textbf{Proof of Lemma~\ref{lemma:tavg} - (ii).}Using the notation $t^'=t-\tau_t$, we get
\eqal{eq:etApp6}{
        \|\mathbf{e}_t\|=&\leq \|\mathbf{g}(\thet_{t^'},o_{t^'})-\mathbf{g}(\thet_{t},o_{t^'})\|+  \|\mathbf{g}(\thet_{t},o_{t^'})-\mathbf{g}(\thet_{t},o_{t})\|
        \\&\leq L\|\thet_{t}-\thet_{t^'}\|+2L( \|\thet_t\|+\sigma)
        \\&
        \leq 2L\|\thet_t\| + {L\epsilon + 2L\sigma}
        \\&
        \leq 2L\|\thet_t\| + {2L(\epsilon + \sigma)}
        \\&
        = 2(L\|\thet_t\| + c).
} Note that, using the above inequality, we have
\eqal{}{
\|\mathbf{e}_t\|\leq& L\|\thet_t\| + 2c.
}
Using this inequality, we can get
\eqal{eq:etAppEq6}{
\E{\alpha^2K_{t,2}}&\leq \alpha^2\E{\|\mathbf{e}_t\|^2}
\\&
\leq \alpha^2\E{\left(L\|\thet_t\| + 2c\right)^2}
\\&
\leq \alpha^2\E{\left(2L^2\|\thet_t\|^2 + 8c^2\right)}
\\&
\leq 4\alpha^2L^2\E{r_t^2} + 12\alpha^2c^2. 
}\hspace*{\fill}~$\square$\vspace{0.4cm}\noindent\\
\noindent\textbf{Proof of Lemma~\ref{lemma:tavg} - (iii).} 
We write 
\begin{equation}
\begin{aligned}
-K_{t,3} &= \langle -\et, \xk - \xs + \alpha\bfgofxo \rangle
\\& 
=  \underbrace{\langle -\et, \xk - \xs \rangle}_{\Delta} + \underbrace{\alpha\langle -\et, \bfgofxo \rangle}_{\bar{\Delta}}.
\end{aligned}
\end{equation}
Note that using \eqref{eq:etAppEq6}, we have 
\begin{equation}
    \|\et\|^2\leq 4L^2{r_t^2} + 12c^2.
\end{equation}
Using this fact, we can get
\eqal{}{
\bar{\Delta} &= \alpha \langle -\et, \bfgofxo \rangle 
\\&
\leq \alpha\|\et\|\|\bfgofxo\|
\\&
\overset{\eqref{eq:halfSq}}\leq \frac{\alpha}{2}\left(\|\et\|^2+\|\bfgofxo\|^2\right)
\\&
\overset{\eqref{eq:boundGsquared}}\leq \frac{\alpha}{2}\left(4L^2{r_t^2} + 12c^2 + 2L^2(2r_t^2 + 3\sigma^2)\right) 
\\&
\leq\frac{\alpha}{2}(8L^2r_t^2 + 18c^2)
\\&
\leq 4\alpha L^2r_t^2 + 9\alpha c^2,
}
implying that 
\eqal{}{
\E{\bar{\Delta}} \leq \alpha L^2(4\E{r_{t}^2} + 9\beta^2).
}
We now proceed to bound $\Delta$ as follows: 
\eqal{}{
\Delta &= \langle -\et, \xk - \xs \rangle = \langle -\bfgofxo + \bfg(\thet_{t-\tau_t}, o_{t-\tau_t}), \xk - \xs\rangle
\\&
= \underbrace{\langle -\bfgofxo + \barg(\xk)-\barg(\xk)+\bfg(\xk, o_{t-\tau_t}), \xk - \xs \rangle}_{\Delta_1}
\\&
+ \underbrace{\langle -\bfg(\xk, o_{t-\tau_t}) + \bfg(\thet_{t-\tau_t}, o_{t-\tau_t}), \xk -\xs\rangle}_{\Delta_2}.
}
Next, noting that $\|\bfg(\xk, o_{t-\tau_t}) - \bfg(\thet_{t-\tau_t}, o_{t-\tau_t})\|\leq L\|\xk-\thet_{t-\tau_t}\|\leq L\epsilon$, we have

\eqal{}{
\Delta_2\leq L\epsilon r_t \leq \frac{\alpha L}{2}\left( r_t^2+\left(\frac{\epsilon}{\alpha}\right)^2\right).
}
Now, using the fact that $\epsilon\leq \alpha$ and $1\leq \sigma\leq \beta$, we have

\eqal{}{
\Delta_2\leq \frac{\alpha L}{2}( r_t^2+\beta^2),
}
and 

\eqal{}{
\Delta_1 &\leq \underbrace{\langle -\bfgofxo + \barg(\xk), \xk - \xs \rangle}_{\Delta_{11}}+\underbrace{\langle -\barg(\xk)+\bfg(\xk, o_{t-\tau_t}), \xk - \xs \rangle}_{\Delta_{12}}.
}
Note that $\Delta_{11} = T_1'$ for $ T_1'$ defined above (see \eqref{eq:defTprime}), and therefore can be bounded accordingly as in \eqref{eq:T1prime}: 
\eqal{}{
\E{\Delta_{11}}\leq 130\alpha\tmix L^2(\E{r_{t}^2} + 2\beta^2).
}
We now bound $\E{\Delta_{12}}$ as follows: 
\eqal{}{
\Delta_{12} &= \langle -\bar{\bfg}(\thet_{t}) + \bfg(\xk, o_{t-\tau_t}), \xk - \xs \rangle 
\\&
= \underbrace{\langle -\bar{\bfg}(\thet_{t-\tau_t}) + \bfg(\xk, o_{t-\tau_t}), \xk - \xs \rangle}_{\Delta_1'} 
\\&
+ \underbrace{\langle -\barg(\xk) +\bar{\bfg}(\thet_{t-\tau_t}), \xk - \xs \rangle}_{\Delta_2'}. 
}
Note that using the fact that $\epsilon \leq \alpha$, we have
\eqal{eq:LipschitzUsedEps}{
\Delta_2' &\leq L\|\xk - \thet_{t-\tau_t}\|\|\xk - \xs\|
\\&
\leq L(\epsilon r_t)
\\&
\leq L\left(\alpha r_t^2 + \frac{\epsilon^2}{\alpha}\right)
\\&
\leq L\alpha(r_t^2+ \beta^2)
.
}
We now bound $\E{\Delta_1'}$ as follows: 
\eqal{}{
\Delta_1'&=\langle -\bar{\bfg}(\thet_{t-\tau_t}) + \bfg(\xk, o_{t-\tau_t}), \xk - \xs \rangle
\\&
= \underbrace{\langle  -\bar{\bfg}(\thet_{t-\tau_t}) + \bfg(\thet_{t-\tau_t}, o_{t-\tau_t}), \xk - \xs \rangle}_{\Delta_{11}'}
\\&
+ \underbrace{\langle -\bfg(\thet_{t-\tau_t}, o_{t-\tau_t})  + \bfg(\xk, o_{t-\tau_t}), \xk - \xs \rangle}_{\Delta_{12}'}.
}
Note that, using Lipschitzness (Assumption \ref{ass:Lipschitz}) and the same calculations used to arrive at \eqref{eq:LipschitzUsedEps}, one can obtain 
\eqal{}{
\Delta_{12}' &\leq \|\bfg(\thet_{t-\tau_t}, o_{t-\tau_t})  - \bfg(\xk, o_{t-\tau_t})\|\|\xk - \xs\|
\\&
\leq L\|\thet_{t-\tau_t} - \xk\|\|\xk - \xs\|
\\&
\leq L\alpha(r_t^2+ \beta^2).
}
Let us also observe that 
\eqal{}{
\Delta_{11}'&= \langle  -\bar{\bfg}(\thet_{t-\tau_t}) + \bfg(\thet_{t-\tau_t}, o_{t-\tau_t}), \xk - \xs \rangle
\\&
= \underbrace{\langle  -\bar{\bfg}(\thet_{t-\tau_t}) + \bar{\bfg}(\thet_{t- \tau_t -\tmix}), \xk - \xs \rangle}_{\bar{\Delta}_{1}}
\\&
+ \underbrace{\langle -\bar{\bfg}(\thet_{t- \tau_t- \tmix}) + \bfg(\thet_{t-\tau_t}, o_{t-\tau_t}), \xk - \xs \rangle}_{\bar{\Delta}_{2}}.
}
Now, using Lemma \ref{lemma:lemmaSrikantOurs} and the fact that $\epsilon \leq \beta$, we have
\eqal{eq:LipschitzAndEpsilon}{
\bar{\Delta}_1 &\leq L\left(\|\thet_{t-\tau_t} - \thet_{t-\tau_t -\tmix}\|\right)\|\xk - \xs\|
\\&
\leq 4\alpha\tmix L^2(\|\thet_{t-\tau_t}\| + \beta)r_t
\\&
\leq 4\alpha\tmix L^2(\|\thet_t\| + \epsilon + \beta)r_t
\\&
\leq 4\alpha\tmix L^2(r_t + 3\beta)r_t
\\& 
= 4\alpha\tmix L^2(r_t^2 + 3\beta r_t)
\\&
\leq 4\alpha\tmix L^2(3r_t^2 + 2\beta^2)
.
}
We also have that 
\eqal{}{
\bar{\Delta}_{2} &= \underbrace{\langle -\bar{\bfg}(\thet_{t-\tau_t - \tmix}) + \bfg(\thet_{t- \tau_t- \tmix}, o_{t-\tau_t}), \xk - \xs \rangle}_{\bar{\Delta}_{21}}
\\&
+ \underbrace{\langle -\bfg(\thet_{t-\tau_t- \tmix}, o_{t-\tau_t}) + \bfg(\thet_{t-\tau_t}, o_{t-\tau_t}),\xk - \xs \rangle}_{\bar{\Delta}_{22}}.
}
Observe that with calculations analogous to the ones done to arrive at  \eqref{eq:LipschitzAndEpsilon}, we can obtain
\eqal{}{
\E{\bar{\Delta}_{22}} \leq 4\alpha\tmix L^2(3\E{r_t^2} + 2\beta^2).}
Now, we write
\eqal{}{
\bar{\Delta}_{21} &= \underbrace{\langle -\bar{\bfg}(\thet_{t-\tau_t - \tmix}) + \bfg(\thet_{t- \tau_t- \tmix}, o_{t-\tau_t}), \thet_{t- \tau_t- \tmix}-\xs \rangle}_{\bar{\Delta}_{211}}
\\&
+ \underbrace{\langle -\bar{\bfg}(\thet_{t-\tau_t - \tmix}) + \bfg(\thet_{t- \tau_t- \tmix}, o_{t-\tau_t}),\xk - \thet_{t- \tau_t- \tmix} \rangle}_{\bar{\Delta}_{212}}.
}
We first bound $\bar{\Delta}_{212}$. Using Lemma \ref{lemma:lemmaSrikantOurs} and \eqref{eq:boundGradNorm}, 
\eqal{}{
\bar{\Delta}_{212} &\leq \|\bar{\bfg}(\thet_{t-\tau_t - \tmix}) - \bfg(\thet_{t- \tau_t- \tmix}, o_{t-\tau_t})\|\|\xk - \thet_{t- \tau_t- \tmix}\|
\\&
\leq 2L(\|\thet_{t-\tau_t - \tmix}\| + \sigma)\|\xk - \thet_{t-\tau_t - \tmix}\|
\\&
\leq 2L(\|\xk\| + \|\xk - \thet_{t-\tau_t - \tmix}\| + \sigma)\|\xk - \thet_{t-\tau_t - \tmix}\|
\\&
\leq 2L(\|\xk\| + \|\xk - \thet_{t-\tau_t - \tmix}\| + \sigma)\|\xk - \thet_{t-\tau_t - \tmix}\|.
}
Furthermore, using \eqref{eq:delta1} from Lemma \ref{lemma:lemmaSrikantOurs} and the fact that $\epsilon \leq \alpha$, we have 
\eqal{eq:tMinuTPr}{
\|\xk - \thet_{t-\tau_t - \tmix}\| &\leq \|\thet_t- \thet_{t-\tau_t}\| + \|\thet_{t-\tau_t} - \thet_{t-\tau_t - \tmix}\|
\\&
\leq \epsilon + \|\thet_{t-\tau_t} - \thet_{t-\tau_t - \tmix}\|
\\&
\leq \epsilon + 4\alpha\tmix L(r_{t-\tau_t} + 2\beta)
\\&
\leq \alpha + 4\alpha\tmix L(r_{t} + \epsilon + 2\beta)
\\&
\leq13\alpha\tmix L(r_{t} + \beta).
}
Hence, given that $\sigma \leq \beta$ and $\alpha\tmix \leq \frac{1}{26L^2}$, 
\eqal{}{
\bar{\Delta}_{212} &\leq 2L(\|\xk\| + \|\xk - \thet_{t-\tau_t - \tmix}\| + \sigma)\|\xk - \thet_{t-\tau_t - \tmix}\|
\\&
\overset{\eqref{eq:halfSq}}\leq \frac{2L}{26\alpha\tmix L}\|\thet_t-\thet_{t-\tmix-\tau_t}\|^2 + \frac{13\alpha\tmix L^2}{2}(\|\thet_t\| + \|\thet_{t} - \thet_{t-\tmix - \tau_t}\| + \beta)^2
\\&
\leq 26\alpha\tmix L^2(r_t^2 + \beta^2) + \frac{13\alpha\tmix L^2}{2}(2(\|\thet_t\| + \|\thet_t - \thet_{t-\tmix-\tau_t}\|)^2 + 2\sigma^2) 
\\&
\leq 26\alpha\tmix L^2(r_t^2 + \beta^2) + {13\alpha\tmix L^2}(2\|\thet_t\|^2 + 2\|\thet_t - \thet_{t-\tmix-\tau_t}\|^2 + \sigma^2)
\\&
\leq 26\alpha\tmix L^2(r_t^2 + \beta^2) + 13\alpha\tmix L^2(4r_t^2 + 4\beta^2 + 2(13^2)\alpha^2\tmix^2L^22(r_t^2 + \beta^2) + \sigma^2)
\\&
\leq 92\alpha\tmix L^2(r_t^2 + \beta^2).
 }
To conclude, we bound $\E{\bar{\Delta}_{211}}$. We use \eqref{eq:mixingEq}, the fact that $\alpha\tmix \leq 1$ and \eqref{eq:tMinuTPr}. For notational  convenience, define $t' \triangleq t-\tau_t - \tmix$. Using manipulations similar to the ones performed above, we can get
\eqal{}{
\E{\bar{\Delta}_{211}} &= \E{\langle \bar{\bfg}(\thet_{t'}) - \bfg(\thet_{t'}, o_{t-\tau_t}), \thet_{t'}-\xs\rangle}
\\&
= \E{\langle \bar{\bfg}(\thet_{t'}) - \E{\bfg(\thet_{t'}, o_{t-\tau_t})|\thet_{t'}, o_{t'}}, \thet_{t'}-\xs\rangle}
\\&
\leq \E{\|\bar{\bfg}(\thet_{t'}) - \E{\bfg(\thet_{t'}, o_{t-\tau_t})\|}\|\thet_{t'} - \xs\|}
\\&
\leq \E{(\alpha)(\|\thet_{t'}\| + \sigma)(\|\thet_{t'} - \thet_t\| + \|\xk - \xs\|)}
\\&
\leq \alpha\E{(r_t + \|\thet_{t'} - \xk\| + \sigma)(r_t +\|\thet_{t'} - \xk\|+ \beta)}
\\& 
\leq 196L^2\alpha (\E{r_t^2} + \beta^2).
}
Putting together the above bounds, we can conclude that
\eqal{}{\E{-K_{t,3}}\leq C_{12}L^2\alpha\tau_{mix}( \E{r_{t}^2} + \beta^2),}
where $C_{12}=440$.
\hspace*{\fill}~$\square$\vspace{0.4cm}\noindent\\

\noindent\textbf{Proof of Lemma~\ref{th:lemmaAtleast}.} The proof follow the same arguments as in~\cite{cohen2021asynchronous}, we provide details only for completeness. Recall the definition of the average delay up to time $T$, $\tau_{avg}\triangleq \frac{1}{T}\sum_{t=1}^{T}\tau_t$.
Consider $U_{2\tau_{avg}}$, the number of steps $t$ for which the delay $\tau_t$ is at least $2\tau_{avg}$. We must have $U_{2\tau_{avg}} \leq \frac{T}{2}$ (otherwise the total sum of delays exceeds $\tau_{avg} T$, contradicting the definition of $\tau_{avg}$). On the other hand, let $k$ be the number of updates that the algorithm makes. Let $t_1 < t_2 < \ldots < t_k$ be the steps in which an update is made. Denote $t_0 = 0$ and $t_{k+1} = T$. Now, fix $i$ and consider the steps at times $s_n = t_i + n$ for $n \in [1, 2, \ldots, t_{i+1} - t_i - 1]$. In all those steps no update takes place and $\thet_{s_n} = \thet_{t_i}$. We must have $\tau_{s_n} > n$ for all $n$ (otherwise $\thet_t = \thet_{t-t_{\tau_t}}$ for $t = s_n$ and an update occurs). In particular, we have that $\tau_{s_n} \geq 2\tau_{avg}$ in at least $t_{i+1} - t_i - 1 - 2\tau_{avg}$ steps. Formally,
\eqal{}{
    \text{\# of steps in $[t_i, t_{i+1}]$ with delay bigger or equal to $2\tau_{avg}$}\\\geq \max\{0,t_{i+1} - t_i - 1 - 2\tau_{avg}\}\\\geq t_{i+1} - t_i - 1 - 2\tau_{avg}.
}
Hence,
\begin{align*}
U_{2\tau_{avg}} &\geq \sum_{i=0}^{k-1} (t_{i+1} - t_i - 1 - 2\tau_{avg}) \\
&= T - k(1 + 2\tau_{avg}).
\end{align*}
Finally, it follows that $T - k(1 + 2\tau_{avg}) \leq \frac{T}{2}$ which implies $k \geq \frac{T}{4(\tau_{avg}+1)}$.\hspace*{\fill}~$\square$\vspace{0.4cm}
\subsubsection{Initial Condition }
Now, we provide a bound on the initial condition $r_{t}^2$ for $t\leq{\tmix}$. Note that
\eqal{}{
r_{t+1}^2 &= r_t^2 +2\alpha \langle \thet_t - \xs,\bfg(\thet_{t-\tau_t}, o_{t-\tau_t}) \rangle + \alpha^2\|\bfg(\thet_{t-\tau_t}, o_{t-\tau_t})\|^2
\\&
\overset{\eqref{eq:halfSq}}{\leq }r_t^2 + \alpha r_t^2 + \alpha\|\bfg(\thet_{t-\tau_t}, o_{t-\tau_t})\|^2 + \alpha^2\|\bfg(\thet_{t-\tau_t}, o_{t-\tau_t})\|^2
\\&
\leq (1+\alpha)r_t^2 + 2\alpha\|\bfg(\thet_{t-\tau_t}, o_{t-\tau_t})\|^2
\\&
\overset{\eqref{eq:boundGsquared}}{\leq}(1+\alpha)r_t^2 + 4\alpha L^2(\|\thet_{t-\tau_t}\|^2 + \sigma^2)
\\&
\leq(1+\alpha)r_t^2 + 4\alpha L^2(2\epsilon^2 + 2\|\thet_{t}\|^2 + \sigma^2)
\\&
\leq (1+17\alpha L^2)r_t^2 + 28\alpha L^2\sigma^2
\\&
\leq (1+17\alpha L^2)^{t+1}r_0^2 + 28\alpha L^2\sigma^2\sum_{j = 0}^{t}(1+\alpha L^2)^j.
}
Using $\alpha \leq \frac{1}{68\tmix L^2}$ and $(1+ x) \leq e^x \leq 2$ for $x\leq 0.25$, we have: 
\eqal{eq:initial_cond}{
r_{\tmix}^2 &\leq (1+17\alpha L^2)^{\tmix}\sigma^2 + 28\alpha L^2\sigma^2\tmix(1+17\alpha L^2)^{\tmix}
\\&
\leq e^{17\alpha L^2\tmix}\sigma^2 + 28\alpha L^2\sigma^2\tmix e^{17\alpha L^2\tmix}
\\&
\leq 2\sigma^2 + 56\alpha L^2\sigma^2\tmix
\\&
\leq 3\sigma^2.
}
\subsubsection{Proof of Theorem \ref{thm:picky}}
Putting together the results of Lemma \ref{lemma:tavg}, we have, for $I_t=1$:
\begin{align}
    	\E{\|\xkp - \xs\|^2 }&=\E{ K_{t,1}} + \E{\alpha^2K_{t,2}} - \E{2\alpha K_{t,3}}\notag\\&\leq \left(1- 2\alpha\mu + 264\alpha^2\tmix L^2\right)\E{r_t^2} + 526\alpha^2\tmix L^2\beta^2
     \notag\\&\quad+4\alpha^2L^2\E{r_t^2} + 12\alpha^2c^2+2C_{12}L^2\alpha^2\tau_{mix}( \E{r_{t}^2} + \beta^2)
     \\&\leq \left(1- 2\alpha\mu + 1152\alpha^2\tmix L^2\right)\E{r_t^2}+1418L^2\alpha^2\tau_{mix}\beta^2.
\end{align}
Assume $t^'_0,t^'_1,t^'_2,\dots t^'_k\in [T]$
are iterations in which updates happen after iteration $\tmax + \tmix$, so that ${\tmix+\tmax} \leq t^'_0\leq t^'_1\leq\cdots\leq t^'_k$. Then, we have
$r_j=r_{t^'_i}$ for all $j\in [t^'_i, t^'_{i+1})$. We can write, for all $j\in [t^'_{i+1}, t^'_{i+2})$: 
	\begin{equation}
		\begin{aligned}  
			\E{r_{j}^2}&=\E{r_{t^'_{i+1}}^2}\\&= {\E{\|\thet_{t^'_{i+1}-1} - \thet + \alpha\mathbf{g}(\thet_{t^'_{i+1}-1}, o_{t^'_{i+1}-1})\|^2}}\\& + \alpha^2{\E{\|\mathbf{e}_{t^'_{i+1}-1}\|^2}}\\& - 2\alpha{\E{\langle \mathbf{e}_{t^'_{i+1}-1} \thet_{t^'_{i+1}-1} - \xs + \alpha\mathbf{g}(\thet_{t^'_{i+1}-1}, o_{t^'_{i+1}-1}) \rangle}}\\&= {\E{\|\thet_{t^'_{i}} - \thet + \alpha\mathbf{g}(\thet_{t^'_{i}}, o_{t^'_{i}})\|^2}}+ \alpha^2{\E{\|\mathbf{e}_{t^'_{i}}\|^2}}\\& - 2\alpha{\E{\langle \mathbf{e}_{t^'_{i}} \thet_{t^'_{i}} - \xs + \alpha\mathbf{g}(\thet_{t^'_{i}}, o_{t^'_{i}}) \rangle}}\\ 
			 &\leq {\left(1- 2\alpha\mu + 1152\alpha^2\tmix L^2\right)}\E{r_{t^'_{i}}^2}\\& + 1418L^2\alpha^2\tau_{mix}\beta^2.
		\end{aligned}
	\end{equation}
By choosing the step size to be $\alpha\leq \frac{\mu}{1152 L^2\tau_{mix}}$, we have
 for all $j\in [t^'_{i+1}, t^'_{i+2})$: 
	\begin{equation}
		\begin{aligned}  
			\E{r_{j}^2}
			 &\leq {\left(1- 2\alpha\mu + 1152\alpha^2\tmix L^2\right)}\E{r_{t^'_{i}}^2} + 1418L^2\alpha^2\tau_{mix}\beta^2
   \\& \leq \underbrace{\left(1- \alpha\mu\right)}_{\rho}\E{r_{t^'_{i}}^2} + 1418L^2\alpha^2\tau_{mix}\beta^2.
		\end{aligned}
	\end{equation}

 By recursively using the above equation, we have

\begin{equation}
		\begin{aligned}  
			\E{r_{T}^2}
            \leq \rho^{k}\E{r_{t^'_0}^2 }+ \gamma,
		\end{aligned}
	\end{equation}

 with 
$\rho=(1- \alpha\mu)$ and $\gamma\triangleq1418L^2\alpha\tmix\beta^2\mu^{-1}$. Based on Lemma~\ref{th:lemmaAtleast}, $k\geq \frac{T}{4\tau_{avg}+4}-\tmix$, and based on~\eqref{eq:initial_cond},  we know $\E{r_{t^'_0}^2} = \E{r_{\tmix}^2}\leq 3\sigma^2 $. This yields: 
\begin{equation}
\begin{aligned}  
\E{r_{T}^2}\leq \rho^{k}\E{r_{t^'_0}^2 }+ \gamma \leq \rho^{k}3\sigma^2+ \gamma&\leq \exp{\left(-\alpha\mu\frac{T}{4\tau_{avg}+4}+\alpha\mu\tmix\right)}3\sigma^2+1418\frac{L^2\alpha\tmix(\epsilon + \sigma)^2}{\mu}
\\&\leq\exp{\left(-\alpha\mu\frac{T}{4\tau_{avg}+4}\right)}6\sigma^2 + 1418\frac{L^2\alpha\tmix(\epsilon + \sigma)^2}{\mu},
\end{aligned}
\end{equation}
where we used $ \exp{\left(\frac{\mu^2}{1152L^2}\right)} \leq \exp{\left(\frac{1}{1152L^2}\right)}\leq 2$. 

Additionally, if we set $\epsilon=\alpha$, $\alpha=\alpha_0=\frac{\mu}{1152L^2\tmix}$, we obtain, defining $C_2 \triangleq 1152$,
\eqal{}{\E{r_T^2}\leq \left(\exp\left(\frac{-{\mu}^2T}{4 C_2{L}^2\tau_{mix}(\tau_{avg}+1)}\right)6+8\right)\sigma^2.}

This completes the proof. 
\newpage
\section{Simulation Results}\label{app:sim}
\begin{figure}[t]
\center
\includegraphics[width=0.5\columnwidth, trim ={0cm 0cm 0cm 0cm}, clip]{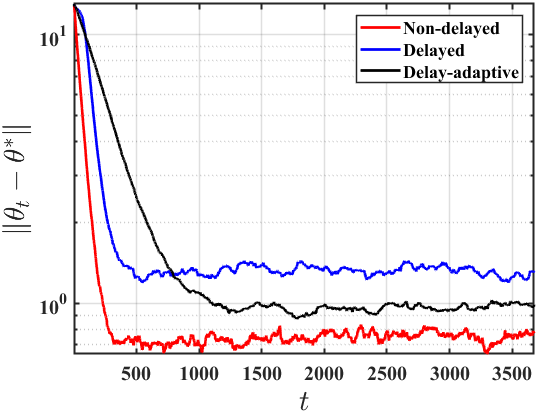}
	\caption{Simulation performance of a TD(0) learning algorithm under delayed updates and Markovian sampling. We compare three different algorithms: a non-delayed TD learning algorithm, a vanilla-delayed algorithm (equivalent to update rule~\eqref{eq:delayedSA}), and a delay-adaptive algorithm (equivalent to update rule~\eqref{eq:algo}).}
\label{fig:1}
\end{figure}
In this section, we show numerical results to validate our theoretical analysis. We simulate a TD(0) learning algorithm for policy evaluation via linear function approximation, with a setting similar to the one adopted by~\cite{dal2023federated} (see technical report) and~\cite{mitra2023temporal} for TD learning under communication constraints. We generate a Markov Decision Process with a state space $\mathcal{S}$ of dimension $|\mathcal{S}| = 20$, resorting to value function approximation with a feature space spanned by $d = 10$ orthonormal basis vectors. We set the discount factor of the problem to $\gamma = 0.5$. We generate the delay sequence by picking a random delay $\tau_{t}$ at each iteration $t$ as a uniform random variable in the range $[1, \tau_{max}]$. We set the maximum delay to $\tau_{max} = 200$. We compare a non-delayed version of TD learning (``non delayed'') with the delayed scenario, for which we implement both the vanilla approach (update rule in~\eqref{eq:delayedSA}) and the proposed delay-adaptive strategy (update rule in \eqref{eq:algo}). For the delay-adaptive algorithm, as required by the theory (Theorem~\ref{thm:picky}), we set $\epsilon = \alpha$. To validate our theoretical results, we compare the results obtained with the different algorithms when the step size is the same for all algorithms. Specifically, we set $\alpha = 0.35$. We show the results in Fig.~\ref{fig:1}. As expected based on the results on Theorem~\ref{thm2} and~\ref{thm:picky}, we see that both the vanilla-delayed and the delay-adaptive algorithms converge to bigger noise balls around the best linear approximation parameter $\xs$. At the same time, we see that the rate of convergence of the delay-adaptive algorithm (for the same choice of step-size) is worse than the vanilla-delayed one, while the convergence ball is smaller (for the delay-adaptive). This is in line with the theoretical results. Indeed, if we substitute the same step size in the bounds of Theorem~\ref{thm2} and~\ref{thm:picky}, we get a worse rate for the delay-adaptive algorithm but a better convergence ball. Note that the main advantage of the delay-adaptive algorithm is the fact that, unlike the vanilla-delayed algorithm, the choice of step size does not inversely depend on the maximum delay in the delay sequence. Thus, the delay-adaptive algorithm allows one to be less conservative in the step-size choice and agnostic of the maximum delay.
\end{document}